\documentclass[acmlarge, nonacm]{acmart}
\AtBeginDocument{}

\usepackage{array}
\usepackage{pifont}
\usepackage{amsmath}
\usepackage{multirow}
\usepackage{listings}
\usepackage{graphicx}
\usepackage{algorithm}
\usepackage{algorithmic}

\begin{document}

\title[ChronoSteer]{ChronoSteer: Bridging Large Language Model and Time Series Foundation Model via Synthetic Cross-Modal Alignment Dataset}

\author{Chengsen Wang}
\authornote{Work performed during the internship at Huawei.}
\authornote{Equal Contribution.}
\email{cswang@bupt.edu.cn}
\orcid{0000-0002-3826-1148}
\affiliation{
  \institution{State Key Laboratory of Networking and Switching Technology, Beijing University of Posts and Telecommunications}
  \city{Beijing}
  \country{China}
}

\author{Qi Qi}
\authornotemark[2]
\email{qiqi8266@bupt.edu.cn}
\orcid{0000-0003-0829-4624}
\affiliation{
  \institution{State Key Laboratory of Networking and Switching Technology, Beijing University of Posts and Telecommunications}
  \city{Beijing}
  \country{China}
}

\author{Zhongwen Rao}
\authornote{Corresponding Authors.}
\email{raozhongwen@huawei.com}
\orcid{0000-0002-1273-4437}
\affiliation{
  \institution{Huawei Technologies Co., Ltd}
  \city{Shenzhen}
  \country{China}
}

\author{Lujia Pan}
\email{panlujia@huawei.com}
\orcid{0000-0002-8988-4740}
\affiliation{
  \institution{Huawei Technologies Co., Ltd}
  \city{Shenzhen}
  \country{China}
}

\author{Jingyu Wang}
\authornotemark[3]
\email{wangjingyu@bupt.edu.cn}
\orcid{0000-0002-2182-2228}
\affiliation{
  \institution{State Key Laboratory of Networking and Switching Technology, Beijing University of Posts and Telecommunications}
  \city{Beijing}
  \country{China}
}

\begin{abstract}

	Conventional forecasting methods are trained end-to-end on unimodal time series, which limits their ability to exploit textual information and undermines their generalization in data-scarce scenarios. Recently, large language models (LLMs) and time series foundation models (TSFMs) have demonstrated powerful capabilities in complex textual reasoning and zero-shot temporal modeling, respectively. Integrating these strengths to construct a multimodal time series foundation model that jointly leverages temporal and textual information for zero-shot future inference has emerged as a promising research direction. However, the scarcity of large-scale, high-quality multimodal datasets remains a fundamental obstacle. To address this challenge, we propose ChronoSteer, a decoupled agentic framework that learns cross-modal alignment from synthetic paired supervision. Specifically, a pretrained LLM first converts textual events into revision instructions that steer the initial unimodal prediction produced by a frozen TSFM. These revision instructions form an intermediate instruction space that bridges the semantic gap between text and time series while fully leveraging pretrained knowledge. Technically, the instructions are discretized into a compact codebook of instruction anchors, effectively mitigating semantic divergence while reducing the cost of dataset construction. Finally, we adopt a two-stage training strategy to recover the fine-grained magnitude information lost during discretization. Furthermore, we release a leakage-controlled multimodal benchmark constructed with temporal separation and textual context available before the prediction window. When paired with an LLM and trained on synthetic cross-modal alignment data, ChronoSteer achieves a 25.8\% improvement in zero-shot prediction accuracy over its unimodal backbone, and outperforms prior state-of-the-art unimodal and multimodal methods by 17.8\% and 22.5\%. These results suggest that synthetic paired supervision is a promising direction for multimodal time series analysis.

\end{abstract}



\maketitle

\section{Introduction}
\label{section:Introduction}

Time series forecasting holds considerable practical value across various domains such as finance \cite{LiuYGG22}, transportation \cite{LiuGZFZMX23}, healthcare \cite{HuangWLLJ25}, and energy \cite{FanFZBZX24}. Human experts usually combine multimodal information to make predictions. For instance, transportation experts often infer flow trends by jointly analyzing historical series and public events. Recently, neural network-based methods \cite{DLinear, PatchTST, HuHQYCX25} have demonstrated strong performance in unimodal forecasting tasks due to powerful modeling capabilities. However, their end-to-end training typically demands substantial computational resources and domain-specific data, which poses high barriers to deployment in low-resource or data-scarce scenarios. To alleviate these challenges, time series foundation models (TSFMs) have emerged as a promising alternative. Pretrained on massive collections of time series, these out-of-the-box models \cite{TimesFM, Moirai} exhibit strong generalization and can perform zero-shot forecasting without additional fine-tuning. Although models like Chronos \cite{Chronos} have shown impressive accuracy, they are pre-trained exclusively on unimodal time series and still lack the capacity to exploit textual information.

Meanwhile, large language models (LLMs) \cite{GPT4, DeepSeekR1} have exhibited remarkable capability in textual reasoning, prompting growing interest in integrating them into time series analysis. Prior attempts to combine language and temporal modeling have followed two broad paradigms. Some approaches \cite{TimeLLM, TGForecaster} transfer LLM weights into time series frameworks and perform cross-modal alignment fine-tuning to build end-to-end models for integrating temporal and textual information. However, they often require dataset-specific adaptation and lack zero-shot generalization. Alternatively, other studies \cite{LLMTIME, LSTPrompt} transform time series into string formats, leveraging prompts to enable LLMs for zero-shot forecasting in unimodal and multimodal scenarios. However, the inherent limitations of LLMs in precise numerical inference and fine-grained temporal reasoning hinder their accuracy in time series analysis \cite{TSFUBenchmark, TSandLanguage}. Another fundamental obstacle for this field is the scarcity of large-scale, high-quality multimodal datasets. \textbf{Consequently, how to effectively integrate the complementary strengths of TSFMs (zero-shot temporal modeling) and LLMs (rich textual reasoning) into a single multimodal framework, especially in the absence of high-quality aligned data, remains a promising yet underexplored direction.}

To address this challenge, we propose ChronoSteer, a decoupled agentic framework that learns cross-modal alignment from synthetic paired supervision. Compared with fine-tuning LLMs to enhance domain-specific capabilities \cite{MLLM4Rec}, equipping lightweight unimodal TSFMs with textual comprehension represents a substantially more computationally efficient path. By freezing the unimodal TSFM backbone and integrating a lightweight branch to process textual information, this paradigm reduces training overhead while strictly preserving the unimodal forecasting accuracy of TSFMs and the event reasoning capabilities of LLMs. Textual descriptions such as backgrounds and events often present semantic complexity and data sparsity, posing significant challenges for direct cross-modal alignment without high-quality paired data. We therefore leverage a pretrained LLM with domain knowledge to reinterpret textual events as concise revision instructions that specify how an initial unimodal prediction produced by a frozen TSFM should be adjusted. This intermediate instruction space both narrows the semantic gap between text and series and fully exploits the pretrained knowledge of LLMs. For example, although an event like \textit{Christmas Holidays} may appear infrequently in raw training datasets, once reinterpreted by LLMs as an instruction such as \textit{Elevate Peaks}, it can be easier for TSFMs to learn and apply. Notably, while some time series editing models \cite{TEdit} have explored attribute-based series tuning, they are restricted to simple attributes and tend to neglect incorporating historical information.

Considering the inherent limitations of LLMs in temporal reasoning \cite{TSFUBenchmark, TSandLanguage}, directly employing them to construct large-scale instruction-series paired datasets is both prohibitively expensive and unreliable. Besides, open-text instructions often exhibit semantic divergence and low signal-to-noise ratios \cite{XiongW0J0Z0L025}, which impose stringent requirements on the quality and scale of aligned data. To address this issue, we adopt a hybrid data synthesis strategy. Specifically, we first employ LLMs to generate revision instructions for a small set of representative samples obtained via clustering. These instructions are subsequently refined and summarized based on expert knowledge and domain-specific priors. Following this summarization, the open-text instructions are discretized into a compact codebook of instruction anchors, which serve as discrete prototypes of temporal adjustment behaviors. For each anchor, we design a corresponding parametric synthesis function to expand the dataset, thereby substantially reducing the cost and complexity of constructing large-scale multimodal datasets. To further recover the fine-grained magnitude information lost during discretization and overcome the expressive limitations of the synthesis functions, we develop a two-stage training framework. The first stage coarsely aligns the model with the anchor-conditioned revision behaviors, while the second stage mines implicit cues from the historical context to refine magnitude-level details and improve temporal consistency.

Moreover, the development of benchmarks \cite{TimeMMD, CiK} for multimodal time series forecasting (MTSF) remains nascent. A significant concern is the potential for data leakage arising from overlaps between benchmark datasets and the corpora used to pretrain LLMs and TSFMs. This can lead to inflated performance metrics during evaluation. To address this concern, we follow the established temporal-update paradigm of ChatTime \cite{ChatTime} to construct MTSFBench-300, a leakage-controlled benchmark covering three representative domains: Energy, Traffic, and Weather. MTSFBench-300 itself is constructed after the evaluated open-source or open-weight checkpoints are fixed and uses only textual context available before the prediction window, so the released checkpoints are not trained on this benchmark and the evaluation does not expose future observations.

The contributions of our paper are summarised as follows:

\begin{itemize}

	\item \textbf{Framework:} We introduce ChronoSteer, a decoupled multimodal framework that effectively bridges LLMs and TSFMs. By freezing the unimodal TSFM backbone and employing a lightweight adapter, ChronoSteer concurrently leverages temporal dynamics and textual instructions for efficient zero-shot future inference without compromising unimodal performance.

	\item \textbf{Strategy:} We design a two-stage training strategy to overcome the scarcity of high-quality multimodal alignment data. By discretizing open-text events into a codebook of instruction anchors and applying parametric synthesis functions, we learn cross-modal alignment without requiring manually annotated real-world cross-modal pairs.

	\item \textbf{Benchmark:} Addressing the risk of data leakage in existing evaluations, we construct MTSFBench-300, a leakage-controlled benchmark spanning the Energy, Traffic, and Weather domains. It provides a semantically aligned platform for evaluating multimodal forecasting under a clear temporal-separation protocol.

	\item \textbf{Performance:} We conduct extensive experiments comparing ChronoSteer with mainstream baselines. The results show that ChronoSteer, trained on synthetic cross-modal alignment pairs, improves zero-shot multimodal forecasting accuracy on MTSFBench-300 under the tested settings.

\end{itemize}

\section{Related Work}
\label{section:Related_Work}

\subsection{Unimodal Time Series Forecasting}
\label{subsection:Unimodal_Time_Series_Forecasting}

As a critical real-world challenge, time series forecasting has attracted considerable attention. Early research mainly relies on statistical methods like ARIMA \cite{ARIMA}, which uses moving averages for prediction. However, these approaches often struggle with the complexity of real-world scenarios. With deep learning progress, neural network-based approaches have grown in popularity. RNNs \cite{LSTM, DeepAR} capture temporal dependency through sequential architecture, yet encounter limitations such as vanishing/exploding gradients and information loss when handling long series. To overcome these challenges, researchers have introduced CNNs \cite{MICN, TimesNet} and Transformers \cite{Informer, iTransformer} to capture long-term temporal dependency. While these methods have achieved notable improvements, their training typically demands significant computational resources, domain expertise, and abundant data, posing high barriers to practical deployment. In response, TSFMs have emerged as a promising alternative. Pretrained on extensive datasets, these models \cite{MOMENT, TimeMoE, Sundial} exhibit robust generalization to support zero-shot forecasting. Nonetheless, most existing TSFMs are trained solely on unimodal time series, lacking the capacity to process multimodal inputs such as text information, which constrains the applicability across diverse scenarios. Unlike these purely unimodal approaches, our proposed ChronoSteer equips frozen TSFMs with a lightweight adapter to interpret textual instructions. This design preserves the superior numerical forecasting capability of TSFMs while extending their applicability to multimodal contexts without expensive retraining.

\subsection{Multimodal Time Series Forecasting}
\label{subsection:Multimodal_Time_Series_Forecasting}

The LLMs \cite{LLaMA2, QwQ32B} with strong text reasoning capabilities have brought new opportunities to time series analysis. Some approaches \cite{S2IPLLM, AutoTimes, ContextAlignment} transfer LLM weights into time series frameworks and apply cross-modal alignment fine-tuning to enable end-to-end methods for integrating temporal and textual information. However, they often require dataset-specific adaptation and lack zero-shot generalization. Moreover, the effectiveness of directly applying text-pretrained weights to model numerical time series still remains uncertain \cite{TanMGAH24}. Alternatively, other studies \cite{LLMTIME, LSTPrompt} transform time series into string formats, leveraging prompts to enable LLMs for zero-shot forecasting in unimodal and multimodal scenarios. However, the inherent limitations of LLMs in temporal understanding, specifically the challenges in extracting complex features from numerical data \cite{TimeSeriesExam, TimerBed} and generating precise values with intricate attributes \cite{LLMTIME}, hinder their ability to perform accurate numerical inference. While instruction fine-tuning brings partial improvements \cite{TPLLM}, the constrained training data undermines broader generalization. The development of multimodal time series foundation models capable of zero-shot future inference by jointly leveraging textual and temporal information remains largely unexplored. As the first attempt, ChatTime \cite{ChatTime} still relies on LLMs to perform zero-shot forecasting, without incorporating TSFMs that are specifically designed for numerical modeling. In contrast to these approaches, ChronoSteer not only achieves zero-shot generalization but also strategically decouples the modeling process: it assigns temporal reasoning to a specialized TSFM while restricting the LLM to semantic instruction generation. Unlike prior methods that force language models to execute precise numerical predictions, our strategy effectively synergizes the numerical precision of TSFMs with the rich semantic understanding of LLMs.

The time series editing task in TEdit \cite{TEdit} shares similarities with our work. However, TEdit focuses on adjusting series based on specified attributes, while our goal is to generate future series conditioned on both instructions and historical data. Ignoring historical information limits their control over modification extent. Moreover, TEdit supports only relatively restricted attributes, allowing only complete changes in trends or cycles. It also lacks zero-shot generalization across various datasets. Different from TEdit, ChronoSteer possesses zero-shot generalization capabilities across diverse scenarios. Moreover, it generates future series conditioned on both textual instructions and historical data, ensuring that the predicted outcome is both semantically consistent with text and temporally coherent with historical context.

Several studies \cite{Terra, TimeMMD} have introduced benchmark datasets for MTSF. However, the sources of these datasets are largely outdated. Considering the ongoing expansion of corpus size required for LLMs and TSFMs, existing benchmarks may have already been included in training data. This overlap leads to information leakage during evaluation, undermining the reliability of experiment results. To address this issue, we construct MTSFBench-300 after the evaluated checkpoints are fixed and use only context available before the prediction window during evaluation. This temporal-separation protocol provides a more rigorous platform for evaluating zero-shot multimodal forecasting capabilities.

\section{Methodology}
\label{section:Methodology}

To enhance the distinction, a standardized notation for future series is introduced: ground truth is $\mathbf{y}$, unimodal prediction is $\bar{\mathbf{y}}$, multimodal prediction is $\hat{\mathbf{y}}$, and function-transformed series is $\tilde{\mathbf{y}}$.

\subsection{Problem Definition}
\label{subsection:Problem_Definition}

This paper explores MTSF, aiming to concurrently leverage both temporal and textual information for future inference. Consistent with most previous TSFMs \cite{Chronos, TimeLLM, ChatTime, TimeGPT}, we focus on univariate time series. As for multivariate time series, they can be decomposed into multiple univariate series under the assumption of variable independence \cite{DLinear}. Specifically, each input instance is represented by a multimodal pair $\left( \mathbf{x}, \mathbf{e} \right)$, where $\mathbf{x} \in \mathbb{R}^{H \times 1}$ denotes a univariate series of historical observations of length $H$, and $\mathbf{e}$ indicates the corresponding textual information. The main goal is to build a generalizable multimodal forecasting framework $\mathcal{M} \left( \cdot, \cdot \right)$ to predict future observations $\hat{\mathbf{y}} \in \mathbb{R}^{P \times 1}$ over $P$ time steps:
\begin{equation}
	\begin{gathered}
		\begin{aligned}
			\hat{\mathbf{y}} = \mathcal{M} \left( \mathbf{x}, \mathbf{e} \right)
		\end{aligned}
	\end{gathered}
	\label{eq1}
\end{equation}

Moreover, we expect the TSFMs to possess zero-shot forecasting capability, meaning that after training on large-scale datasets, the model can generate accurate predictions in downstream scenarios without any fine-tuning. To ensure fairer evaluation, benchmarks should be constructed after the evaluated checkpoints are fixed and should use only information available before the prediction window.

\subsection{Framework Overview}
\label{subsection:Framework_Overview}

\begin{figure}[t]
	\centering
	\resizebox{0.9\linewidth}{!}
	{
		\includegraphics{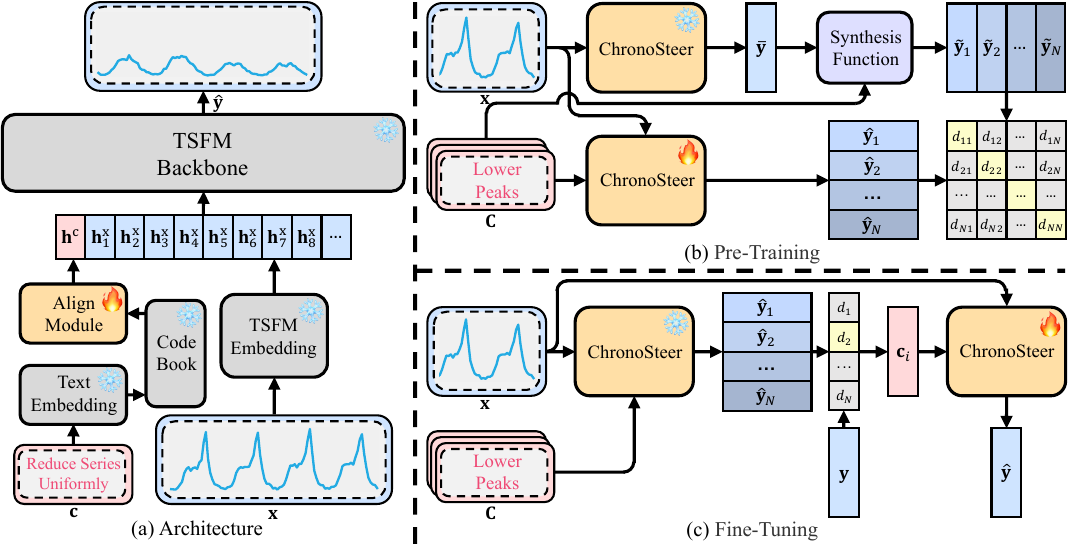}
	}
	\caption{The overview of ChronoSteer. Spark icons indicate trainable parameters, while snowflake icons denote frozen components. (a) illustrates the overall architecture, preserving the unimodal TSFM backbone while integrating a lightweight branch to process textual instruction. To mitigate the scarcity of cross-modal paired data, we introduce a two-stage training strategy: (b) synthetic data driven pre-training followed by (c) pseudo label guided fine-tuning.}
	\label{fig1}
\end{figure}

Textual descriptions such as contextual backgrounds and event narratives often involve high semantic complexity and data sparsity, which create significant challenges for achieving effective cross-modal alignment without high-quality paired data. To address this issue, we propose a decoupled multimodal forecasting framework that bridges LLMs and TSFMs through textual revision instructions. Specifically, the TSFM $\mathcal{M}_\text{series}$ first generates unimodal predictions $\bar{\mathbf{y}} \in \mathbb{R}^{P \times 1}$ based on historical observations $\mathbf{x}$. Next, an LLM $\mathcal{M}_\text{text}$ generates textual revision instructions $\mathbf{c}$ by comprehensively analyzing the historical observations $\mathbf{x}$, the unimodal predictions $\bar{\mathbf{y}}$, and the associated textual context $\mathbf{e}$. Finally, $\mathcal{M}_\text{series}$ steered by instructions $\mathbf{c}$ produces the multimodal predictions $\hat{\mathbf{y}}$. This three-stage decomposition shifts the LLM task from value generation to event influence analysis, which aligns better with its textual reasoning strengths. The intermediate instruction space then offers a simpler interface between text semantics and TSFM prediction. This inference paradigm can be formalized as:
\begin{equation}
	\begin{gathered}
		\begin{aligned}
			\bar{\mathbf{y}} & = \mathcal{M}_\text{series} \left( \mathbf{x}, \varnothing \right)                \\
			\mathbf{c}       & = \mathcal{M}_\text{text} \left( \mathbf{x}, \bar{\mathbf{y}}, \mathbf{e} \right) \\
			\hat{\mathbf{y}} & = \mathcal{M}_\text{series} \left( \mathbf{x}, \mathbf{c} \right)                 \\
		\end{aligned}
	\end{gathered}
	\label{eq2}
\end{equation}

This work aims to bridge LLMs and TSFMs by developing a multimodal TSFM $\mathcal{M}_\text{series}$ that can be dynamically guided by textual revision instructions. The detailed process for generating these revision instructions, described in Appendix \ref{appendix:Detailed_Prediction_Showcase}, is beyond the primary scope of this paper. With the ongoing advancements in LLMs \cite{DeepSeekR1}, the accuracy of semantic transformation is expected to improve continuously, which further strengthens the practicality of our framework.

As illustrated in Figure \ref{fig1}(a), we develop ChronoSteer, a multimodal TSFM capable of being steered by textual revision instructions. Technically, we introduce a lightweight branch to process textual instructions while preserving the original unimodal backbone of the TSFM. In the absence of textual input, ChronoSteer naturally reverts to the unimodal forecasting mode of the backbone, ensuring stability and adaptability in practical applications. Furthermore, to mitigate the scarcity of cross-modal instruction-series paired data, we design a two-stage training strategy (Figures \ref{fig1}(b) and \ref{fig1}(c)) built on synthetic alignment pairs. Our experiments show that this training strategy can improve zero-shot multimodal forecasting without relying on manually annotated cross-modal pairs.

\subsection{ChronoSteer Architecture}
\label{subsection:ChronoSteer_Architecture}

The architecture of ChronoSteer is illustrated in Figure \ref{fig1}(a). Given an input pair $\left( \mathbf{x}, \mathbf{c} \right)$, the model first performs feature embedding. Specifically, the time series $\mathbf{x}$ is transformed into a sequence of tokens $\mathbf{H}^\text{x}=\left\{ \mathbf{h}_1^\text{x}, \ldots, \mathbf{h}_L^\text{x} \right\} \in \mathbb{R}^{L \times D_\text{series}}$ through the embedding layer in pre-trained unimodal TSFM. Here, $L$ represents the number of embedding tokens, which typically corresponds to the number of patches, and $D_\text{series}$ is the dimension of time series embedding. Meanwhile, the textual instruction $\mathbf{c}$ is mapped into a text embedding $\mathbf{h}^\text{c} \in \mathbb{R}^{1 \times D_\text{text}}$ using a pre-trained text embedding model, where $D_\text{text}$ is the dimension of the text embedding.

To alleviate semantic divergence in instructions \cite{XiongW0J0Z0L025} and the scarcity of paired data, as discussed in Section \ref{subsubsection:Multimodal_Triplet_Construction}, we introduce an anchor matching mechanism. We construct a compact codebook $\mathbf{C} = \left\{ \mathbf{c}_1, \ldots, \mathbf{c}_N \right\} \in \mathbb{R}^{N \times D_\text{text}}$ that contains $N$ anchor revision instructions (with $N=9$ in our experiments). These anchors act as discrete prototypes that represent different types of temporal adjustment behaviors. The cosine similarity retrieval is employed to map $\mathbf{h}^\text{c}$ to its nearest anchor $\bar{\mathbf{h}}^\text{c} \in \mathbb{R}^{1 \times D_\text{text}}$:
\begin{equation}
	\begin{gathered}
		\begin{aligned}
			\bar{\mathbf{h}}^\text{c} =\mathbf{c}_i,\ \ \operatorname{where} \ \ i=\arg\max _{j \in\left[1, N\right]} \cos\left(\mathbf{h}^\text{c}, \mathbf{c}_j\right)
		\end{aligned}
	\end{gathered}
	\label{eq3}
\end{equation}

A Multilayer Perceptron (MLP) is used for alignment of dimension and semantics:
\begin{equation}
	\begin{gathered}
		\begin{aligned}
			\hat{\mathbf{h}}^\text{c} = \operatorname{ReLU} \left( \bar{\mathbf{h}}^\text{c} \mathbf{W}_1 + \mathbf{b}_1 \right) \mathbf{W}_2 + \mathbf{b}_2
		\end{aligned}
	\end{gathered}
	\label{eq4}
\end{equation}
where $\mathbf{W}_1 \in \mathbb{R}^{D_\text{text} \times D_\text{hidden}}$ and $\mathbf{W}_2 \in \mathbb{R}^{D_\text{hidden} \times D_\text{series}}$ are weights, $D_\text{hidden}$ is the hidden layer dimension. $\mathbf{b}_1 \in \mathbb{R}^{1 \times D_\text{hidden}}$, $\mathbf{b}_2 \in \mathbb{R}^{1 \times D_\text{series}}$ are biases. Notably, only this module needs to be optimized.

Given that Transformers tend to allocate more attention to the initial token \cite{UPFT}, $\hat{\mathbf{h}}^\text{c} \in \mathbb{R}^{1 \times D_\text{series}}$ is concatenated to the start of $\mathbf{H}^\text{x}$, resulting in a combined representation $\mathbf{H} = \left[ \hat{\mathbf{h}}^\text{c}; \mathbf{H}^\text{x} \right] \in \mathbb{R}^{ \left( L+1 \right) \times D_\text{series}}$. After processing by frozen TSFM backbone, the final prediction $\hat{\mathbf{y}}$ is produced.

\begin{algorithm}[t]
	\centering
	\caption{Multimodal Triplet Construction}
	\label{alg1}
	\begin{algorithmic}[1]
	\STATE \textbf{Input:} Historical series $\mathbf{X}$, Instruction codebook $\mathbf{C}$, Transformation functions $\mathcal{F}$, ChronoSteer $\mathcal{M}_\text{series}$ with frozen initial parameters.
	\STATE \textbf{Output:} Pre-training dataset $\mathcal{D}_\text{PT}$.
	\STATE

	\STATE Initialize $\mathcal{D}_\text{PT} \leftarrow \varnothing$
	\FOR{each historical series $\mathbf{x}_i \in \mathbf{X}$}
	\STATE $\bar{\mathbf{y}}_i \leftarrow \mathcal{M}_\text{series}\left(\mathbf{x}_i, \varnothing\right)$
	\FOR{each anchor $j \in \left(\{1, \ldots, N\right\}$}
	\STATE $\tilde{\mathbf{y}}_{ij} \leftarrow f_j\left((\bar{\mathbf{y}}_i\right)$
	\STATE Add triplet $\left((\mathbf{x}_i, \mathbf{c}_j, \tilde{\mathbf{y}}_{ij}\right)$ to $\mathcal{D}_\text{PT}$
	\ENDFOR
	\ENDFOR
	\STATE \textbf{return} $\mathcal{D}_\text{PT}$
\end{algorithmic}
\end{algorithm}

\subsection{Synthetic Data Driven Pre-training}
\label{subsection:Synthetic_Data_Driven_Pre-training}

\subsubsection{Multimodal Triplet Construction}
\label{subsubsection:Multimodal_Triplet_Construction}

We utilize Monash Repository \cite{Monash} as our training data source. A sliding window (length 160, stride 32) slices the original series into about 15M slices. Each slice is independently normalized and split into historical series $\mathbf{x}$ (length 128) and prediction series $\mathbf{y}$ (length 32). K-means clustering \cite{Kmeans} is applied to optimize efficiency, like ChatTime \cite{ChatTime}, grouping samples into 1K, 10K and 100K clusters, respectively. We randomly sample one slice from each cluster to balance diversity and efficiency.

A practical approach to constructing the instruction-series paired dataset is to input both the unimodal prediction $\bar{\mathbf{y}}$ generated by the TSFMs and the corresponding ground-truth labels $\mathbf{y}$ into LLMs, prompting them to produce revision instructions. We validated this approach on the representative dataset containing 1K samples. However, due to the inherent limitations of LLMs in temporal reasoning \cite{TSFUBenchmark, TSandLanguage}, the accuracy of the generated instructions in coarse-grained revision directions is only about 68\%, as evaluated by five human experts. The performance in specifying fine-grained adjustment magnitudes is even less satisfactory. Prior research has shown that open-text instructions often suffer from semantic ambiguity and a low signal-to-noise ratio \cite{XiongW0J0Z0L025}, which imposes stringent requirements on the quality and scale of aligned data. Considering that directly synthesizing large-scale data with LLMs is both costly and unreliable, we adopt a more conservative but controllable function-based strategy.

Specifically, these open-text instructions are refined and summarized based on expert knowledge. Considering both the coverage of the codebook over 1K representative samples and the feasibility of constructing synthetic functions, we consolidate the 1K open-text instructions into nine core anchors, which represent discrete prototypes of temporal adjustment behaviors. For each anchor (\textit{Keep Unchanged}, \textit{Increase Trend}, \textit{Reduce Trend}, \textit{Expand Amplitude}, \textit{Compress Amplitude}, \textit{Elevate Peaks}, \textit{Lower Peaks}, \textit{Raise Troughs}, and \textit{Deepen Troughs}), we design a corresponding transformation function, as described in Algorithm \ref{alg5}. These functions are then applied to 100K slices, generating the ChronoSteer-100K-PT dataset, which contains 900K precisely aligned synthetic triplets $\left( \mathbf{x}, \mathbf{c}, \tilde{\mathbf{y}} \right)$. Since fine-grained adjustment magnitudes are difficult to obtain in evaluation or practice, we exclude them from the anchor instructions. This process is illustrated in the first row of Figure \ref{fig1}(b) and Algorithm \ref{alg1}. Specifically, for each historical series $\mathbf{x}$, a unimodal TSFM first predicts the initial output $\bar{\mathbf{y}}$. Subsequently, the nine transformation functions, each corresponding to an anchor revision instruction within the instruction codebook $\mathbf{C}=\left\{ \mathbf{c}_1, \ldots, \mathbf{c}_N \right\}$, are applied to $\bar{\mathbf{y}}$, producing enhanced series $\tilde{\mathbf{Y}} = \left\{ \tilde{\mathbf{y}}_1, \ldots, \tilde{\mathbf{y}}_N \right\} \in \mathbb{R}^{N \times P}$.

\subsubsection{Cross-Modal Contrastive Pretraining}
\label{subsubsection:Cross-Modal_Contrastive_Pretraining}

\begin{algorithm}[t]
	\centering
	\caption{Cross-Modal Contrastive Pretraining}
	\label{alg2}
	\begin{algorithmic}[1]
	\STATE \textbf{Input:} Pre-training dataset $\mathcal{D}_\text{PT}$, Loss weight $\alpha$, ChronoSteer $\mathcal{M}_\text{series}$ with trainable initial parameters.
	\STATE \textbf{Output:} ChronoSteer $\mathcal{M}_\text{series}$ with pre-trained parameters.
	\STATE

	\WHILE{not converged}
	\STATE Sample $\left(\mathbf{x}, \mathbf{C}, \tilde{\mathbf{Y}}\right)$ from $\mathcal{D}_\text{PT}$
	\STATE Initialize multimodal predictions $\hat{\mathbf{Y}} \leftarrow \varnothing$

	\FOR{each instruction $\mathbf{c}_j \in \mathbf{C}$}
	\STATE $\hat{\mathbf{y}}_j \leftarrow \mathcal{M}_\text{series}\left(\mathbf{x}, \mathbf{c}_j\right)$
	\STATE Add $\hat{\mathbf{y}}_j$ to $\hat{\mathbf{Y}}$
	\ENDFOR
	\STATE $\mathcal{L}_{\text{total}} \leftarrow \mathcal{L}_{\text{MSE}}\left(\hat{\mathbf{Y}}, \tilde{\mathbf{Y}}\right) + \alpha \mathcal{L}_{\text{Con}}\left(\hat{\mathbf{Y}}, \tilde{\mathbf{Y}}\right)$

	\STATE Update $\mathcal{M}_\text{series}$ to minimize $\mathcal{L}_{\text{total}}$
	\ENDWHILE
	\STATE \textbf{return} $\mathcal{M}_\text{series}$
\end{algorithmic}
\end{algorithm}

As illustrated in the second row of Figure \ref{fig1}(b) and Algorithm \ref{alg2}, for each triplet $\left( \mathbf{x}, \mathbf{C}, \tilde{\mathbf{Y}} \right)$ in ChronoSteer-100K-PT, multimodal predictions $\hat{\mathbf{Y}} = \left\{ \hat{\mathbf{y}}_1, \ldots, \hat{\mathbf{y}}_N \right\} \in \mathbb{R}^{N \times P}$ are generated by ChronoSteer:
\begin{equation}
	\begin{gathered}
		\begin{aligned}
			\hat{\mathbf{y}}_i = \operatorname{ChronoSteer} \left( \mathbf{x}, \mathbf{c}_i \right)
		\end{aligned}
	\end{gathered}
	\label{eq5}
\end{equation}

In addition to the Mean Squared Error (MSE) loss $\mathcal{L}_{\text{MSE}}$, a contrastive loss \cite{CLIP} is introduced to further enhance cross-modal alignment. This loss minimizes the distance between the function output $\tilde{\mathbf{y}}_i$ and the predicted output $\hat{\mathbf{y}}_i$ for the same text instruction $\mathbf{c}_i$, while maximizing the distance to outputs from other instructions.
\begin{equation}
	\begin{gathered}
		\begin{aligned}
			\mathcal{L}_{\text{Con}} =-\frac{1}{2 N} \sum_{i=1}^N\Bigg[ & \log \frac{\exp \left(\cos \left(\hat{\mathbf{y}}_i, \tilde{\mathbf{y}}_i\right)\right)}{\sum_{j=1}^N \exp \left(\cos \left(\hat{\mathbf{y}}_i, \tilde{\mathbf{y}}_j\right) \right)} + \log \frac{\exp \left(\cos \left(\tilde{\mathbf{y}}_i, \hat{\mathbf{y}}_i\right)\right)}{\sum_{j=1}^N \exp \left(\cos \left(\tilde{\mathbf{y}}_i, \hat{\mathbf{y}}_j\right)\right)}\Bigg]
		\end{aligned}
	\end{gathered}
	\label{eq6}
\end{equation}

Finally, the complete loss function for a single instance is:
\begin{equation}
	\begin{gathered}
		\begin{aligned}
			\mathcal{L}_{\text{total}} = \mathcal{L}_{\text{MSE}} + \alpha \mathcal{L}_{\text{Con}}
		\end{aligned}
	\end{gathered}
	\label{eq7}
\end{equation}
where $\alpha$ is used to balance the relative contributions.

\subsection{Pseudo Label Guided Fine-Tuning}
\label{subsection:Pseudo_Label_Guided_Fine-Tuning}

\subsubsection{Pseudo Label Generation}
\label{subsubsection:Pseudo_Label_Generation}

\begin{algorithm}[t]
	\centering
	\caption{Pseudo Label Generation}
	\label{alg3}
	\begin{algorithmic}[1]
	\STATE \textbf{Input:} Historical series $\mathbf{X}$, Ground truth series $\mathbf{Y}$, Instruction codebook $\mathbf{C}$, ChronoSteer $\mathcal{M}_\text{series}$ with frozen pre-trained parameters.
	\STATE \textbf{Output:} Fine-tuning dataset $\mathcal{D}_\text{FT}$.
	\STATE

	\STATE Initialize $\mathcal{D}_\text{FT} \leftarrow \varnothing$
	\FOR{each sample $\left(\mathbf{x}_i, \mathbf{y}_i\right) \in \left(\mathbf{X}, \mathbf{Y}\right)$}
	\FOR{each anchor $j \in \left\{1, \ldots, N\right\}$}
	\STATE $\hat{\mathbf{y}}_{ij} \leftarrow \mathcal{M}_\text{series}\left(\mathbf{x}_i, \mathbf{c}_j\right)$
	\ENDFOR
	\STATE $k \leftarrow \arg\min_{j} \operatorname{MSE}\left(\hat{\mathbf{y}}_{ij}, \mathbf{y}_i\right)$
	\STATE $\hat{\mathbf{c}}_i \leftarrow \mathbf{c}_k$

	\STATE Add triplet $\left(\mathbf{x}_i, \hat{\mathbf{c}}_i, \mathbf{y}_i\right)$ to $\mathcal{D}_\text{FT}$
	\ENDFOR
	\STATE \textbf{return} $\mathcal{D}_\text{FT}$
\end{algorithmic}
\end{algorithm}

\begin{algorithm}[t]
	\centering
	\caption{Self-Supervised Continuous Fine-Tuning}
	\label{alg4}
	\begin{algorithmic}[1]
	\STATE \textbf{Input:} Fine-tuning dataset $\mathcal{D}_\text{FT}$, ChronoSteer $\mathcal{M}_\text{series}$ with trainable pre-trained parameters.
	\STATE \textbf{Output:} ChronoSteer $\mathcal{M}_\text{series}$ with fine-tuned parameters.
	\STATE

	\WHILE{not converged}
	\STATE Sample $\left(\mathbf{x}, \hat{\mathbf{c}}, \mathbf{y}\right)$ from $\mathcal{D}_\text{FT}$
	\STATE $\hat{\mathbf{y}} \leftarrow \mathcal{M}_\text{series}\left(\mathbf{x}, \hat{\mathbf{c}}\right)$
	\STATE $\mathcal{L}_{\text{MSE}} \leftarrow \operatorname{MSE}\left(\hat{\mathbf{y}}, \mathbf{y}\right)$

	\STATE Update $\mathcal{M}_\text{series}$ to minimize $\mathcal{L}_{\text{MSE}}$
	\ENDWHILE
	\STATE \textbf{return} $\mathcal{M}_\text{series}$
\end{algorithmic}
\end{algorithm}

Considering that directly employing LLMs to construct large-scale instruction-series paired datasets is both prohibitively expensive and unreliable, we adopt a function-based synthesis strategy without specifying magnitude during the pretraining stage. These functions provide controllable directional supervision, but they do not fully reproduce the nonlinear, delayed, or domain-specific real-world event effects. Using only function-transformed future series $\tilde{\mathbf{y}}$ as targets discards fine-grained magnitude information and confines the model to the predefined transformation functions. To address this issue, we adopt observed future series $\mathbf{y}$ as the optimization target during fine-tuning. This approach promotes the model to extract implicit cues from the historical context, refine magnitude-level details, and improve temporal consistency.

Technically, we propose an optimal matching-based pseudo-label generation strategy, shown in the left part of Figure \ref{fig1}(c) and Algorithm \ref{alg3}, to create revision instructions from unimodal predictions $\bar{\mathbf{y}}$ to align with $\mathbf{y}$. For each $\mathbf{x}$, we freeze the pre-trained ChronoSteer and enumerate all anchors in the codebook to obtain a multimodal prediction set $\hat{\mathbf{Y}} = \left\{ \hat{\mathbf{y}}_1, \ldots, \hat{\mathbf{y}}_N \right\} \in \mathbb{R}^{N \times P}$, where each candidate corresponds to one anchor instruction. The prediction $\hat{\mathbf{y}}_i$ closest to $\mathbf{y}$ is selected based on MSE, and its corresponding anchor instruction $\mathbf{c}_i$ is assigned as the pseudo-label $\hat{\mathbf{c}}$:
\begin{equation}
	\begin{gathered}
		\begin{aligned}
			\hat{\mathbf{c}} = \mathbf{c}_i,\ \ \text{where} \ \ i = \arg\min_{j \in \left[1,N\right]} \operatorname{MSE}\left(\hat{\mathbf{y}}_j, \mathbf{y}\right)
		\end{aligned}
	\end{gathered}
	\label{eq8}
\end{equation}

This pseudo-labeling process remains stable because the labels are generated offline and kept fixed during fine-tuning. For each sample, we enumerate all anchors in the finite codebook, generate the corresponding candidate forecasts, and select the anchor whose forecast is closest to the ground truth in terms of MSE. Because the labels are not updated iteratively, our method avoids the label drift commonly observed in self-training. For example, if the forecast under \textit{Lower Peaks} is closer to the ground truth than those under \textit{Increase Trend} and \textit{Keep Unchanged}, then \textit{Lower Peaks} is used as the pseudo-label. In the end, we obtain the ChronoSteer-100K-FT dataset containing 100K aligned synthetic triplets $\left( \mathbf{x}, \hat{\mathbf{c}}, \mathbf{y} \right)$, where $\mathbf{y}$ is the observed future series and the association between $\hat{\mathbf{c}}$ and $\mathbf{y}$ is automatically established by anchor matching rather than manually annotated from real-world cross-modal pairs.

\subsubsection{Self-Supervised Continuous Fine-Tuning}
\label{subsubsection:Self-Supervised_Continuous_Fine-Tuning}

The fine-tuning process is illustrated in the right side of Figure \ref{fig1}(c) and Algorithm \ref{alg4}. For each sample in ChronoSteer-100K-FT, ChronoSteer generates corresponding predictions:
\begin{equation}
	\begin{gathered}
		\begin{aligned}
			\hat{\mathbf{y}} = \operatorname{ChronoSteer} \left( \mathbf{x}, \hat{\mathbf{c}} \right)
		\end{aligned}
	\end{gathered}
	\label{eq9}
\end{equation}

During fine-tuning, parameter optimization exclusively uses MSE as the loss function. \textcolor{black}{Although the observed future series $\mathbf{y}$ is used as the optimization target, it is not manually annotated but directly obtained from the future segment of the same unlabeled sequence. Following common practice in time series analysis \cite{Zhang2024SelfSupervised}, we therefore regard this forecasting objective as self-supervised.}

\section{Benchmark}
\label{section:Benchmark}

\begin{table}[t]
	\centering
	\caption{The statistics of each sub-dataset. A lower ADF test statistic denotes more stationary series.}
	\resizebox{0.85\linewidth}{!}
	{
		\begin{tabular}{@{}c|c@{\hspace{24pt}}c@{\hspace{24pt}}c@{\hspace{24pt}}c@{\hspace{24pt}}l@{}}
	\toprule
	Domain  & Number & Period                     & Frequency  & ADF   & \multicolumn{1}{c}{Information} \\ \midrule
	Energy  & 100    & 2025-01-01$\sim$2025-01-31 & 30 minutes & -8.08 & photovoltaic generation         \\
	Traffic & 100    & 2025-01-01$\sim$2025-01-31 & 60 minutes & -5.68 & traffic flow                    \\
	Weather & 100    & 2025-01-01$\sim$2025-01-31 & 60 minutes & -3.77 & particulate matter              \\ \bottomrule
\end{tabular}
	}
	\label{tab1}
\end{table}

In contrast to the abundant benchmark datasets available in domains such as text-image \cite{YuYLWL0WW24, LiuDZLZZYWHLCL24} and text-video \cite{XuMYR16, KrishnaHRFN17}, the development of benchmark datasets for MTSF remains comparatively underdeveloped. Considering the ongoing expansion of corpus size required for LLMs, existing benchmarks \cite{Terra, TimeMMD, CiK, ChatTime} may have already been included in their training data. This overlap leads to information leakage during evaluation, undermining the reliability of experiment results. To address this concern, we adhere to the established temporal-update paradigm \cite{ChatTime} to construct MTSFBench-300 as a leakage-controlled benchmark for MTSF. This benchmark encompasses three key domains (Energy, Traffic, and Weather) and incorporates data updated to January 2025. MTSFBench-300 is designed as a leakage-controlled benchmark rather than a large-scale benchmark, and its 300 samples, with 100 samples per domain, limit the statistical power and external generalizability of the evaluation. We therefore interpret the results together with the experiments under different historical and prediction lengths, the repeated runs, and the statistical significance tests reported in Section \ref{subsection:Main_Result}. Detailed statistical characteristics are provided in Table \ref{tab1}.

The Energy dataset is constructed using the UK PV generation system. The raw data consists of 317 GSPs, from which seven faulty units are excluded after a data integrity assessment. To ensure spatial representativeness, 100 GSPs are selected through clustering based on geographic coordinates. The PV generation series is recorded at a 30-minute resolution and includes contextual information such as background, city, forecast-only weather (to prevent future information leakage), and date. Time series slices are generated using a 14-day sliding window (7 days of historical data and 7 days for prediction) with a one-day stride. Finally, 100 representative samples are retained by clustering based on multidimensional characteristics, including geospatial information, weather conditions, and holidays. The dataset exhibits substantial diurnal variability, with non-zero values occurring exclusively during daylight hours. Appendix \ref{subappendix:Prediction_Showcase_on_Energy} provides corresponding text prompts and visual representations.

The Traffic dataset is derived from the highway sensor network in California. Following LargeST \cite{LiuXL0W00LHZ23}, 8,600 valid nodes are selected from an initial pool of 18,954 sensors. After integrating POIs, spatial clustering is performed based on geographic coordinates and POI information, yielding 100 representative nodes. Hourly traffic flow series is collected and supplemented with contextual information, including background, city, forecast-only weather, and date. Time series slices are generated using a 14-day sliding window (7 days of historical data and 7 days for prediction) with a one-day stride. Subsequently, 100 representative samples are retained through multidimensional clustering, considering factors such as geospatial information, POIs, weather conditions, and holidays. The dataset exhibits a relatively stable distribution. Appendix \ref{subappendix:Prediction_Showcase_on_Traffic} provides corresponding text prompts and visual representations.

The Weather dataset focuses on monitoring PM10 concentrations in major global cities. From a total of 47,868 cities, 771 core cities with populations exceeding one million are selected. With clustering sampling based on geographic coordinates, climate types, and altitude, 100 representative cities are chosen for further analysis. Hourly PM10 concentration series is collected and enriched with contextual information, including background, forecast-only weather, and date. Time series slices are generated using a 14-day sliding window (7 days of historical data and 7 days for prediction) with a one-day stride. Finally, 100 representative samples are selected through clustering, considering multidimensional attributes such as geospatial information, climate, elevation, weather conditions, and holidays. The dataset demonstrates significant non-stationarity. Appendix \ref{subappendix:Prediction_Showcase_on_Weather} provides corresponding text prompts and visual representations.

\section{Experiment}
\label{section:Experiment}

\subsection{Experimental Setup}
\label{subsection:Experimental_Setup}

\subsubsection{Dataset}
\label{subsubsection:Dataset}

The synthetic cross-modal dataset from Section \ref{section:Methodology} is used to pre-train and fine-tune ChronoSteer, and is split into training and validation sets at an 8:2 ratio to reduce overfitting and support hyperparameter selection. Once fine-tuning is complete, the weights of ChronoSteer are frozen to enable zero-shot evaluation on benchmark. We evaluate on MTSFBench-300, which is constructed after the evaluated checkpoints are fixed and uses only context available before the prediction window during evaluation. This benchmark encompasses three key domains (Energy, Traffic, and Weather) with different frequencies and incorporates data updated to January 2025. We conduct evaluation experiments on MTSFBench-300, adhering to established zero-shot protocols \cite{ChatTime}. Specifically, the prediction length is fixed at 1 day, while historical lengths vary across \{2, 3, 4, 7\} days. The historical window is maintained larger than the prediction window to ensure sufficient information for zero-shot forecasting. Further details about benchmark are provided in Section \ref{section:Benchmark}.

\subsubsection{Baseline}
\label{subsubsection:Baseline}

Apart from ChatTime, the development of multimodal TSFMs capable of zero-shot inference by jointly leveraging textual and temporal information remains unexplored. To more comprehensively evaluate the effectiveness of ChronoSteer, we additionally include several baselines capable of zero-shot forecasting. Specifically, we consider Chronos-Bolt \cite{Chronos}, TimesFM-2.0 \cite{TimesFM}, and Moirai-1.1 \cite{Moirai} as baselines for unimodal forecasting, and DeepSeek-R1 \cite{DeepSeekR1}, QwQ-32B \cite{QwQ32B}, and ChatTime-7B \cite{ChatTime} as baselines for multimodal forecasting. Although these LLMs are not specifically designed for time series, other end-to-end methods \cite{PatchTST, TimeLLM, TGForecaster, SELLM} are excluded from the comparison due to their lack of zero-shot capabilities. Due to constraints on computing resources, evaluations of the LLMs are conducted via their official APIs, while the remaining baselines are tested on our identical hardware. For DeepSeek-R1 and QwQ-32B, the APIs are used only as inference endpoints for their released open-source or open-weight checkpoints. Details are as follows:

\begin{itemize}

	\item DeepSeek-R1: The first large-scale reasoning model, with over 100 billion parameters, specializes in mathematics, programming, and logical reasoning tasks. Released in January 2025, it is post-trained via reinforcement learning based on DeepSeek-V3-Base (released in December 2024). \textcolor{black}{The reinforcement learning data of DeepSeek-R1 mainly come from mathematical competitions \cite{DeepSeekR1}, with a knowledge cutoff of July 2024.} We use the API provided on its official website, with all parameters set to default values (\textit{top\_p=1.0, temperature=1.0}).

	\item QwQ-32B: A reinforcement learning-based model performs comparable to DeepSeek-R1 with 671B parameters in mathematical and programming tasks, using only 32B parameters. Released in March 2025, it is post-trained via reinforcement learning based on Qwen2.5-32B (released in September 2024). \textcolor{black}{The reinforcement learning of QwQ-32B focuses on mathematical and coding capabilities \cite{QwQ32B}, with a knowledge cutoff of November 2024.} We use the API provided on its official website, with all parameters set to default values (\textit{top\_k=40, top\_p=0.8, temperature=0.7}).

	\item ChatTime-7B: A multimodal foundation model tailored for time series analysis, supports dual-modal input and output of series and text. Released in December 2024, it is fine-tuned via LoRA \cite{LoRA} based on LLaMA-2-7B (released in July 2023). We use the weight provided on its official website, with all parameters set to default values (\textit{num\_samples=8, top\_k=100, top\_p=1.0, temperature=1.0}).

	\item Chronos-Bolt: A pretrained time series forecasting model based on the T5 encoder-decoder architecture, trained on nearly 100 billion time series observations. Released in November 2024. We use the weight provided on its official website, with all parameters set to default values.

	\item TimesFM-2.0: A pretrained time series foundation model developed by Google Research for time series forecasting, achieving up to 25\% improvement over v1.0 on leading benchmarks. Released in December 2024. We use the weight provided on its official website, with all parameters set to default values.

	\item Moirai-1.1: A masked encoder-based universal time series forecasting transformer pretrained on the LOTSA dataset, which demonstrates significant improvements over v1.0 in low-frequency scenarios. Released in June 2024. We use the weight provided on its official website, with all parameters set to default values (\textit{num\_samples=100}).

\end{itemize}

\textcolor{black}{MTSFBench-300 contains target observations from January 2025, all of which were already available within that month. The benchmark was assembled from these data in March 2025. ChatTime-7B and the three TSFM baselines were released before this period, while the reported knowledge cutoffs for DeepSeek-R1 and QwQ-32B both fall in 2024. These temporal boundaries rule out any overlap between the benchmark targets and the training data of the evaluated models. Moreover, each evaluation sample includes only textual context available before the prediction window and does not reveal any future observations.}

\subsubsection{Implementation}
\label{subsubsection:Implementation}

Given its widespread adoption and standardized architecture, we choose Chronos-Bolt-base \cite{Chronos} as the unimodal TSFM backbone in our experiments, with an embedding dimension of 768. The text embedding model is set to BGE-M3 \cite{BgeM3}, whose output embeddings have a dimensionality of 1024. Textual events are transformed into revision instructions using the LLM QwQ-32B \cite{QwQ32B}, with all parameters set to default values (\textit{top\_k=40, top\_p=0.8, temperature=0.7}). The prompts for generating these textual revision instructions and their corresponding answers are detailed in Appendix \ref{appendix:Detailed_Prediction_Showcase}. Additionally, ablation studies on the choice of LLM, text embedding model, and unimodal TSFM backbone are provided in Appendix \ref{appendix:Detailed_Ablation_Study}, offering further insights into the impact of each module.

During both the pre-training and fine-tuning stages of ChronoSteer, the Adam optimizer is consistently employed. The model is trained for up to 100 epochs, with early stopping triggered if the validation loss does not decrease over 10 consecutive epochs, mitigating overfitting. In the pre-training stage, which involves contrastive learning tasks with multiple anchor revision instructions, a batch size of 32 is used. During fine-tuning, the batch size is increased to 256. Through grid search, the optimal hyperparameters for ChronoSteer are determined: an initial learning rate of 0.001 (selected from \{0.0001, 0.0005, 0.001, 0.005\}), a hidden layer dimension of 1024 in the alignment module (selected from \{512, 1024, 2048, 4096\}), and a contrastive loss weight of 0.001 during pre-training (selected from \{0.0001, 0.001, 0.01, 0.1\}). Section \ref{subsection:Hyperparameter_Analysis} provides a detailed analysis of each hyperparameter.

All experiments are conducted using Python 3.10.13 and PyTorch 2.1.2. The prediction accuracy is evaluated using mean squared error (MSE) and mean absolute error (MAE), where lower values indicate better model performance. To ensure the statistical representativeness of our results, all experiments are repeated across ten different random seeds, and we report both the mean and standard deviation. Notably, TimesFM and Chronos exhibit deterministic behavior during inference, yielding a standard deviation of zero. While ChronoSteer inherits this determinism from Chronos, we seek to comprehensively evaluate its robustness. Therefore, unlike the baselines which are evaluated solely in inference mode, we conduct the ChronoSteer experiments by re-initiating the process from the pre-trained stage for each of the ten repetitions. To facilitate open science and community collaboration, we will make all data, source code, and pre-trained weight checkpoints publicly accessible following the release. These resources will offer substantial support for subsequent research.

\subsection{Main Result}
\label{subsection:Main_Result}

\begin{table}[t]
	\centering
	\caption{The prediction errors of ChronoSteer and baselines across various historical windows. The historical windows vary across \{2, 3, 4, 7\} days, while the prediction window is fixed at 1 day. A lower value denotes superior accuracy. The optimal and suboptimal results are highlighted in bold and underlined, respectively. $\dagger$ indicates that ChronoSteer Multimodal significantly outperforms the strongest non-ChronoSteer baseline in the same row under a two-sided paired Wilcoxon signed-rank test with Holm-Bonferroni correction over all primary comparisons ($p<0.05$).}
	\resizebox{\linewidth}{!}
	{
		\begin{tabular}{@{}ccc|c|c|c@{\hspace{3pt}}c|c@{\hspace{3pt}}c|c@{\hspace{3pt}}c|c@{\hspace{3pt}}c@{}}
	\toprule
	\multicolumn{3}{c|}{\multirow{2}{*}{Method}}                  & TimesFM-2.0                              & Moirai-1.1 & \multicolumn{2}{c|}{DeepSeek-R1} & \multicolumn{2}{c|}{QwQ-32B} & \multicolumn{2}{c|}{ChatTime-7B} & \multicolumn{2}{c}{ChronoSteer}                                                                                                                                                             \\ \cmidrule(l){4-13}
	\multicolumn{3}{c|}{}                                         & Unimodal                                 & Unimodal   & Unimodal                         & Multimodal                   & Unimodal                         & Multimodal                      & Unimodal        & Multimodal                   & Unimodal        & Multimodal                                                                             \\ \midrule
	\multicolumn{1}{c|}{\multirow{8}{*}{\rotatebox{90}{Energy}}}  & \multicolumn{1}{c|}{\multirow{2}{*}{2D}} & MSE        & 0.728$\pm$0.000                  & 0.754$\pm$0.011              & 1.615$\pm$0.070                  & \underline{ 0.711$\pm$0.037}    & 3.766$\pm$0.167 & 0.733$\pm$0.037              & 1.250$\pm$0.012 & 0.820$\pm$0.013              & 0.765$\pm$0.000              & \textbf{0.573$\pm$0.012}$^{\dagger}$ \\
	\multicolumn{1}{c|}{}                                         & \multicolumn{1}{c|}{}                    & MAE        & 0.288$\pm$0.000                  & 0.294$\pm$0.003              & 0.411$\pm$0.021                  & \underline{ 0.274$\pm$0.016}    & 0.649$\pm$0.043 & 0.279$\pm$0.018              & 0.368$\pm$0.005 & 0.295$\pm$0.003              & 0.277$\pm$0.000              & \textbf{0.265$\pm$0.007}$^{\dagger}$ \\ \cmidrule(l){2-13}
	\multicolumn{1}{c|}{}                                         & \multicolumn{1}{c|}{\multirow{2}{*}{3D}} & MSE        & 0.695$\pm$0.000                  & 0.716$\pm$0.012              & 0.982$\pm$0.045                  & 0.690$\pm$0.043                 & 2.422$\pm$0.099 & \underline{ 0.672$\pm$0.028} & 1.254$\pm$0.011 & 0.823$\pm$0.010              & 0.763$\pm$0.000              & \textbf{0.591$\pm$0.008}$^{\dagger}$ \\
	\multicolumn{1}{c|}{}                                         & \multicolumn{1}{c|}{}                    & MAE        & 0.320$\pm$0.000                  & 0.281$\pm$0.005              & 0.342$\pm$0.009                  & 0.271$\pm$0.018                 & 0.636$\pm$0.022 & \underline{ 0.267$\pm$0.022} & 0.360$\pm$0.005 & 0.297$\pm$0.003              & 0.268$\pm$0.000              & \textbf{0.267$\pm$0.004} \\ \cmidrule(l){2-13}
	\multicolumn{1}{c|}{}                                         & \multicolumn{1}{c|}{\multirow{2}{*}{4D}} & MSE        & 0.688$\pm$0.000                  & 0.727$\pm$0.014              & 0.892$\pm$0.057                  & 0.650$\pm$0.054                 & 1.404$\pm$0.074 & \underline{ 0.648$\pm$0.046} & 1.248$\pm$0.018 & 0.825$\pm$0.010              & 0.697$\pm$0.000              & \textbf{0.613$\pm$0.011}$^{\dagger}$ \\
	\multicolumn{1}{c|}{}                                         & \multicolumn{1}{c|}{}                    & MAE        & 0.279$\pm$0.000                  & 0.284$\pm$0.004              & 0.301$\pm$0.019                  & 0.263$\pm$0.012                 & 0.419$\pm$0.020 & 0.270$\pm$0.015              & 0.365$\pm$0.004 & 0.294$\pm$0.004              & \underline{ 0.259$\pm$0.000} & \textbf{0.253$\pm$0.005}$^{\dagger}$ \\ \cmidrule(l){2-13}
	\multicolumn{1}{c|}{}                                         & \multicolumn{1}{c|}{\multirow{2}{*}{7D}} & MSE        & 0.581$\pm$0.000                  & 0.620$\pm$0.010              & 0.735$\pm$0.010                  & 0.663$\pm$0.034                 & 1.340$\pm$0.101 & 0.618$\pm$0.038              & 1.250$\pm$0.022 & 0.817$\pm$0.010              & \underline{ 0.573$\pm$0.000} & \textbf{0.524$\pm$0.006}$^{\dagger}$ \\
	\multicolumn{1}{c|}{}                                         & \multicolumn{1}{c|}{}                    & MAE        & 0.288$\pm$0.000                  & 0.259$\pm$0.003              & 0.284$\pm$0.013                  & 0.251$\pm$0.018                 & 0.423$\pm$0.018 & 0.266$\pm$0.019              & 0.363$\pm$0.005 & 0.295$\pm$0.005              & \underline{ 0.246$\pm$0.000} & \textbf{0.243$\pm$0.005}$^{\dagger}$ \\ \midrule
	\multicolumn{1}{c|}{\multirow{8}{*}{\rotatebox{90}{Traffic}}} & \multicolumn{1}{c|}{\multirow{2}{*}{2D}} & MSE        & 0.681$\pm$0.000                  & 0.853$\pm$0.013              & 1.008$\pm$0.074                  & 0.646$\pm$0.036                 & 1.660$\pm$0.059 & 0.638$\pm$0.029              & 0.932$\pm$0.008 & \underline{ 0.600$\pm$0.006} & 0.669$\pm$0.000              & \textbf{0.512$\pm$0.005}$^{\dagger}$ \\
	\multicolumn{1}{c|}{}                                         & \multicolumn{1}{c|}{}                    & MAE        & 0.568$\pm$0.000                  & 0.655$\pm$0.008              & 0.633$\pm$0.019                  & \underline{ 0.544$\pm$0.026}    & 0.925$\pm$0.033 & 0.577$\pm$0.037              & 0.681$\pm$0.006 & 0.551$\pm$0.009              & 0.571$\pm$0.000              & \textbf{0.508$\pm$0.013}$^{\dagger}$ \\ \cmidrule(l){2-13}
	\multicolumn{1}{c|}{}                                         & \multicolumn{1}{c|}{\multirow{2}{*}{3D}} & MSE        & 0.544$\pm$0.000                  & 0.655$\pm$0.007              & 0.636$\pm$0.028                  & 0.587$\pm$0.027                 & 1.431$\pm$0.093 & \underline{ 0.530$\pm$0.042} & 0.816$\pm$0.009 & 0.581$\pm$0.010              & 0.608$\pm$0.000              & \textbf{0.401$\pm$0.007}$^{\dagger}$ \\
	\multicolumn{1}{c|}{}                                         & \multicolumn{1}{c|}{}                    & MAE        & 0.516$\pm$0.000                  & 0.569$\pm$0.010              & 0.554$\pm$0.032                  & \underline{ 0.507$\pm$0.028}    & 0.902$\pm$0.030 & 0.509$\pm$0.038              & 0.591$\pm$0.008 & 0.536$\pm$0.006              & 0.551$\pm$0.000              & \textbf{0.448$\pm$0.006}$^{\dagger}$ \\ \cmidrule(l){2-13}
	\multicolumn{1}{c|}{}                                         & \multicolumn{1}{c|}{\multirow{2}{*}{4D}} & MSE        & 0.526$\pm$0.000                  & 0.522$\pm$0.007              & 0.544$\pm$0.022                  & 0.511$\pm$0.037                 & 1.362$\pm$0.073 & \underline{ 0.437$\pm$0.020} & 0.666$\pm$0.007 & 0.449$\pm$0.005              & 0.559$\pm$0.000              & \textbf{0.367$\pm$0.006}$^{\dagger}$ \\
	\multicolumn{1}{c|}{}                                         & \multicolumn{1}{c|}{}                    & MAE        & 0.492$\pm$0.000                  & 0.482$\pm$0.007              & 0.538$\pm$0.019                  & 0.509$\pm$0.029                 & 0.873$\pm$0.027 & 0.465$\pm$0.029              & 0.579$\pm$0.010 & \underline{ 0.460$\pm$0.007} & 0.522$\pm$0.000              & \textbf{0.430$\pm$0.007}$^{\dagger}$ \\ \cmidrule(l){2-13}
	\multicolumn{1}{c|}{}                                         & \multicolumn{1}{c|}{\multirow{2}{*}{7D}} & MSE        & \underline{ 0.294$\pm$0.000}     & 0.421$\pm$0.005              & 0.446$\pm$0.021                  & 0.450$\pm$0.028                 & 1.049$\pm$0.050 & 0.379$\pm$0.028              & 0.555$\pm$0.007 & 0.301$\pm$0.005              & 0.344$\pm$0.000              & \textbf{0.242$\pm$0.004}$^{\dagger}$ \\
	\multicolumn{1}{c|}{}                                         & \multicolumn{1}{c|}{}                    & MAE        & 0.361$\pm$0.000                  & 0.459$\pm$0.007              & 0.470$\pm$0.033                  & 0.475$\pm$0.029                 & 0.807$\pm$0.035 & 0.423$\pm$0.024              & 0.524$\pm$0.012 & \underline{ 0.351$\pm$0.004} & 0.390$\pm$0.000              & \textbf{0.332$\pm$0.006}$^{\dagger}$ \\ \midrule
	\multicolumn{1}{c|}{\multirow{8}{*}{\rotatebox{90}{Weather}}} & \multicolumn{1}{c|}{\multirow{2}{*}{2D}} & MSE        & 1.565$\pm$0.000                  & 2.536$\pm$0.051              & 3.502$\pm$0.177                  & \underline{ 1.537$\pm$0.088}    & 2.391$\pm$0.098 & 1.768$\pm$0.101              & 2.737$\pm$0.035 & 2.742$\pm$0.024              & 2.340$\pm$0.000              & \textbf{1.532$\pm$0.029}$^{\dagger}$ \\
	\multicolumn{1}{c|}{}                                         & \multicolumn{1}{c|}{}                    & MAE        & 0.973$\pm$0.000                  & 1.196$\pm$0.015              & 1.361$\pm$0.058                  & \underline{ 0.925$\pm$0.063}    & 1.194$\pm$0.056 & 1.110$\pm$0.093              & 1.262$\pm$0.009 & 1.327$\pm$0.022              & 1.126$\pm$0.000              & \textbf{0.910$\pm$0.019}$^{\dagger}$ \\ \cmidrule(l){2-13}
	\multicolumn{1}{c|}{}                                         & \multicolumn{1}{c|}{\multirow{2}{*}{3D}} & MSE        & \underline{ 1.307$\pm$0.000}     & 1.996$\pm$0.027              & 3.038$\pm$0.133                  & 1.500$\pm$0.147                 & 1.948$\pm$0.076 & 1.770$\pm$0.133              & 2.685$\pm$0.041 & 2.572$\pm$0.041              & 1.511$\pm$0.000              & \textbf{0.898$\pm$0.015}$^{\dagger}$ \\
	\multicolumn{1}{c|}{}                                         & \multicolumn{1}{c|}{}                    & MAE        & 0.886$\pm$0.000                  & 1.081$\pm$0.016              & 1.302$\pm$0.050                  & \underline{ 0.885$\pm$0.057}    & 1.121$\pm$0.051 & 1.028$\pm$0.073              & 1.301$\pm$0.015 & 1.290$\pm$0.020              & 0.967$\pm$0.000              & \textbf{0.726$\pm$0.014}$^{\dagger}$ \\ \cmidrule(l){2-13}
	\multicolumn{1}{c|}{}                                         & \multicolumn{1}{c|}{\multirow{2}{*}{4D}} & MSE        & \underline{ 1.157$\pm$0.000}     & 1.774$\pm$0.010              & 2.687$\pm$0.121                  & 1.467$\pm$0.079                 & 1.944$\pm$0.119 & 1.680$\pm$0.112              & 2.082$\pm$0.024 & 2.123$\pm$0.028              & 1.358$\pm$0.000              & \textbf{0.800$\pm$0.014}$^{\dagger}$ \\
	\multicolumn{1}{c|}{}                                         & \multicolumn{1}{c|}{}                    & MAE        & \underline{ 0.836$\pm$0.000}     & 1.009$\pm$0.013              & 1.260$\pm$0.057                  & 0.893$\pm$0.059                 & 1.043$\pm$0.045 & 1.020$\pm$0.071              & 1.068$\pm$0.015 & 1.085$\pm$0.011              & 0.909$\pm$0.000              & \textbf{0.673$\pm$0.008}$^{\dagger}$ \\ \cmidrule(l){2-13}
	\multicolumn{1}{c|}{}                                         & \multicolumn{1}{c|}{\multirow{2}{*}{7D}} & MSE        & \underline{ 1.065$\pm$0.000}     & 1.416$\pm$0.018              & 2.318$\pm$0.129                  & 1.371$\pm$0.110                 & 1.844$\pm$0.076 & 1.656$\pm$0.098              & 1.641$\pm$0.029 & 1.448$\pm$0.025              & 1.116$\pm$0.000              & \textbf{0.754$\pm$0.021}$^{\dagger}$ \\
	\multicolumn{1}{c|}{}                                         & \multicolumn{1}{c|}{}                    & MAE        & \underline{ 0.804$\pm$0.000}     & 0.897$\pm$0.020              & 1.095$\pm$0.038                  & 0.832$\pm$0.047                 & 1.082$\pm$0.037 & 0.959$\pm$0.080              & 0.933$\pm$0.018 & 0.896$\pm$0.015              & 0.822$\pm$0.000              & \textbf{0.646$\pm$0.012}$^{\dagger}$ \\ \bottomrule
\end{tabular}
	}
	\label{tab2}
\end{table}

\begin{table}[t]
	\centering
	\caption{The prediction errors of ChronoSteer and baselines across various prediction windows. The historical window is fixed at 7 days, while the prediction windows vary across \{6 hours, 12 hours, 2 days, 4 days\}. A lower value denotes superior accuracy. The optimal and suboptimal results are highlighted in bold and underlined, respectively. $\dagger$ indicates that ChronoSteer Multimodal significantly outperforms the strongest non-ChronoSteer baseline in the same row under a two-sided paired Wilcoxon signed-rank test with Holm-Bonferroni correction over all primary comparisons ($p<0.05$).}
	\resizebox{\linewidth}{!}
	{
		\begin{tabular}{@{}ccc|c|c|c@{\hspace{3pt}}c|c@{\hspace{3pt}}c|c@{\hspace{3pt}}c|c@{\hspace{3pt}}c@{}}
	\toprule
	\multicolumn{3}{c|}{\multirow{2}{*}{Method}}                  & TimesFM-2.0                               & Moirai-1.1 & \multicolumn{2}{c|}{DeepSeek-R1} & \multicolumn{2}{c|}{QwQ-32B} & \multicolumn{2}{c|}{ChatTime-7B} & \multicolumn{2}{c}{ChronoSteer}                                                                                                                                   \\ \cmidrule(l){4-13}
	\multicolumn{3}{c|}{}                                         & Unimodal                                  & Unimodal   & Unimodal                         & Multimodal                   & Unimodal                         & Multimodal                      & Unimodal        & Multimodal      & Unimodal        & Multimodal                                                                \\ \midrule
	\multicolumn{1}{c|}{\multirow{8}{*}{\rotatebox{90}{Energy}}}  & \multicolumn{1}{c|}{\multirow{2}{*}{6H}}  & MSE        & 0.037$\pm$0.000                  & 0.038$\pm$0.001              & 0.067$\pm$0.003                  & 0.055$\pm$0.003                 & 0.090$\pm$0.004 & 0.048$\pm$0.003 & 0.066$\pm$0.001 & 0.062$\pm$0.001 & \underline{ 0.037$\pm$0.000} & \textbf{0.019$\pm$0.000} \\
	\multicolumn{1}{c|}{}                                         & \multicolumn{1}{c|}{}                     & MAE        & \underline{ 0.078$\pm$0.000}     & 0.080$\pm$0.001              & 0.129$\pm$0.010                  & 0.091$\pm$0.054                 & 0.145$\pm$0.012 & 0.090$\pm$0.005 & 0.111$\pm$0.002 & 0.093$\pm$0.002 & 0.079$\pm$0.000              & \textbf{0.063$\pm$0.001} \\ \cmidrule(l){2-13}
	\multicolumn{1}{c|}{}                                         & \multicolumn{1}{c|}{\multirow{2}{*}{12H}} & MSE        & 0.312$\pm$0.000                  & 0.403$\pm$0.005              & 0.507$\pm$0.031                  & 0.440$\pm$0.034                 & 0.614$\pm$0.047 & 0.585$\pm$0.050 & 0.488$\pm$0.010 & 0.419$\pm$0.006 & \underline{ 0.281$\pm$0.000} & \textbf{0.232$\pm$0.005}$^{\dagger}$ \\
	\multicolumn{1}{c|}{}                                         & \multicolumn{1}{c|}{}                     & MAE        & 0.203$\pm$0.000                  & 0.175$\pm$0.004              & 0.267$\pm$0.029                  & 0.251$\pm$0.016                 & 0.404$\pm$0.017 & 0.323$\pm$0.034 & 0.313$\pm$0.004 & 0.271$\pm$0.003 & \underline{ 0.148$\pm$0.000} & \textbf{0.117$\pm$0.001}$^{\dagger}$ \\ \cmidrule(l){2-13}
	\multicolumn{1}{c|}{}                                         & \multicolumn{1}{c|}{\multirow{2}{*}{2D}}  & MSE        & 0.644$\pm$0.000                  & 0.681$\pm$0.014              & 1.050$\pm$0.090                  & 0.679$\pm$0.035                 & 1.378$\pm$0.068 & 0.709$\pm$0.048 & 1.293$\pm$0.020 & 0.861$\pm$0.015 & \underline{ 0.583$\pm$0.000} & \textbf{0.527$\pm$0.007}$^{\dagger}$ \\
	\multicolumn{1}{c|}{}                                         & \multicolumn{1}{c|}{}                     & MAE        & 0.289$\pm$0.000                  & 0.246$\pm$0.003              & 0.508$\pm$0.036                  & 0.311$\pm$0.024                 & 0.583$\pm$0.027 & 0.308$\pm$0.009 & 0.547$\pm$0.006 & 0.404$\pm$0.007 & \underline{ 0.236$\pm$0.000} & \textbf{0.176$\pm$0.004}$^{\dagger}$ \\ \cmidrule(l){2-13}
	\multicolumn{1}{c|}{}                                         & \multicolumn{1}{c|}{\multirow{2}{*}{4D}}  & MSE        & 0.667$\pm$0.000                  & 0.804$\pm$0.012              & 1.098$\pm$0.083                  & 0.730$\pm$0.053                 & 1.411$\pm$0.064 & 0.718$\pm$0.038 & 1.326$\pm$0.027 & 0.879$\pm$0.016 & \underline{ 0.597$\pm$0.000} & \textbf{0.530$\pm$0.009}$^{\dagger}$ \\
	\multicolumn{1}{c|}{}                                         & \multicolumn{1}{c|}{}                     & MAE        & 0.319$\pm$0.000                  & 0.280$\pm$0.003              & 0.519$\pm$0.029                  & 0.342$\pm$0.024                 & 0.617$\pm$0.028 & 0.356$\pm$0.024 & 0.590$\pm$0.011 & 0.420$\pm$0.007 & \underline{ 0.237$\pm$0.000} & \textbf{0.188$\pm$0.004}$^{\dagger}$ \\ \midrule
	\multicolumn{1}{c|}{\multirow{8}{*}{\rotatebox{90}{Traffic}}} & \multicolumn{1}{c|}{\multirow{2}{*}{6H}}  & MSE        & 0.180$\pm$0.000                  & 0.237$\pm$0.003              & 0.284$\pm$0.025                  & 0.250$\pm$0.018                 & 0.349$\pm$0.026 & 0.295$\pm$0.018 & 0.271$\pm$0.004 & 0.255$\pm$0.005 & \underline{ 0.177$\pm$0.000} & \textbf{0.143$\pm$0.002}$^{\dagger}$ \\
	\multicolumn{1}{c|}{}                                         & \multicolumn{1}{c|}{}                     & MAE        & \underline{ 0.210$\pm$0.000}     & 0.313$\pm$0.004              & 0.324$\pm$0.023                  & 0.285$\pm$0.023                 & 0.398$\pm$0.019 & 0.333$\pm$0.027 & 0.345$\pm$0.003 & 0.276$\pm$0.004 & 0.211$\pm$0.000              & \textbf{0.178$\pm$0.005}$^{\dagger}$ \\ \cmidrule(l){2-13}
	\multicolumn{1}{c|}{}                                         & \multicolumn{1}{c|}{\multirow{2}{*}{12H}} & MSE        & \underline{ 0.192$\pm$0.000}     & 0.380$\pm$0.005              & 0.297$\pm$0.027                  & 0.264$\pm$0.020                 & 0.370$\pm$0.014 & 0.326$\pm$0.016 & 0.316$\pm$0.003 & 0.252$\pm$0.002 & 0.225$\pm$0.000              & \textbf{0.187$\pm$0.003} \\
	\multicolumn{1}{c|}{}                                         & \multicolumn{1}{c|}{}                     & MAE        & \underline{ 0.263$\pm$0.000}     & 0.410$\pm$0.006              & 0.418$\pm$0.023                  & 0.375$\pm$0.027                 & 0.521$\pm$0.020 & 0.409$\pm$0.032 & 0.411$\pm$0.007 & 0.344$\pm$0.008 & 0.278$\pm$0.000              & \textbf{0.228$\pm$0.003}$^{\dagger}$ \\ \cmidrule(l){2-13}
	\multicolumn{1}{c|}{}                                         & \multicolumn{1}{c|}{\multirow{2}{*}{2D}}  & MSE        & \underline{ 0.380$\pm$0.000}     & 0.523$\pm$0.007              & 0.626$\pm$0.048                  & 0.455$\pm$0.028                 & 1.334$\pm$0.032 & 0.420$\pm$0.022 & 0.581$\pm$0.008 & 0.500$\pm$0.008 & 0.430$\pm$0.000              & \textbf{0.303$\pm$0.006}$^{\dagger}$ \\
	\multicolumn{1}{c|}{}                                         & \multicolumn{1}{c|}{}                     & MAE        & \underline{ 0.390$\pm$0.000}     & 0.468$\pm$0.003              & 0.662$\pm$0.059                  & 0.450$\pm$0.014                 & 0.974$\pm$0.034 & 0.443$\pm$0.048 & 0.591$\pm$0.012 & 0.522$\pm$0.008 & 0.426$\pm$0.000              & \textbf{0.343$\pm$0.006}$^{\dagger}$ \\ \cmidrule(l){2-13}
	\multicolumn{1}{c|}{}                                         & \multicolumn{1}{c|}{\multirow{2}{*}{4D}}  & MSE        & \underline{ 0.399$\pm$0.000}     & 0.559$\pm$0.009              & 0.651$\pm$0.035                  & 0.488$\pm$0.037                 & 1.773$\pm$0.042 & 0.456$\pm$0.044 & 0.635$\pm$0.009 & 0.541$\pm$0.012 & 0.455$\pm$0.000              & \textbf{0.347$\pm$0.008}$^{\dagger}$ \\
	\multicolumn{1}{c|}{}                                         & \multicolumn{1}{c|}{}                     & MAE        & \underline{ 0.378$\pm$0.000}     & 0.502$\pm$0.007              & 0.644$\pm$0.032                  & 0.480$\pm$0.029                 & 1.271$\pm$0.045 & 0.440$\pm$0.031 & 0.581$\pm$0.013 & 0.506$\pm$0.008 & 0.416$\pm$0.000              & \textbf{0.352$\pm$0.007}$^{\dagger}$ \\ \midrule
	\multicolumn{1}{c|}{\multirow{8}{*}{\rotatebox{90}{Weather}}} & \multicolumn{1}{c|}{\multirow{2}{*}{6H}}  & MSE        & \underline{ 0.394$\pm$0.000}     & 0.647$\pm$0.010              & 0.676$\pm$0.027                  & 0.499$\pm$0.032                 & 0.798$\pm$0.038 & 0.581$\pm$0.036 & 0.616$\pm$0.012 & 0.518$\pm$0.008 & 0.512$\pm$0.000              & \textbf{0.361$\pm$0.006}$^{\dagger}$ \\
	\multicolumn{1}{c|}{}                                         & \multicolumn{1}{c|}{}                     & MAE        & \underline{ 0.449$\pm$0.000}     & 0.554$\pm$0.009              & 0.738$\pm$0.036                  & 0.591$\pm$0.042                 & 0.881$\pm$0.051 & 0.591$\pm$0.056 & 0.712$\pm$0.013 & 0.591$\pm$0.009 & 0.502$\pm$0.000              & \textbf{0.419$\pm$0.006}$^{\dagger}$ \\ \cmidrule(l){2-13}
	\multicolumn{1}{c|}{}                                         & \multicolumn{1}{c|}{\multirow{2}{*}{12H}} & MSE        & \underline{ 0.769$\pm$0.000}     & 1.007$\pm$0.015              & 1.287$\pm$0.051                  & 0.929$\pm$0.091                 & 1.513$\pm$0.100 & 0.974$\pm$0.058 & 1.174$\pm$0.022 & 0.985$\pm$0.018 & 0.882$\pm$0.000              & \textbf{0.583$\pm$0.016}$^{\dagger}$ \\
	\multicolumn{1}{c|}{}                                         & \multicolumn{1}{c|}{}                     & MAE        & \underline{ 0.664$\pm$0.000}     & 0.742$\pm$0.010              & 1.136$\pm$0.059                  & 0.783$\pm$0.071                 & 1.253$\pm$0.072 & 0.838$\pm$0.081 & 1.017$\pm$0.016 & 0.872$\pm$0.015 & 0.710$\pm$0.000              & \textbf{0.540$\pm$0.012}$^{\dagger}$ \\ \cmidrule(l){2-13}
	\multicolumn{1}{c|}{}                                         & \multicolumn{1}{c|}{\multirow{2}{*}{2D}}  & MSE        & \underline{ 1.685$\pm$0.000}     & 2.100$\pm$0.021              & 2.648$\pm$0.111                  & 1.730$\pm$0.176                 & 3.333$\pm$0.156 & 1.901$\pm$0.136 & 2.632$\pm$0.048 & 2.246$\pm$0.035 & 1.772$\pm$0.000              & \textbf{1.319$\pm$0.017}$^{\dagger}$ \\
	\multicolumn{1}{c|}{}                                         & \multicolumn{1}{c|}{}                     & MAE        & \underline{ 0.951$\pm$0.000}     & 1.074$\pm$0.015              & 1.564$\pm$0.098                  & 0.962$\pm$0.079                 & 1.931$\pm$0.118 & 1.152$\pm$0.124 & 1.476$\pm$0.034 & 1.260$\pm$0.021 & 0.983$\pm$0.000              & \textbf{0.760$\pm$0.012}$^{\dagger}$ \\ \cmidrule(l){2-13}
	\multicolumn{1}{c|}{}                                         & \multicolumn{1}{c|}{\multirow{2}{*}{4D}}  & MSE        & \underline{ 2.297$\pm$0.000}     & 2.859$\pm$0.040              & 3.706$\pm$0.219                  & 2.459$\pm$0.264                 & 4.583$\pm$0.322 & 2.573$\pm$0.166 & 3.529$\pm$0.050 & 3.084$\pm$0.032 & 2.539$\pm$0.000              & \textbf{1.886$\pm$0.043}$^{\dagger}$ \\
	\multicolumn{1}{c|}{}                                         & \multicolumn{1}{c|}{}                     & MAE        & \underline{ 1.042$\pm$0.000}     & 1.147$\pm$0.014              & 1.697$\pm$0.145                  & 1.113$\pm$0.038                 & 2.101$\pm$0.089 & 1.186$\pm$0.080 & 1.602$\pm$0.027 & 1.393$\pm$0.024 & 1.090$\pm$0.000              & \textbf{0.853$\pm$0.014}$^{\dagger}$ \\ \bottomrule
\end{tabular}
	}
	\label{tab3}
\end{table}

\begin{table}[t]
	\centering
	\caption{The prediction errors of ChronoSteer and editing variants. The historical windows vary across \{2, 3, 4, 7\} days, while the prediction window is fixed at 1 day. A lower value denotes superior accuracy. The optimal results are highlighted in bold.}
	\resizebox{0.85\linewidth}{!}
	{
		\begin{tabular}{@{}cc|cc|cccccc@{}}
	\toprule
	\multicolumn{2}{c|}{\multirow{3}{*}{Method}}                  & \multicolumn{2}{c|}{Ours}        & \multicolumn{6}{c}{Editing}                                                                                                                                                                                                                 \\ \cmidrule(l){3-10}
	\multicolumn{2}{c|}{}                                         & \multicolumn{2}{c|}{ChronoSteer} & \multicolumn{2}{c|}{DeepSeek-R1} & \multicolumn{2}{c|}{QwQ-32B} & \multicolumn{2}{c}{Function}                                                                                                                                              \\ \cmidrule(l){3-10}
	\multicolumn{2}{c|}{}                                         & MSE                              & MAE                              & MSE                          & \multicolumn{1}{c|}{MAE}     & MSE                                  & \multicolumn{1}{c|}{MAE} & MSE                                  & MAE                               \\ \midrule
	\multicolumn{1}{c|}{\multirow{4}{*}{\rotatebox{90}{Energy}}}  & 2D                               & \textbf{0.573$\pm$0.012}         & \textbf{0.265$\pm$0.007}     & 0.605$\pm$0.045              & \multicolumn{1}{c|}{0.278$\pm$0.013} & 0.591$\pm$0.054          & \multicolumn{1}{c|}{0.266$\pm$0.019} & 0.811$\pm$0.033 & 0.366$\pm$0.011 \\
	\multicolumn{1}{c|}{}                                         & 3D                               & \textbf{0.591$\pm$0.008}         & \textbf{0.267$\pm$0.004}     & 0.672$\pm$0.038              & \multicolumn{1}{c|}{0.272$\pm$0.019} & 0.631$\pm$0.047          & \multicolumn{1}{c|}{0.274$\pm$0.025} & 0.803$\pm$0.017 & 0.365$\pm$0.013 \\
	\multicolumn{1}{c|}{}                                         & 4D                               & \textbf{0.613$\pm$0.011}         & \textbf{0.253$\pm$0.005}     & 0.633$\pm$0.057              & \multicolumn{1}{c|}{0.271$\pm$0.024} & 0.625$\pm$0.041          & \multicolumn{1}{c|}{0.262$\pm$0.020} & 0.788$\pm$0.027 & 0.362$\pm$0.016 \\
	\multicolumn{1}{c|}{}                                         & 7D                               & \textbf{0.524$\pm$0.006}         & \textbf{0.243$\pm$0.005}     & 0.577$\pm$0.043              & \multicolumn{1}{c|}{0.266$\pm$0.016} & 0.564$\pm$0.035          & \multicolumn{1}{c|}{0.256$\pm$0.017} & 0.657$\pm$0.022 & 0.389$\pm$0.009 \\ \midrule
	\multicolumn{1}{c|}{\multirow{4}{*}{\rotatebox{90}{Traffic}}} & 2D                               & \textbf{0.512$\pm$0.005}         & \textbf{0.508$\pm$0.013}     & 0.534$\pm$0.031              & \multicolumn{1}{c|}{0.527$\pm$0.043} & 0.543$\pm$0.035          & \multicolumn{1}{c|}{0.532$\pm$0.034} & 0.648$\pm$0.023 & 0.603$\pm$0.022 \\
	\multicolumn{1}{c|}{}                                         & 3D                               & \textbf{0.401$\pm$0.007}         & \textbf{0.448$\pm$0.006}     & 0.474$\pm$0.021              & \multicolumn{1}{c|}{0.465$\pm$0.020} & 0.467$\pm$0.059          & \multicolumn{1}{c|}{0.466$\pm$0.030} & 0.484$\pm$0.021 & 0.527$\pm$0.017 \\
	\multicolumn{1}{c|}{}                                         & 4D                               & \textbf{0.367$\pm$0.006}         & \textbf{0.430$\pm$0.007}     & 0.409$\pm$0.016              & \multicolumn{1}{c|}{0.433$\pm$0.033} & 0.400$\pm$0.042          & \multicolumn{1}{c|}{0.442$\pm$0.026} & 0.495$\pm$0.019 & 0.547$\pm$0.014 \\
	\multicolumn{1}{c|}{}                                         & 7D                               & \textbf{0.242$\pm$0.004}         & \textbf{0.332$\pm$0.006}     & 0.334$\pm$0.021              & \multicolumn{1}{c|}{0.402$\pm$0.030} & 0.378$\pm$0.016          & \multicolumn{1}{c|}{0.386$\pm$0.040} & 0.462$\pm$0.015 & 0.523$\pm$0.029 \\ \midrule
	\multicolumn{1}{c|}{\multirow{4}{*}{\rotatebox{90}{Weather}}} & 2D                               & \textbf{1.532$\pm$0.029}         & \textbf{0.910$\pm$0.019}     & 1.551$\pm$0.093              & \multicolumn{1}{c|}{0.921$\pm$0.054} & 1.632$\pm$0.122          & \multicolumn{1}{c|}{1.055$\pm$0.044} & 2.125$\pm$0.049 & 1.077$\pm$0.040 \\
	\multicolumn{1}{c|}{}                                         & 3D                               & \textbf{0.898$\pm$0.015}         & \textbf{0.726$\pm$0.014}     & 1.485$\pm$0.065              & \multicolumn{1}{c|}{0.914$\pm$0.086} & 1.620$\pm$0.161          & \multicolumn{1}{c|}{0.983$\pm$0.058} & 1.164$\pm$0.064 & 0.818$\pm$0.022 \\
	\multicolumn{1}{c|}{}                                         & 4D                               & \textbf{0.800$\pm$0.014}         & \textbf{0.673$\pm$0.008}     & 1.309$\pm$0.088              & \multicolumn{1}{c|}{0.870$\pm$0.061} & 1.410$\pm$0.116          & \multicolumn{1}{c|}{0.942$\pm$0.071} & 1.055$\pm$0.038 & 0.786$\pm$0.027 \\
	\multicolumn{1}{c|}{}                                         & 7D                               & \textbf{0.754$\pm$0.021}         & \textbf{0.646$\pm$0.012}     & 0.964$\pm$0.043              & \multicolumn{1}{c|}{0.746$\pm$0.056} & 1.234$\pm$0.069          & \multicolumn{1}{c|}{0.838$\pm$0.072} & 0.868$\pm$0.030 & 0.716$\pm$0.033 \\ \bottomrule
\end{tabular}
	}
	\label{tab4}
\end{table}

\begin{table}[t]
	\centering
	\begin{minipage}{0.47\textwidth}
		\centering
		\caption{The codebook coverage and instruction accuracy of ChronoSteer. The historical windows vary across \{2, 3, 4, 7\} days, while prediction window is fixed at 1. A higher value denotes superior accuracy, except for Optimal MSE.}
		\resizebox{\linewidth}{!}
		{
			\begin{tabular}{@{}cc|cc|cc@{}}
	\toprule
	\multicolumn{2}{c|}{\multirow{2}{*}{Comp.}}                   & \multicolumn{2}{c|}{Codebook} & \multicolumn{2}{c}{LLM}                                                       \\ \cmidrule(l){3-6}
	\multicolumn{2}{c|}{}                                         & Coverage Ratio                & Optimal MSE             & Match Acc       & Effective Acc                     \\ \midrule
	\multicolumn{1}{c|}{\multirow{4}{*}{\rotatebox{90}{Energy}}}  & 2D                            & 0.987$\pm$0.009         & 0.412$\pm$0.007 & 0.453$\pm$0.009 & 0.320$\pm$0.006 \\
	\multicolumn{1}{c|}{}                                         & 3D                            & 0.988$\pm$0.006         & 0.368$\pm$0.010 & 0.447$\pm$0.011 & 0.559$\pm$0.011 \\
	\multicolumn{1}{c|}{}                                         & 4D                            & 0.996$\pm$0.005         & 0.376$\pm$0.006 & 0.409$\pm$0.007 & 0.645$\pm$0.020 \\
	\multicolumn{1}{c|}{}                                         & 7D                            & 0.643$\pm$0.006         & 0.397$\pm$0.009 & 0.399$\pm$0.008 & 0.237$\pm$0.005 \\ \midrule
	\multicolumn{1}{c|}{\multirow{4}{*}{\rotatebox{90}{Traffic}}} & 2D                            & 0.977$\pm$0.008         & 0.279$\pm$0.006 & 0.139$\pm$0.003 & 0.775$\pm$0.016 \\
	\multicolumn{1}{c|}{}                                         & 3D                            & 0.964$\pm$0.007         & 0.325$\pm$0.006 & 0.388$\pm$0.009 & 0.819$\pm$0.008 \\
	\multicolumn{1}{c|}{}                                         & 4D                            & 0.979$\pm$0.008         & 0.296$\pm$0.006 & 0.408$\pm$0.006 & 0.840$\pm$0.010 \\
	\multicolumn{1}{c|}{}                                         & 7D                            & 0.944$\pm$0.010         & 0.177$\pm$0.003 & 0.291$\pm$0.008 & 0.678$\pm$0.011 \\ \midrule
	\multicolumn{1}{c|}{\multirow{4}{*}{\rotatebox{90}{Weather}}} & 2D                            & 0.995$\pm$0.005         & 0.643$\pm$0.015 & 0.380$\pm$0.008 & 0.679$\pm$0.009 \\
	\multicolumn{1}{c|}{}                                         & 3D                            & 0.987$\pm$0.008         & 0.638$\pm$0.010 & 0.556$\pm$0.012 & 0.848$\pm$0.009 \\
	\multicolumn{1}{c|}{}                                         & 4D                            & 0.991$\pm$0.005         & 0.586$\pm$0.011 & 0.522$\pm$0.009 & 0.889$\pm$0.016 \\
	\multicolumn{1}{c|}{}                                         & 7D                            & 0.978$\pm$0.007         & 0.619$\pm$0.018 & 0.518$\pm$0.011 & 0.885$\pm$0.024 \\ \bottomrule
\end{tabular}
		}
		\label{tab5}
	\end{minipage}
	\hfill
	\begin{minipage}{0.51\textwidth}
		\centering
		\caption{The inference time and memory usage of ChronoSteer across various historical windows. The historical windows vary across \{2, 3, 4, 7\} days, while the prediction window is fixed at 1 day. A lower value denotes superior efficiency.}
		\resizebox{\linewidth}{!}
		{
			\begin{tabular}{@{}cc|cc|cc@{}}
	\toprule
	\multicolumn{2}{c|}{\multirow{2}{*}{Method}}                  & \multicolumn{2}{c|}{Unimodal} & \multicolumn{2}{c}{Multimodal}                                                              \\ \cmidrule(l){3-6}
	\multicolumn{2}{c|}{}                                         & Time ($10^{-3}$ s)            & Param ($10^{2}$ MB)            & Time ($10^{-3}$ s) & Param ($10^{2}$ MB)                   \\ \midrule
	\multicolumn{1}{c|}{\multirow{4}{*}{\rotatebox{90}{Energy}}}  & 2D                            & 8.370$\pm$0.187                & 7.831$\pm$0.000    & 7.972$\pm$0.118     & 7.901$\pm$0.000 \\
	\multicolumn{1}{c|}{}                                         & 3D                            & 7.875$\pm$0.158                & 7.831$\pm$0.000    & 8.010$\pm$0.144     & 7.901$\pm$0.000 \\
	\multicolumn{1}{c|}{}                                         & 4D                            & 7.929$\pm$0.184                & 7.831$\pm$0.000    & 7.959$\pm$0.129     & 7.901$\pm$0.000 \\
	\multicolumn{1}{c|}{}                                         & 7D                            & 8.061$\pm$0.168                & 7.831$\pm$0.000    & 8.005$\pm$0.196     & 7.901$\pm$0.000 \\ \midrule
	\multicolumn{1}{c|}{\multirow{4}{*}{\rotatebox{90}{Traffic}}} & 2D                            & 7.922$\pm$0.141                & 7.831$\pm$0.000    & 7.837$\pm$0.120     & 7.901$\pm$0.000 \\
	\multicolumn{1}{c|}{}                                         & 3D                            & 8.021$\pm$0.138                & 7.831$\pm$0.000    & 7.980$\pm$0.108     & 7.901$\pm$0.000 \\
	\multicolumn{1}{c|}{}                                         & 4D                            & 8.009$\pm$0.172                & 7.831$\pm$0.000    & 7.847$\pm$0.126     & 7.901$\pm$0.000 \\
	\multicolumn{1}{c|}{}                                         & 7D                            & 8.002$\pm$0.090                & 7.831$\pm$0.000    & 7.961$\pm$0.139     & 7.901$\pm$0.000 \\ \midrule
	\multicolumn{1}{c|}{\multirow{4}{*}{\rotatebox{90}{Weather}}} & 2D                            & 7.929$\pm$0.130                & 7.831$\pm$0.000    & 7.907$\pm$0.196     & 7.901$\pm$0.000 \\
	\multicolumn{1}{c|}{}                                         & 3D                            & 8.014$\pm$0.148                & 7.831$\pm$0.000    & 7.922$\pm$0.143     & 7.901$\pm$0.000 \\
	\multicolumn{1}{c|}{}                                         & 4D                            & 7.939$\pm$0.108                & 7.831$\pm$0.000    & 7.979$\pm$0.128     & 7.901$\pm$0.000 \\
	\multicolumn{1}{c|}{}                                         & 7D                            & 7.918$\pm$0.146                & 7.831$\pm$0.000    & 8.007$\pm$0.115     & 7.901$\pm$0.000 \\ \bottomrule
\end{tabular}
		}
		\label{tab6}
	\end{minipage}
\end{table}

Table \ref{tab2} shows that ChronoSteer, trained on synthetic alignment pairs, outperforms the compared methods across the tested historical lengths. When integrated with an LLM, ChronoSteer achieves a 25.8\% improvement over its unimodal TSFM backbone, Chronos \cite{Chronos} (referred to as ChronoSteer-Unimodal). Furthermore, it also outperforms previous state-of-the-art unimodal (TimesFM-2.0) and multimodal (DeepSeek-R1) methods by 17.8\% and 22.5\%.

As illustrated in Table \ref{tab2}, both temporal and textual information contribute to enhanced prediction accuracy. In the absence of textual data, the unimodal TSFM, TimesFM, yields the best performance. Even when other baselines incorporate textual context, TimesFM and Chronos remain competitive due to their superior temporal modeling capabilities. Further analysis reveals that for short historical series, multimodal LLMs can partially compensate for limited temporal modeling by leveraging their textual reasoning capabilities, thereby outperforming unimodal TSFMs. However, as the historical series length increases, temporal information becomes sufficient; consequently, multimodal LLMs, which are limited in their ability to perform precise numerical inference, begin to underperform.

To complement the fixed prediction window experiments, we invert the experimental setup to evaluate model performance across varying prediction horizons while fixing the historical window. Specifically, we fix the historical window at 7 days and vary the prediction window across \{6 hours, 12 hours, 2 days, 4 days\}. The historical window is maintained larger than the prediction window to ensure sufficient information for zero-shot forecasting. Table \ref{tab3} shows that ChronoSteer maintains its advantage across the tested prediction windows. Moreover, consistent with observations in Table \ref{tab2}, existing multimodal approaches lose their competitive edge when sufficient historical context is available. This insight motivates our research: by synergizing the textual reasoning strengths of LLMs with the temporal modeling capabilities of TSFMs, we aim to develop a multimodal TSFM that concurrently leverages both modalities for effective zero-shot future inference.

Beyond reporting the mean and standard deviation, we further conduct paired statistical significance tests to examine whether the improvements are consistent at the sample level. For each domain, window setting, and metric, ChronoSteer Multimodal is compared with the strongest non-ChronoSteer baseline in the same row using a two-sided paired Wilcoxon signed-rank test on per-sample errors, followed by Holm-Bonferroni correction over all 48 primary comparisons. The significance markers in Tables \ref{tab2} and \ref{tab3} show that the improvement is significant in 44 of 48 primary comparisons, which provides evidence beyond mean and standard deviation reporting. The few non-significant cases correspond to rows with very small error gaps, while the overall ranking remains consistent with the main conclusion.

To further examine the role of the proposed training strategy, we conduct three additional comparative experiments, summarized in Table \ref{tab4}. Two LLM-based variants directly modify the initial predictions of the unimodal TSFM by incorporating the historical series and textual information ($ \hat{\mathbf{y}} = \mathcal{M}_\text{text} \left( \mathbf{x}, \bar{\mathbf{y}}, \mathbf{e} \right)$). The Function variant instead adjusts $\bar{\mathbf{y}}$ using the transformation function tied to the anchor most aligned with the textual instruction ($ \tilde{\mathbf{y}} = \mathcal{F}_\text{c} \left( \bar{\mathbf{y}} \right)$). Both LLM-based variants benefit from the initial predictions and improve over their direct multimodal counterparts in Table \ref{tab2}, but they remain less accurate than ChronoSteer. The Function variant performs the worst, indicating that fixed transformation alone are insufficient for temporally consistent revisions.

Furthermore, we evaluate the coverage of the compact codebook and the accuracy of LLM-generated revision instructions to investigate factors limiting system performance. A sample is considered covered if there exists an anchor instruction that reduces the prediction error compared to the unimodal backbone. As shown in the first two columns of Table \ref{tab5}, the discretized codebook covers more than 95.2\% of samples from real-world scenarios (col 1). If the optimal instruction is selected for each sample (col 2), the prediction error can be further reduced by approximately 34.6\% relative to the actual results in Table \ref{tab2}, indicating that the codebook possesses sufficient representational capacity. Section \ref{subsection:Codebook_Analysis} provides further analysis regarding these anchors.

Therefore, the primary performance bottleneck lies in the accuracy of revision instructions generated by the LLMs. The last two columns of Table \ref{tab5} report this accuracy, showing that only 40.9\% (col 3) of generated instructions match the optimal anchors. However, matching the exact optimal anchor is not a strict prerequisite for reducing prediction error, as anchor coverage is not exhaustive. Conversely, missing the optimal anchor does not preclude error reduction. Multiple anchors can benefit the same sample rather than following a strict one-to-one correspondence (e.g., both \textit{Increase Trend} and \textit{Raise Troughs} may improve the forecast in some cases). Table \ref{tab5} (col 4) shows that the LLM identifies a beneficial anchor in 68.1\% of scenarios. This rate measures whether the generated instruction reduces prediction error, whereas the 40.9\% exact-match rate measures whether it selects the single best anchor. While this success rate is not high, it demonstrates practical value. Section \ref{subsection:Prediction_Showcase} details further analysis on instruction generation. These observations once again highlight the inherent limitations of LLMs in temporal reasoning \cite{TSFUBenchmark, TSandLanguage}, which, although beyond the scope of this study, remain an important direction for future research. Specifically, prompt refinement, candidate instruction reranking, and the fine-tuning of specialized small models are potential directions worth exploring. Such approaches could augment domain knowledge for event analysis and deepen the comprehension of time-series features. Accurately understanding the current state appears to be a challenging yet essential prerequisite for generating precise incremental corrections.

To assess the practicality of ChronoSteer in real-world scenarios, we compare it with its unimodal backbone, Chronos, focusing on inference speed and memory consumption. A lower value denotes superior efficiency. The results in Table \ref{tab6} demonstrate that by integrating a lightweight textual instruction processing branch and employing a training strategy that freezes the backbone network, ChronoSteer inference speed remains nearly identical to that of Chronos, even showing a slight enhancement, while memory usage increases by only 0.9\%. In fact, the true efficiency bottleneck for the entire system, regarding both latency and memory, stems from the LLM component responsible for instruction generation. Our existing hardware resources fail to support even local deployment, which necessitates reliance on official API calls as a substitute. Fortunately, the task assigned to the LLM involves generating high-level text instructions rather than directly predicting values for every timestamp. Consequently, under comparable conditions, our method requires significantly fewer tokens than other multimodal foundation models, especially when the prediction window is long. Fine-tuning smaller language models such as SmolLM2-135M \cite{SmolLM2} or Gemma-3-270M \cite{Gemma} presents a viable solution for the future. In conclusion, we fully preserve the unimodal prediction accuracy of Chronos while extending a multimodal interface at a minimal cost. This approach effectively addresses scenarios where users are willing to invest more resources to pursue higher prediction precision.

\subsection{Prediction Showcase}
\label{subsection:Prediction_Showcase}

\begin{figure}[t]
	\centering
	\resizebox{0.95\linewidth}{!}
	{
		\includegraphics{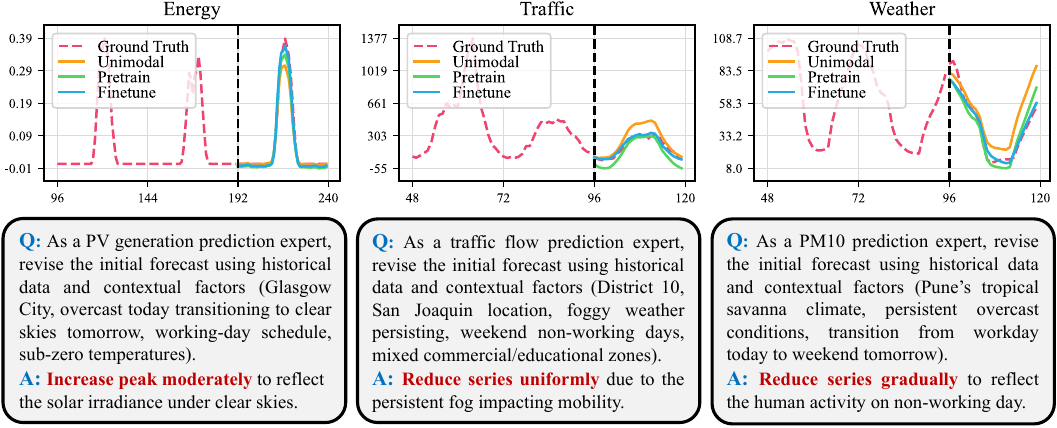}
	}
	\caption{The illustration of prediction showcases for ChronoSteer, including core text context and revision instructions. The historical and prediction windows are fixed at 4 days and 1 day, respectively.}
	\label{fig2}
\end{figure}

\begin{table}[t]
	\centering
	\caption{The prediction errors and instruction accuracy of ChronoSteer on different training stages. The historical windows vary across \{2, 3, 4, 7\} days, while the prediction window is fixed at 1 day. A lower value denotes superior accuracy, except for Effective Acc.}
	\resizebox{\linewidth}{!}
	{
		\begin{tabular}{@{}cc|cc|ccc|ccc|ccc@{}}
	\toprule
	\multicolumn{2}{c|}{\multirow{2}{*}{Method}}                  & \multicolumn{2}{c|}{Unimodal} & \multicolumn{3}{c|}{Function} & \multicolumn{3}{c|}{Pretrain} & \multicolumn{3}{c}{Finetune}                                                                                                                                                 \\ \cmidrule(l){3-13}
	\multicolumn{2}{c|}{}                                         & MSE                           & MAE                           & MSE                           & MAE                          & Effect. Acc     & MSE             & MAE             & Effect. Acc     & MSE             & MAE             & Effect. Acc                       \\ \midrule
	\multicolumn{1}{c|}{\multirow{4}{*}{\rotatebox{90}{Energy}}}  & 2D                            & 0.765$\pm$0.000               & 0.277$\pm$0.000               & 0.811$\pm$0.033              & 0.366$\pm$0.011 & 0.191$\pm$0.005 & 0.669$\pm$0.016 & 0.298$\pm$0.006 & 0.206$\pm$0.005 & 0.573$\pm$0.012 & 0.265$\pm$0.007 & 0.320$\pm$0.006 \\
	\multicolumn{1}{c|}{}                                         & 3D                            & 0.763$\pm$0.000               & 0.268$\pm$0.000               & 0.803$\pm$0.017              & 0.365$\pm$0.013 & 0.362$\pm$0.006 & 0.680$\pm$0.013 & 0.291$\pm$0.004 & 0.411$\pm$0.007 & 0.591$\pm$0.008 & 0.267$\pm$0.004 & 0.559$\pm$0.011 \\
	\multicolumn{1}{c|}{}                                         & 4D                            & 0.697$\pm$0.000               & 0.259$\pm$0.000               & 0.788$\pm$0.027              & 0.362$\pm$0.016 & 0.482$\pm$0.012 & 0.634$\pm$0.017 & 0.279$\pm$0.005 & 0.567$\pm$0.006 & 0.613$\pm$0.011 & 0.253$\pm$0.005 & 0.645$\pm$0.020 \\
	\multicolumn{1}{c|}{}                                         & 7D                            & 0.573$\pm$0.000               & 0.246$\pm$0.000               & 0.657$\pm$0.022              & 0.389$\pm$0.009 & 0.159$\pm$0.003 & 0.560$\pm$0.013 & 0.263$\pm$0.005 & 0.188$\pm$0.004 & 0.524$\pm$0.006 & 0.243$\pm$0.005 & 0.237$\pm$0.005 \\ \midrule
	\multicolumn{1}{c|}{\multirow{4}{*}{\rotatebox{90}{Traffic}}} & 2D                            & 0.669$\pm$0.000               & 0.571$\pm$0.000               & 0.648$\pm$0.023              & 0.603$\pm$0.022 & 0.647$\pm$0.007 & 0.564$\pm$0.015 & 0.552$\pm$0.011 & 0.688$\pm$0.011 & 0.512$\pm$0.005 & 0.508$\pm$0.013 & 0.775$\pm$0.016 \\
	\multicolumn{1}{c|}{}                                         & 3D                            & 0.608$\pm$0.000               & 0.551$\pm$0.000               & 0.484$\pm$0.021              & 0.527$\pm$0.017 & 0.577$\pm$0.010 & 0.405$\pm$0.007 & 0.477$\pm$0.009 & 0.664$\pm$0.012 & 0.401$\pm$0.007 & 0.448$\pm$0.006 & 0.819$\pm$0.008 \\
	\multicolumn{1}{c|}{}                                         & 4D                            & 0.559$\pm$0.000               & 0.522$\pm$0.000               & 0.495$\pm$0.019              & 0.547$\pm$0.014 & 0.752$\pm$0.014 & 0.384$\pm$0.007 & 0.468$\pm$0.008 & 0.785$\pm$0.007 & 0.367$\pm$0.006 & 0.430$\pm$0.007 & 0.840$\pm$0.010 \\
	\multicolumn{1}{c|}{}                                         & 7D                            & 0.344$\pm$0.000               & 0.390$\pm$0.000               & 0.462$\pm$0.015              & 0.523$\pm$0.029 & 0.578$\pm$0.010 & 0.279$\pm$0.005 & 0.376$\pm$0.005 & 0.613$\pm$0.013 & 0.242$\pm$0.004 & 0.332$\pm$0.006 & 0.678$\pm$0.011 \\ \midrule
	\multicolumn{1}{c|}{\multirow{4}{*}{\rotatebox{90}{Weather}}} & 2D                            & 2.340$\pm$0.000               & 1.126$\pm$0.000               & 2.125$\pm$0.049              & 1.077$\pm$0.040 & 0.639$\pm$0.009 & 1.959$\pm$0.037 & 1.035$\pm$0.018 & 0.628$\pm$0.010 & 1.532$\pm$0.029 & 0.910$\pm$0.019 & 0.679$\pm$0.009 \\
	\multicolumn{1}{c|}{}                                         & 3D                            & 1.511$\pm$0.000               & 0.967$\pm$0.000               & 1.164$\pm$0.064              & 0.818$\pm$0.022 & 0.715$\pm$0.012 & 1.020$\pm$0.015 & 0.773$\pm$0.019 & 0.709$\pm$0.015 & 0.898$\pm$0.015 & 0.726$\pm$0.014 & 0.848$\pm$0.009 \\
	\multicolumn{1}{c|}{}                                         & 4D                            & 1.358$\pm$0.000               & 0.909$\pm$0.000               & 1.055$\pm$0.038              & 0.786$\pm$0.027 & 0.757$\pm$0.021 & 0.975$\pm$0.014 & 0.762$\pm$0.013 & 0.826$\pm$0.014 & 0.800$\pm$0.014 & 0.673$\pm$0.008 & 0.889$\pm$0.016 \\
	\multicolumn{1}{c|}{}                                         & 7D                            & 1.116$\pm$0.000               & 0.822$\pm$0.000               & 0.868$\pm$0.030              & 0.716$\pm$0.033 & 0.760$\pm$0.011 & 0.847$\pm$0.020 & 0.685$\pm$0.015 & 0.803$\pm$0.013 & 0.754$\pm$0.021 & 0.646$\pm$0.012 & 0.885$\pm$0.024 \\ \bottomrule
\end{tabular}
	}
	\label{tab7}
\end{table}

To visually illustrate the performance of ChronoSteer across different stages, the top row of Figure \ref{fig2} displays prediction results using a 1-day prediction window and a 4-day historical window (with only the last 2 days shown due to space constraints). The bottom row presents core textual information and revision instructions generated by the LLM. Further details, including the visual analysis of the anchor revision instructions, are provided in Appendix \ref{appendix:Detailed_Prediction_Showcase}. As shown in Figure \ref{fig2}, ChronoSteer, after two-stage fine-tuning, effectively utilizes textual revision instructions to steer the outputs of the unimodal TSFM. Furthermore, when the instructions are somewhat imprecise, ChronoSteer robustly balances them with historical series for improved temporal consistency.

In the Energy dataset, foggy weather causes a gradual decline in peaks within the historical series. Without textual context, a unimodal TSFM extrapolates this trend and forecasts even lower peaks. Here the instruction is appropriate. The weather forecast indicates \textit{"transitioning to clear skies tomorrow"}, so the LLM generates an instruction to raise the peaks, and ChronoSteer adjusts its prediction upward. After fine-tuning, the adjustment magnitude more closely matches the observed future series.

In the Traffic dataset, persistent heavy fog causes a further reduction in traffic flow peaks. Without textual information, a unimodal TSFM predicts a typical non-working day pattern similar to the previous day. This example shows an imprecise instruction case. The LLM provides the revision instruction \textit{"reduce series uniformly"}, although the event should mainly affect the peaks rather than the entire series. After pre-training, ChronoSteer follows the instruction strictly, resulting in unrealistic troughs. After fine-tuning, it integrates the instruction with historical series and partly mitigates this issue.

In the Weather dataset, the transition from weekdays to non-working days reduces air pollution due to decreased human activity. Without textual information, a unimodal TSFM struggles to capture this pattern and overestimates future values. This example shows that a directionally correct instruction may still need historical context to determine the appropriate magnitude. The revision instruction guides ChronoSteer to lower the prediction. After fine-tuning, ChronoSteer predicts troughs and peaks more accurately by balancing this instruction with historical context.

Finally, we conduct a decoupled analysis to determine the accuracy of LLM-generated revision instructions and evaluate the ability of ChronoSteer to rectify imprecise LLM instructions using historical series. We compare the impact of these instructions on ChronoSteer across different stages in Table \ref{tab7}. We employ auxiliary metrics for indirect measurement as MTSFBench-300 lacks ground-truth anchor labels. First, the Function method directly modifies the original unimodal predictions by using the transformation function corresponding to the anchor most aligned with the textual instructions. This method fails to significantly improve prediction accuracy compared to unimodal predictions despite the LLM instructions being effective in 55.2\% of scenarios, as it lacks adjustment magnitude information from historical series. The pre-trained ChronoSteer captures adjustment magnitudes more precisely even though it lacks error correction capabilities. While the instruction effectiveness rate of 59.1\%, which reflects the true accuracy of LLM-generated revision instructions, shows no significant increase over the Function method, prediction accuracy improves by 12.7\%. The fine-tuned ChronoSteer leverages historical and observed future series to correct erroneous instructions to a certain extent. This version corrects an additional 9.0\% of erroneous instructions compared to the pre-trained model and further enhances accuracy by 11.3\%.

\begin{table}[t]
	\centering
	\caption{The prediction errors of ChronoSteer and ablation variants. The historical windows vary across \{2, 3, 4, 7\} days, while the prediction window is fixed at 1 day. A lower value denotes superior accuracy. The optimal results are highlighted in bold.}
	\resizebox{0.85\linewidth}{!}
	{
		\begin{tabular}{@{}cc|cc|cccccc@{}}
	\toprule
	\multicolumn{2}{c|}{\multirow{3}{*}{Method}}                  & \multicolumn{2}{c|}{Ours}        & \multicolumn{6}{c}{Variant}                                                                                                                                                                                                                              \\ \cmidrule(l){3-10}
	\multicolumn{2}{c|}{}                                         & \multicolumn{2}{c|}{ChronoSteer} & \multicolumn{2}{c|}{w/o Contrast} & \multicolumn{2}{c|}{w/o Finetune} & \multicolumn{2}{c}{w/o MLP}                                                                                                                                                      \\ \cmidrule(l){3-10}
	\multicolumn{2}{c|}{}                                         & \multicolumn{1}{c}{MSE}          & \multicolumn{1}{c|}{MAE}          & \multicolumn{1}{c}{MSE}           & \multicolumn{1}{c|}{MAE}    & \multicolumn{1}{c}{MSE}              & \multicolumn{1}{c|}{MAE} & \multicolumn{1}{c}{MSE}              & \multicolumn{1}{c}{MAE}                   \\ \midrule
	\multicolumn{1}{c|}{\multirow{4}{*}{\rotatebox{90}{Energy}}}  & 2D                               & \textbf{0.573$\pm$0.012}          & \textbf{0.265$\pm$0.007}          & 0.659$\pm$0.012             & \multicolumn{1}{l|}{0.298$\pm$0.004} & 0.669$\pm$0.016          & \multicolumn{1}{l|}{0.298$\pm$0.006} & 0.662$\pm$0.014         & 0.323$\pm$0.006 \\
	\multicolumn{1}{c|}{}                                         & 3D                               & \textbf{0.591$\pm$0.008}          & \textbf{0.267$\pm$0.004}          & 0.633$\pm$0.013             & \multicolumn{1}{l|}{0.277$\pm$0.003} & 0.680$\pm$0.013          & \multicolumn{1}{l|}{0.291$\pm$0.004} & 0.631$\pm$0.012         & 0.300$\pm$0.006 \\
	\multicolumn{1}{c|}{}                                         & 4D                               & \textbf{0.613$\pm$0.011}          & \textbf{0.253$\pm$0.005}          & 0.630$\pm$0.011             & \multicolumn{1}{l|}{0.271$\pm$0.004} & 0.634$\pm$0.017          & \multicolumn{1}{l|}{0.279$\pm$0.005} & 0.626$\pm$0.013         & 0.290$\pm$0.005 \\
	\multicolumn{1}{c|}{}                                         & 7D                               & \textbf{0.524$\pm$0.006}          & \textbf{0.243$\pm$0.005}          & 0.556$\pm$0.006             & \multicolumn{1}{l|}{0.262$\pm$0.004} & 0.560$\pm$0.013          & \multicolumn{1}{l|}{0.263$\pm$0.005} & 0.540$\pm$0.008         & 0.260$\pm$0.005 \\ \midrule
	\multicolumn{1}{c|}{\multirow{4}{*}{\rotatebox{90}{Traffic}}} & 2D                               & \textbf{0.512$\pm$0.005}          & \textbf{0.508$\pm$0.013}          & 0.557$\pm$0.007             & \multicolumn{1}{l|}{0.525$\pm$0.010} & 0.564$\pm$0.015          & \multicolumn{1}{l|}{0.552$\pm$0.011} & 0.607$\pm$0.010         & 0.592$\pm$0.012 \\
	\multicolumn{1}{c|}{}                                         & 3D                               & \textbf{0.401$\pm$0.007}          & \textbf{0.448$\pm$0.006}          & 0.419$\pm$0.008             & \multicolumn{1}{l|}{0.452$\pm$0.005} & 0.405$\pm$0.007          & \multicolumn{1}{l|}{0.477$\pm$0.009} & 0.420$\pm$0.005         & 0.461$\pm$0.005 \\
	\multicolumn{1}{c|}{}                                         & 4D                               & \textbf{0.367$\pm$0.006}          & \textbf{0.430$\pm$0.007}          & 0.385$\pm$0.006             & \multicolumn{1}{l|}{0.461$\pm$0.012} & 0.384$\pm$0.007          & \multicolumn{1}{l|}{0.468$\pm$0.008} & 0.379$\pm$0.008         & 0.452$\pm$0.008 \\
	\multicolumn{1}{c|}{}                                         & 7D                               & \textbf{0.242$\pm$0.004}          & \textbf{0.332$\pm$0.006}          & 0.254$\pm$0.007             & \multicolumn{1}{l|}{0.341$\pm$0.008} & 0.279$\pm$0.005          & \multicolumn{1}{l|}{0.376$\pm$0.005} & 0.244$\pm$0.003         & 0.344$\pm$0.007 \\ \midrule
	\multicolumn{1}{c|}{\multirow{4}{*}{\rotatebox{90}{Weather}}} & 2D                               & \textbf{1.532$\pm$0.029}          & \textbf{0.910$\pm$0.019}          & 1.590$\pm$0.035             & \multicolumn{1}{l|}{0.925$\pm$0.019} & 1.959$\pm$0.037          & \multicolumn{1}{l|}{1.035$\pm$0.018} & 1.705$\pm$0.021         & 0.968$\pm$0.024 \\
	\multicolumn{1}{c|}{}                                         & 3D                               & \textbf{0.898$\pm$0.015}          & \textbf{0.726$\pm$0.014}          & 0.948$\pm$0.015             & \multicolumn{1}{l|}{0.750$\pm$0.011} & 1.020$\pm$0.015          & \multicolumn{1}{l|}{0.773$\pm$0.019} & 0.996$\pm$0.012         & 0.741$\pm$0.010 \\
	\multicolumn{1}{c|}{}                                         & 4D                               & \textbf{0.800$\pm$0.014}          & \textbf{0.673$\pm$0.008}          & 0.867$\pm$0.015             & \multicolumn{1}{l|}{0.698$\pm$0.020} & 0.975$\pm$0.014          & \multicolumn{1}{l|}{0.762$\pm$0.013} & 0.887$\pm$0.015         & 0.712$\pm$0.010 \\
	\multicolumn{1}{c|}{}                                         & 7D                               & \textbf{0.754$\pm$0.021}          & \textbf{0.646$\pm$0.012}          & 0.794$\pm$0.009             & \multicolumn{1}{l|}{0.675$\pm$0.013} & 0.847$\pm$0.020          & \multicolumn{1}{l|}{0.685$\pm$0.015} & 0.760$\pm$0.024         & 0.652$\pm$0.014 \\ \bottomrule
\end{tabular}
	}
	\label{tab8}
\end{table}

\subsection{Ablation Study}
\label{subsection:Ablation_Study}

To validate the effectiveness of each component, we evaluate ChronoSteer and its variants in Table \ref{tab8}. The results show that modifying any component will reduce the performance, confirming its architectural soundness.

In w/o Contrastive, the contrastive learning loss is omitted during pretraining. This absence hinders the ability of ChronoSteer to effectively distinguish between different anchor instructions, thereby reducing the cross-modal alignment performance. Specifically, removing operation leads to a prediction accuracy drop of 7.7\%, 4.7\%, and 4.5\% on the Energy, Traffic, and Weather datasets, respectively.

In w/o Finetune, the fine-tuning stage is omitted. Since the model has already converged after pretraining, its parameters are directly evaluated. During fine-tuning, observed future series combined with pseudo-labeling techniques serve as supervised targets, allowing the model to recover the fine-grained magnitude information lost during discretization and to mitigate the expressive limitations of the synthetic transformation functions. Without fine-tuning, prediction accuracy drops by 10.4\%, 8.2\%, and 16.1\% on the Energy, Traffic, and Weather datasets, respectively.

In w/o MLP, the MLP in the alignment module is replaced with a linear layer, similar to the design in LLaVA \cite{LLaVA}. This substitution causes a substantial performance drop. Unlike LLaVA, which represents an image with multiple tokens, our model encodes each textual instruction as a single token. This design requires a more expressive nonlinear mapper to achieve effective cross-modal alignment. The decision to use a single-token embedding arises from the scarcity of high-quality cross-modal alignment datasets. Using multiple tokens not only increases computational cost but also tends to produce collapsing solutions without large-scale alignment data, whereas a well-pretrained text embedding model is better suited for distinguishing instruction semantics.

\subsection{Codebook Analysis}
\label{subsection:Codebook_Analysis}

\begin{table}[t]
	\centering
	\caption{The prediction errors and codebook coverage of ChronoSteer with various anchor configurations. The historical windows vary across \{2, 3, 4, 7\} days, while the prediction window is fixed at 1 day. We report the results averaged across these different historical windows. A lower value denotes superior accuracy, except for Coverage Ratio.}
	\resizebox{\linewidth}{!}
	{
		\begin{tabular}{@{}cc@{\hspace{3pt}}c@{\hspace{3pt}}c@{\hspace{3pt}}c|c@{\hspace{3pt}}c@{\hspace{3pt}}c|c@{\hspace{3pt}}c@{\hspace{3pt}}c|c@{\hspace{3pt}}c@{\hspace{3pt}}c@{}}
	\toprule
	\multicolumn{1}{c|}{\multirow{2}{*}{Embed}} & \multirow{2}{*}{Trend} & \multirow{2}{*}{Ampl.} & \multirow{2}{*}{Peak} & \multirow{2}{*}{Trough} & \multicolumn{3}{c|}{Energy} & \multicolumn{3}{c|}{Traffic} & \multicolumn{3}{c}{Weather}                                                                                                                                                                   \\ \cmidrule(l){6-14}
	\multicolumn{1}{c|}{}                       &                        &                        &                       &                         & LLM MSE                     & Cover. Ratio                 & Opt. MSE                    & LLM MSE                  & Cover. Ratio             & Opt.l MSE                & LLM MSE                  & Cover. Ratio             & Opt. MSE                 \\ \midrule
	\multicolumn{5}{c|}{Unimodal}               & 0.699$\pm$0.000        & -                      & -                     & 0.545$\pm$0.000         & -                           & -                            & 1.581$\pm$0.000             & -                        & -                                                                                                                                    \\ \midrule
	\multicolumn{1}{c|}{\ding{55}}              & \ding{55}              & \ding{55}              & \ding{55}             & \ding{55}               & 0.706$\pm$0.052             & -                            & -                           & 0.534$\pm$0.032          & -                        & -                        & 1.608$\pm$0.072          & -                        & -                        \\
	\multicolumn{1}{c|}{\ding{51}}              & \ding{55}              & \ding{55}              & \ding{55}             & \ding{55}               & 0.672$\pm$0.048             & -                            & -                           & 0.531$\pm$0.040          & -                        & -                        & 1.539$\pm$0.065          &                          & -                        \\ \midrule
	\multicolumn{1}{c|}{\ding{51}}              & \ding{51}              & \ding{55}              & \ding{55}             & \ding{55}               & 0.597$\pm$0.008             & 0.882$\pm$0.014              & 0.427$\pm$0.005             & 0.451$\pm$0.006          & 0.924$\pm$0.009          & 0.277$\pm$0.004          & 1.114$\pm$0.011          & 0.977$\pm$0.011          & 0.659$\pm$0.012          \\
	\multicolumn{1}{c|}{\ding{51}}              & \ding{55}              & \ding{51}              & \ding{55}             & \ding{55}               & 0.625$\pm$0.014             & 0.737$\pm$0.010              & 0.472$\pm$0.006             & 0.500$\pm$0.008          & 0.869$\pm$0.009          & 0.298$\pm$0.003          & 1.321$\pm$0.018          & 0.944$\pm$0.010          & 0.724$\pm$0.010          \\
	\multicolumn{1}{c|}{\ding{51}}              & \ding{55}              & \ding{55}              & \ding{51}             & \ding{55}               & 0.656$\pm$0.007             & 0.728$\pm$0.004              & 0.495$\pm$0.006             & 0.532$\pm$0.006          & 0.843$\pm$0.010          & 0.312$\pm$0.004          & 1.367$\pm$0.017          & 0.936$\pm$0.011          & 0.816$\pm$0.009          \\
	\multicolumn{1}{c|}{\ding{51}}              & \ding{55}              & \ding{55}              & \ding{55}             & \ding{51}               & 0.644$\pm$0.009             & 0.710$\pm$0.010              & 0.534$\pm$0.006             & 0.525$\pm$0.010          & 0.718$\pm$0.010          & 0.336$\pm$0.006          & 1.351$\pm$0.027          & 0.737$\pm$0.012          & 0.841$\pm$0.012          \\ \midrule
	\multicolumn{1}{c|}{\ding{51}}              & \ding{51}              & \ding{51}              & \ding{55}             & \ding{55}               & 0.586$\pm$0.008             & 0.888$\pm$0.011              & 0.413$\pm$0.005             & 0.417$\pm$0.006          & 0.942$\pm$0.012          & 0.274$\pm$0.004          & 1.085$\pm$0.012          & 0.982$\pm$0.011          & 0.637$\pm$0.008          \\
	\multicolumn{1}{c|}{\ding{51}}              & \ding{51}              & \ding{55}              & \ding{51}             & \ding{55}               & 0.609$\pm$0.008             & 0.810$\pm$0.006              & 0.441$\pm$0.006             & 0.470$\pm$0.008          & 0.917$\pm$0.009          & 0.282$\pm$0.004          & 1.200$\pm$0.020          & 0.975$\pm$0.011          & 0.675$\pm$0.010          \\
	\multicolumn{1}{c|}{\ding{51}}              & \ding{51}              & \ding{55}              & \ding{55}             & \ding{51}               & 0.609$\pm$0.007             & 0.808$\pm$0.010              & 0.437$\pm$0.005             & 0.460$\pm$0.004          & 0.923$\pm$0.013          & 0.278$\pm$0.003          & 1.173$\pm$0.020          & 0.971$\pm$0.013          & 0.659$\pm$0.009          \\
	\multicolumn{1}{c|}{\ding{51}}              & \ding{55}              & \ding{51}              & \ding{51}             & \ding{55}               & 0.616$\pm$0.008             & 0.793$\pm$0.011              & 0.449$\pm$0.004             & 0.479$\pm$0.006          & 0.892$\pm$0.010          & 0.286$\pm$0.006          & 1.308$\pm$0.014          & 0.968$\pm$0.013          & 0.693$\pm$0.006          \\
	\multicolumn{1}{c|}{\ding{51}}              & \ding{55}              & \ding{51}              & \ding{55}             & \ding{51}               & 0.615$\pm$0.006             & 0.742$\pm$0.010              & 0.461$\pm$0.006             & 0.488$\pm$0.005          & 0.884$\pm$0.008          & 0.295$\pm$0.004          & 1.312$\pm$0.021          & 0.955$\pm$0.010          & 0.715$\pm$0.008          \\
	\multicolumn{1}{c|}{\ding{51}}              & \ding{55}              & \ding{55}              & \ding{51}             & \ding{51}               & 0.624$\pm$0.006             & 0.728$\pm$0.011              & 0.476$\pm$0.007             & 0.526$\pm$0.009          & 0.864$\pm$0.010          & 0.304$\pm$0.005          & 1.327$\pm$0.023          & 0.940$\pm$0.010          & 0.794$\pm$0.005          \\ \midrule
	\multicolumn{1}{c|}{\ding{51}}              & \ding{51}              & \ding{51}              & \ding{51}             & \ding{55}               & 0.586$\pm$0.005             & 0.891$\pm$0.008              & 0.407$\pm$0.005             & 0.402$\pm$0.008          & 0.943$\pm$0.005          & 0.271$\pm$0.005          & 1.089$\pm$0.016          & 0.985$\pm$0.010          & 0.633$\pm$0.010          \\
	\multicolumn{1}{c|}{\ding{51}}              & \ding{51}              & \ding{51}              & \ding{55}             & \ding{51}               & 0.594$\pm$0.008             & 0.888$\pm$0.009              & 0.421$\pm$0.009             & 0.432$\pm$0.008          & 0.921$\pm$0.012          & 0.275$\pm$0.004          & 1.100$\pm$0.022          & 0.985$\pm$0.008          & 0.651$\pm$0.008          \\
	\multicolumn{1}{c|}{\ding{51}}              & \ding{51}              & \ding{55}              & \ding{51}             & \ding{51}               & 0.597$\pm$0.008             & 0.871$\pm$0.009              & 0.434$\pm$0.006             & 0.451$\pm$0.008          & 0.919$\pm$0.011          & 0.275$\pm$0.005          & 1.166$\pm$0.019          & 0.978$\pm$0.007          & 0.660$\pm$0.010          \\
	\multicolumn{1}{c|}{\ding{51}}              & \ding{55}              & \ding{51}              & \ding{51}             & \ding{51}               & 0.612$\pm$0.008             & 0.801$\pm$0.005              & 0.450$\pm$0.005             & 0.475$\pm$0.008          & 0.906$\pm$0.008          & 0.284$\pm$0.003          & 1.296$\pm$0.019          & 0.966$\pm$0.014          & 0.670$\pm$0.010          \\ \midrule
	\multicolumn{1}{c|}{\ding{51}}              & \ding{51}              & \ding{51}              & \ding{51}             & \ding{51}               & \textbf{0.583$\pm$0.006}    & \textbf{0.894$\pm$0.011}     & \textbf{0.402$\pm$0.004}    & \textbf{0.398$\pm$0.005} & \textbf{0.945$\pm$0.014} & \textbf{0.269$\pm$0.003} & \textbf{1.039$\pm$0.006} & \textbf{0.981$\pm$0.014} & \textbf{0.622$\pm$0.008} \\ \bottomrule
\end{tabular}
	}
	\label{tab9}
\end{table}

We further investigate the necessity and rationality of the discrete codebook mechanism with the experimental results summarized in Table \ref{tab9}. Considering the costs associated with LLM API calls, this series utilizes the 10K samples obtained from the clustering in Section \ref{subsubsection:Multimodal_Triplet_Construction} for training rather than the full 100K samples. All other experimental settings remain consistent with Section \ref{subsubsection:Implementation}.

In the first set of experiments, we examine the effect of the anchor mechanism. We feed the unimodal predictions $\bar{\mathbf{y}}$ generated by the TSFM and the corresponding ground truth labels $\mathbf{y}$ into the LLM to generate revision instructions. We utilize these natural text instructions for training without applying discretization. The results indicate that neither embedding natural text into multiple control tokens via GPT-2 \cite{GPT2} nor into a single control token via BGE-M3 \cite{BgeM3} yields significant benefits. Previous studies indicate that open-ended text instructions often suffer from semantic ambiguity and a low signal-to-noise ratio \cite{XiongW0J0Z0L025}, and this characteristic imposes strict requirements on both the quality and scale of alignment data. Direct embedding into multiple tokens hinders model convergence and even leads to negative performance gains. Although embedding models reduce noise by summarizing instructions into a single token, directly synthesizing large-scale cross-modal alignment datasets with LLMs remains expensive, inefficient, and unreliable. In contrast, our discrete codebook mechanism enables the training of multimodal models within the current context of scarce high-quality data.

In the second set of experiments, we conduct ablation studies on the four types of anchors within the codebook. Experimental results demonstrate that the codebook coverage rate improves as anchors are progressively added and eventually reaches 94.0\%. The nine anchors reflect a practical balance among coverage, interpretability, and the feasibility of designing reliable transformation functions. The nine anchors are intended to cover common directional revision behaviors under limited paired supervision, rather than to exhaust all possible event effects in open-world scenarios. The ablation results in Table \ref{tab9} show that adding anchor groups increases coverage to 94.0\%, which supports the current design within the tested benchmark setting. Specifically, anchors controlling trend and amplitude yield greater performance improvements compared to those controlling peaks and troughs. Peak and trough anchors still improve coverage by handling local changes that cannot be represented by global trend or amplitude controls. At the same time, compound, delayed, nonlinear, continuous, or domain-specific effects may require combinations of anchors, richer instructions, or new transformation functions. Simply enlarging the codebook is not sufficient, because each new anchor must be paired with a reliable transformation function to provide controllable training targets. Future research directions include designing more refined anchors with corresponding transformation functions, exploring anchor composition or continuous instruction embeddings, or fine-tuning smaller LLMs \cite{SmolLM2, Gemma} to generate accurate revision instructions more efficiently.

Overall, although the discrete codebook mechanism and synthetic cross-modal associations limit the generalization of ChronoSteer to some extent, they form a necessary compromise given the current scarcity of high-quality alignment data. Furthermore, the experimental results indicate that these core anchors cover most benchmark cases and yield useful predictive gains under the tested setting.

\subsection{Hyperparameter Analysis}
\label{subsection:Hyperparameter_Analysis}

To assess the sensitivity of ChronoSteer to hyperparameter selections, we evaluate its prediction performance under various configurations. The historical length is set to 4 days and the prediction length to 1 day. The key hyperparameters examined include the learning rate, the contrastive loss weight during pre-training, and the hidden layer dimension of MLP in alignment module. To further support the design of the synthetic cross-modal supervision pipeline, we also provide a sensitivity analysis on the parameter ranges of the synthetic transformation functions in Appendix \ref{appendix:Detailed_Synthetic_Function}.

\begin{figure}[t]
	\centering
	\resizebox{0.9\linewidth}{!}
	{
		\includegraphics{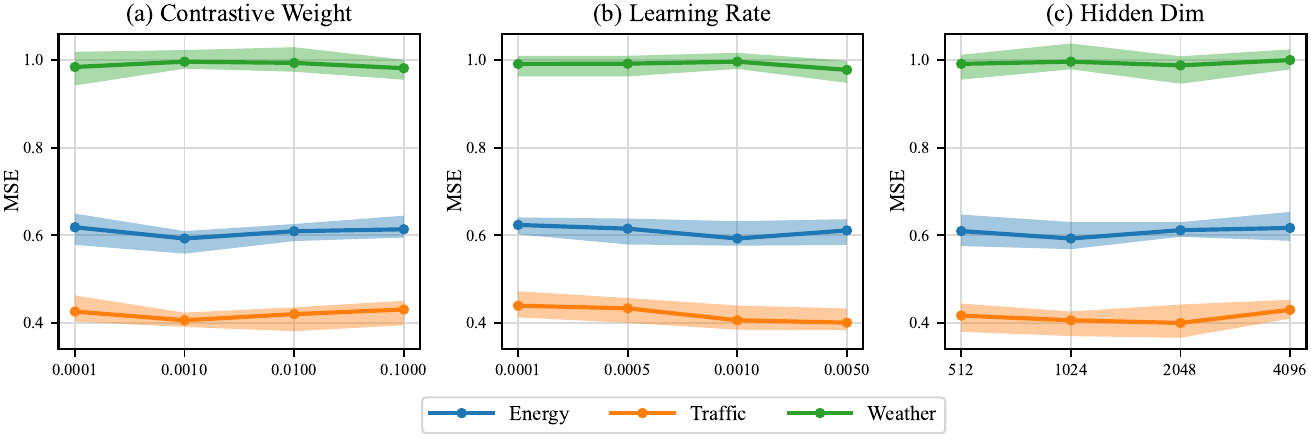}
	}
	\caption{The prediction errors of ChronoSteer with various hyperparameter configurations. The historical windows vary across \{2, 3, 4, 7\} days, while the prediction window is fixed at 1 day. We report the results averaged across these different historical windows. A lower value denotes superior accuracy.}
	\label{fig3}
\end{figure}

The contrastive loss weight during the pre-training stage directly affects the ability of ChronoSteer to distinguish anchor revision instructions. A lower weight shifts optimization focus toward MSE loss, reducing discriminative power. Conversely, a higher weight leads to over-reliance on contrastive loss, potentially hindering the capture of fine-grained temporal characteristics. As shown in Figure \ref{fig3}(a), increasing the contrastive loss weight initially boosts prediction accuracy on the Energy and Traffic datasets before declining, consistent with theoretical expectations. Notably, the unstable Weather dataset shows robustness to this parameter, achieving peak performance at a weight of 0.1.

The learning rate is a key parameter affecting model training. A small learning rate may cause slow convergence and risk trapping the model in local optima, while a large learning rate, though speeding up convergence, can lead to instability or divergence. As shown in Figure \ref{fig3}(b), ChronoSteer consistently improves prediction accuracy on the Traffic and Weather datasets as the learning rate increases, revealing a clear pattern. However, due to the high proportion of zero-value samples in the Energy dataset, it shows greater sensitivity to the learning rate and achieves optimal performance at a learning rate of 0.001.

The dimension of the hidden vector influences the learning capacity and representational power. A smaller dimension reduces computational overhead but may hinder the capture of complex characteristics. Conversely, a larger hidden vector improves expressiveness and generalization. However, this often comes at the cost of higher computation and an elevated risk of overfitting, particularly when training data is limited or of poor quality. As shown in Figure \ref{fig3}(c), prediction accuracy across all datasets initially increases and then decreases as the hidden dimension grows, consistent with theory. Optimal performance is typically achieved when it falls between 1024 and 2048.

\subsection{Robustness Analysis}
\label{subsection:Robustness_Analysis}

\begin{table}[t]
	\centering
	\caption{The prediction errors of ChronoSteer with various freezing strategies. The historical windows vary across \{2, 3, 4, 7\} days, while the prediction window is fixed at 1 day. A lower value denotes superior accuracy and robustness.}
	\resizebox{0.75\linewidth}{!}
	{
		\begin{tabular}{@{}cc|c|cc|cc|cc@{}}
	\toprule
	\multirow{2}{*}{LLM}       & \multirow{2}{*}{Align}     & \multirow{2}{*}{Hist} & \multicolumn{2}{c|}{Energy} & \multicolumn{2}{c|}{Traffic} & \multicolumn{2}{c}{Weather}                                                       \\ \cmidrule(l){4-9}
	                           &                            &                       & MSE                         & MAE                          & MSE                         & MAE             & MSE             & MAE             \\ \midrule
	\multirow{4}{*}{\ding{55}} & \multirow{4}{*}{\ding{51}} & 2D                    & 0.573$\pm$0.006             & 0.266$\pm$0.003              & 0.513$\pm$0.004             & 0.509$\pm$0.003 & 1.529$\pm$0.013 & 0.913$\pm$0.013 \\
	                           &                            & 3D                    & 0.591$\pm$0.006             & 0.267$\pm$0.002              & 0.401$\pm$0.003             & 0.446$\pm$0.005 & 0.902$\pm$0.010 & 0.726$\pm$0.007 \\
	                           &                            & 4D                    & 0.611$\pm$0.005             & 0.253$\pm$0.003              & 0.366$\pm$0.002             & 0.431$\pm$0.002 & 0.802$\pm$0.010 & 0.671$\pm$0.008 \\
	                           &                            & 7D                    & 0.523$\pm$0.005             & 0.243$\pm$0.002              & 0.242$\pm$0.002             & 0.331$\pm$0.005 & 0.760$\pm$0.007 & 0.645$\pm$0.006 \\ \midrule
	\multirow{4}{*}{\ding{51}} & \multirow{4}{*}{\ding{55}} & 2D                    & 0.572$\pm$0.006             & 0.266$\pm$0.004              & 0.514$\pm$0.007             & 0.505$\pm$0.007 & 1.537$\pm$0.022 & 0.909$\pm$0.012 \\
	                           &                            & 3D                    & 0.593$\pm$0.009             & 0.267$\pm$0.003              & 0.399$\pm$0.005             & 0.449$\pm$0.008 & 0.902$\pm$0.012 & 0.725$\pm$0.011 \\
	                           &                            & 4D                    & 0.606$\pm$0.009             & 0.254$\pm$0.004              & 0.368$\pm$0.008             & 0.426$\pm$0.008 & 0.801$\pm$0.012 & 0.671$\pm$0.008 \\
	                           &                            & 7D                    & 0.525$\pm$0.008             & 0.243$\pm$0.005              & 0.241$\pm$0.003             & 0.332$\pm$0.004 & 0.762$\pm$0.009 & 0.643$\pm$0.006 \\ \midrule
	\multirow{4}{*}{\ding{51}} & \multirow{4}{*}{\ding{51}} & 2D                    & 0.573$\pm$0.012             & 0.265$\pm$0.007              & 0.512$\pm$0.005             & 0.508$\pm$0.013 & 1.532$\pm$0.029 & 0.910$\pm$0.019 \\
	                           &                            & 3D                    & 0.591$\pm$0.008             & 0.267$\pm$0.004              & 0.401$\pm$0.007             & 0.448$\pm$0.006 & 0.898$\pm$0.015 & 0.726$\pm$0.014 \\
	                           &                            & 4D                    & 0.613$\pm$0.011             & 0.253$\pm$0.005              & 0.367$\pm$0.006             & 0.430$\pm$0.007 & 0.800$\pm$0.014 & 0.673$\pm$0.008 \\
	                           &                            & 7D                    & 0.524$\pm$0.006             & 0.243$\pm$0.005              & 0.242$\pm$0.004             & 0.332$\pm$0.006 & 0.754$\pm$0.021 & 0.646$\pm$0.012 \\ \bottomrule
\end{tabular}
	}
	\label{tab10}
\end{table}

Table \ref{tab10} presents the statistical properties of the evaluation metrics for ChronoSteer across three independent runs to assess its robustness. Three experiment setups are designed to evaluate system performance:

\begin{itemize}

	\item In the first experiment, the revision instructions derived from textual events are fixed, while ChronoSteer undergoes three independent runs.

	\item In the second experiment, ChronoSteer remains constant, and QwQ-32B is invoked three times independently to generate different revision instructions from the same textual events.

	\item In the third experiment, neither the revision instructions derived from textual events nor ChronoSteer is fixed, enabling three fully independent runs.

\end{itemize}

The results in Table \ref{tab10} reveal that when the revision instructions are fixed, ChronoSteer demonstrates strong robustness, with a minimal standard deviation indicating high output consistency. Furthermore, due to its fine-tuning on observed future series, which enables instruction correction based on historical series (as detailed in Section \ref{subsection:Prediction_Showcase}), the system maintains low error even when revision instructions vary. This highlights its consistency and stability.

\section{Discussion and Conclusion}
\label{section:Conclusion}

In this work, we propose ChronoSteer, a agentic multimodal forecasting framework that integrates the complementary strengths of TSFMs and LLMs for zero-shot future inference using both temporal signals and textual event descriptions. ChronoSteer adopts a decoupled design, in which a pretrained LLM translates textual events into revision instructions that steer the predictions of a frozen TSFM, thereby bridging the semantic gap between text and time series while effectively leveraging pretrained knowledge. To address the scarcity of paired alignment data, we develop a two-stage training pipeline built on a synthetic cross-modal dataset. It first performs coarse anchor-based alignment and then refines fine-grained magnitude details. We also introduce MTSFBench-300, a leakage-controlled benchmark to support more transparent evaluation of multimodal forecasting models.

Despite these encouraging results, ChronoSteer still has limitations on both sides of the pipeline, namely event-to-instruction translation and instruction-to-series forecasting. On the LLM side, when textual events are ambiguous, underspecified, or only weakly correlated with the target series, the resulting instructions may be insufficiently informative or even misleading, thereby limiting the gains over the unimodal TSFM. Improving the validity and reliability of generated instructions therefore remains an important direction. Prompt refinement, candidate instruction reranking, specialized small models, and recent advances in agent-based reasoning \cite{MoiraiAgent} may serve as useful complements for event understanding and instruction generation.

On the forecasting side, the current codebook provides a practical low-resource interface between textual events and temporal revisions, but its discrete design cannot fully represent complex real-world event dynamics. Discretization still constrains open-domain generalization, and the current synthetic functions cannot fully describe nonlinear, delayed, continuous, compound, or domain-specific event effects. The two-stage training strategy recovers fine-grained magnitude information from historical context, but more flexible instruction-based forecasting will likely require higher-quality paired training data, whether synthetic or real.

For evaluation coverage, MTSFBench-300 provides carefully curated samples across three representative domains, but its scale remains limited. Experiments with varying historical and prediction windows, repeated runs, and paired significance tests improve the statistical reliability of the evaluation, while broader leakage-controlled benchmarks are still needed to support stronger external validity.

ChronoSteer can also be extended beyond the current univariate setting. Most existing TSFMs are univariate, and methods such as AdaPTS \cite{AdaPTS} adapt them to multivariate forecasting through lightweight adapters, suggesting a similar route for ChronoSteer in future work. For richer multimodal inputs, external signals such as structured metadata, calendar information, and visual observations can first be converted into revision instructions. A more flexible instruction space may further support such inputs once higher-quality paired resources become available.

Overall, ChronoSteer offers a useful foundation for MTSF at the current stage of the field. Further progress will likely depend on higher-quality paired training data and more rigorous evaluation benchmarks.



\section*{Acknowledgments}
\label{section:Acknowledgments}

This work was supported in part by the National Key R\&D Program of China (Grant No. 2024YFE0200800), the National Natural Science Foundation of China under Grants (62401080, 62471055, U23B2001, 62321001, 62201072), the Fundamental and Interdisciplinary Disciplines Breakthrough Plan of the Ministry of Education of China(JYB2025XDXM107), the High-Quality Development Project of the MIIT(2440STCZB2584), the Ministry of Education and China Mobile Joint Fund (MCM20200202, MCM20180101), Supported by State Key Lab of Processors, Institute of Computing Technology, CAS under Grant No.CLQ202604, the Fundamental Research Funds for the Central Universities.

\bibliographystyle{ACM-Reference-Format}
\bibliography{main}


\newpage
\appendix

\section{Detailed Synthetic Function}
\label{appendix:Detailed_Synthetic_Function}

\begin{algorithm}[tp]
	\centering
	\caption{The synthetic function for revision instruction anchors.}
	\label{alg5}
	\definecolor{codeblue}{rgb}{0.25,0.5,0.5}
	\lstset{
		backgroundcolor=\color{white},
		basicstyle=\fontsize{8.5pt}{8.5pt}\ttfamily\selectfont,
		columns=fullflexible,
		breaklines=true,
		captionpos=b,
		commentstyle=\fontsize{8.5pt}{8.5pt}\color{codeblue},
		keywordstyle=\fontsize{8.5pt}{8.5pt}\color{codeblue},
	}
	\begin{lstlisting}[language=python]
    import numpy as np
    from sklearn.preprocessing import MinMaxScaler

    def normalize(series):
        scaler = MinMaxScaler()
        normed = (scaler.fit_transform(series.reshape(-1, 1)).reshape(-1) - 0.5) * 2
        return normed, scaler

    def denormalize(series, scaler):
        return scaler.inverse_transform(series.reshape(-1, 1) / 2 + 0.5).reshape(-1)

    def tanh_curve(length, factor, n=8):
        return np.tanh(np.linspace(0, n, length)) * factor

    def keep_unchanged(series):
        return series

    def increase_trend(series, min_factor=0.2, max_factor=0.8):
        series, scaler = normalize(series)
        trend = tanh_curve(len(series), np.random.uniform(min_factor, max_factor))
        return denormalize(series + trend, scaler)

    def reduce_trend(series, min_factor=0.2, max_factor=0.8):
        series, scaler = normalize(series)
        trend = tanh_curve(len(series), np.random.uniform(min_factor, max_factor))
        return denormalize(series - trend, scaler)

    def expand_amplitude(series, min_factor=0.2, max_factor=0.8):
        series, scaler = normalize(series)
        factor = tanh_curve(len(series), np.random.uniform(min_factor, max_factor))
        return denormalize(series * (1 + factor), scaler)

    def compress_amplitude(series, min_factor=0.2, max_factor=0.8):
        series, scaler = normalize(series)
        factor = tanh_curve(len(series), np.random.uniform(min_factor, max_factor))
        return denormalize(series * (1 - factor), scaler)

    def elevate_peaks(series, min_factor=0.1, max_factor=0.4):
        series, scaler = normalize(series)
        amp = np.random.uniform(min_factor, max_factor)
        return denormalize((series + 1) * (1 + amp) - 1, scaler)

    def lower_peaks(series, min_factor=0.1, max_factor=0.4):
        series, scaler = normalize(series)
        amp = np.random.uniform(min_factor, max_factor)
        return denormalize((series + 1) * (1 - amp) - 1, scaler)

    def raise_troughs(series, min_factor=0.1, max_factor=0.4):
        series, scaler = normalize(series)
        amp = np.random.uniform(min_factor, max_factor)
        return denormalize((series - 1) * (1 - amp) + 1, scaler)

    def deepen_troughs(series, min_factor=0.1, max_factor=0.4):
        series, scaler = normalize(series)
        amp = np.random.uniform(min_factor, max_factor)
        return denormalize((series - 1) * (1 + amp) + 1, scaler)
\end{lstlisting}
\end{algorithm}

By applying cluster sampling and anchor extraction to the Monash Repository, we obtain 100K high-quality time series slices and identify nine core anchors that represent discrete prototypes of temporal adjustment behaviors. For each anchor, we design a corresponding transformation function, as detailed in Algorithm \ref{alg5}, where the specific perturbation distributions are defined within the default parameters. These functions are subsequently applied to the 100K slices to yield the ChronoSteer-100K-PT dataset, comprising 900K precisely aligned synthetic triplets $\left( \mathbf{x}, \mathbf{c}, \tilde{\mathbf{y}} \right)$.

To examine whether the parameter ranges used in the synthetic transformation functions are overly heuristic, we conduct a compact sensitivity analysis on the perturbation strength by following the protocol in Section \ref{subsection:Hyperparameter_Analysis}. Starting from the default perturbation ranges in Algorithm \ref{alg5}, we introduce a global scaling factor $\beta \in \{0.5, 1.0, 2.0\}$ to construct three settings: \textit{conservative}, \textit{default}, and \textit{aggressive}. This design preserves the relative semantics of each anchor while testing whether ChronoSteer depends on a narrowly tuned perturbation magnitude.

The results in Table \ref{tab:range_sensitivity} indicate that the default ranges achieve the best overall performance, while both the conservative and aggressive variants remain competitive. This behavior is consistent with the intended role of the synthetic functions. When $\beta$ is too small, the synthesized targets under-adjust the initial forecast and weaken the alignment signal. When $\beta$ is too large, the perturbations become less faithful to realistic revision magnitudes and may over-emphasize local adjustments. Nevertheless, the degradation under both alternatives is limited, suggesting that ChronoSteer is robust to moderate range changes rather than relying on a brittle hand-tuned setting. Overall, the default parameter ranges provide the stable trade-off between controllability and realism.

\section{Detailed Ablation Study}
\label{appendix:Detailed_Ablation_Study}

\subsection{Ablation Study on Large Language Model}
\label{subappendix:Ablation_Study_on_Large_Language_Model}

In the primary experiment, the LLM QwQ-32B is employed to transform textual events into revision instructions. To assess how different LLMs affect the prediction accuracy of the system, this section replaces QwQ-32B with alternative models for comparison. Specifically, four models are tested: standard LLM Qwen2.5-32B, reasoning-enhanced LLM QwQ-32B, standard vLLM Qwen2.5-32B-VL, and reasoning-enhanced vLLM QVQ-72B-Preview.

As shown in Table \ref{tab11}, the reasoning-enhanced LLM QwQ-32B demonstrates significantly higher instruction transformation accuracy than the standard LLM Qwen2.5-32B, particularly on unstable datasets. Improvements in prediction accuracy reach 2.1\%, 3.5\%, and 18.5\% on the Energy, Traffic, and Weather datasets, respectively.

Previous research indicates that LLMs struggle to comprehend temporal characteristics effectively. Representing time series as images and leveraging vLLMs can better capture complex temporal patterns. Based on this, we introduce the standard vLLM Qwen2.5-32B-VL and the reasoning-enhanced vLLM QVQ-72B-Preview for comparative experiments. Results show that the standard vLLM Qwen2.5-32B-VL achieves competitive performance, closely matching the reasoning-enhanced LLM QwQ-32B. For instance, on the Energy dataset, its prediction accuracy outperforms QwQ-32B by 2.7\%, while on the Traffic dataset, it trails by only 3.5\%. However, the accuracy of Qwen2.5-32B-VL on the unstable Weather dataset is 17.8\% lower than QwQ-32B. This may be due to cross-modal fine-tuning reducing textual reasoning capability. Despite this, it still outperforms the standard LLM Qwen2.5-32B by 3.5\%. Surprisingly, the reasoning-enhanced vLLM QVQ-72B-Preview performs the worst across all datasets. This underperformance may stem from the preview version being in an optimization phase, where insufficient stability leads to suboptimal results.

\begin{table}[tp]
	\centering
	\caption{The prediction errors of ChronoSteer with different parameter ranges in the synthetic transformation functions. The historical windows vary across \{2, 3, 4, 7\} days, while the prediction window is fixed at 1 day. A lower value denotes superior accuracy.}
	\resizebox{0.8\linewidth}{!}
	{
		\begin{tabular}{@{}cc|ll|ll|ll|ll@{}}
	\toprule
	\multicolumn{2}{c|}{\multirow{2}{*}{Method}}                  & \multicolumn{2}{c|}{Unimodal} & \multicolumn{2}{c|}{Conservative (0.5$\times$)} & \multicolumn{2}{c|}{Default (1.0$\times$)} & \multicolumn{2}{c}{Aggressive (2.0$\times$)}                                                                                                                            \\ \cmidrule(l){3-10}
	\multicolumn{2}{c|}{}                                         & \multicolumn{1}{c}{MSE}       & \multicolumn{1}{c|}{MAE}                        & \multicolumn{1}{c}{MSE}                    & \multicolumn{1}{c|}{MAE}                     & \multicolumn{1}{c}{MSE} & \multicolumn{1}{c|}{MAE} & \multicolumn{1}{c}{MSE} & \multicolumn{1}{c}{MAE}                   \\ \midrule
	\multicolumn{1}{c|}{\multirow{4}{*}{\rotatebox{90}{Energy}}}  & 2D                            & 0.765$\pm$0.000                                 & 0.277$\pm$0.000                            & 0.583$\pm$0.004                              & 0.266$\pm$0.005         & 0.573$\pm$0.012          & 0.265$\pm$0.007         & 0.586$\pm$0.007         & 0.266$\pm$0.009 \\
	\multicolumn{1}{c|}{}                                         & 3D                            & 0.763$\pm$0.000                                 & 0.268$\pm$0.000                            & 0.600$\pm$0.013                              & 0.267$\pm$0.013         & 0.591$\pm$0.008          & 0.267$\pm$0.004         & 0.601$\pm$0.010         & 0.267$\pm$0.009 \\
	\multicolumn{1}{c|}{}                                         & 4D                            & 0.697$\pm$0.000                                 & 0.259$\pm$0.000                            & 0.618$\pm$0.016                              & 0.254$\pm$0.017         & 0.613$\pm$0.011          & 0.253$\pm$0.005         & 0.617$\pm$0.011         & 0.253$\pm$0.016 \\
	\multicolumn{1}{c|}{}                                         & 7D                            & 0.573$\pm$0.000                                 & 0.246$\pm$0.000                            & 0.527$\pm$0.015                              & 0.243$\pm$0.006         & 0.524$\pm$0.006          & 0.243$\pm$0.005         & 0.526$\pm$0.011         & 0.243$\pm$0.006 \\ \midrule
	\multicolumn{1}{c|}{\multirow{4}{*}{\rotatebox{90}{Traffic}}} & 2D                            & 0.669$\pm$0.000                                 & 0.571$\pm$0.000                            & 0.522$\pm$0.013                              & 0.512$\pm$0.003         & 0.512$\pm$0.005          & 0.508$\pm$0.013         & 0.520$\pm$0.006         & 0.512$\pm$0.011 \\
	\multicolumn{1}{c|}{}                                         & 3D                            & 0.608$\pm$0.000                                 & 0.551$\pm$0.000                            & 0.420$\pm$0.011                              & 0.457$\pm$0.012         & 0.401$\pm$0.007          & 0.448$\pm$0.006         & 0.411$\pm$0.013         & 0.453$\pm$0.010 \\
	\multicolumn{1}{c|}{}                                         & 4D                            & 0.559$\pm$0.000                                 & 0.522$\pm$0.000                            & 0.381$\pm$0.020                              & 0.438$\pm$0.005         & 0.367$\pm$0.006          & 0.430$\pm$0.007         & 0.376$\pm$0.015         & 0.440$\pm$0.002 \\
	\multicolumn{1}{c|}{}                                         & 7D                            & 0.344$\pm$0.000                                 & 0.390$\pm$0.000                            & 0.251$\pm$0.007                              & 0.336$\pm$0.015         & 0.242$\pm$0.004          & 0.332$\pm$0.006         & 0.248$\pm$0.021         & 0.336$\pm$0.013 \\ \midrule
	\multicolumn{1}{c|}{\multirow{4}{*}{\rotatebox{90}{Weather}}} & 2D                            & 2.340$\pm$0.000                                 & 1.126$\pm$0.000                            & 1.572$\pm$0.008                              & 0.926$\pm$0.013         & 1.532$\pm$0.029          & 0.910$\pm$0.019         & 1.585$\pm$0.010         & 0.923$\pm$0.013 \\
	\multicolumn{1}{c|}{}                                         & 3D                            & 1.511$\pm$0.000                                 & 0.967$\pm$0.000                            & 0.950$\pm$0.009                              & 0.741$\pm$0.003         & 0.898$\pm$0.015          & 0.726$\pm$0.014         & 0.929$\pm$0.005         & 0.744$\pm$0.005 \\
	\multicolumn{1}{c|}{}                                         & 4D                            & 1.358$\pm$0.000                                 & 0.909$\pm$0.000                            & 0.855$\pm$0.023                              & 0.693$\pm$0.008         & 0.800$\pm$0.014          & 0.673$\pm$0.008         & 0.822$\pm$0.014         & 0.683$\pm$0.004 \\
	\multicolumn{1}{c|}{}                                         & 7D                            & 1.116$\pm$0.000                                 & 0.822$\pm$0.000                            & 0.782$\pm$0.016                              & 0.662$\pm$0.012         & 0.754$\pm$0.021          & 0.646$\pm$0.012         & 0.776$\pm$0.020         & 0.654$\pm$0.012 \\ \bottomrule
\end{tabular}
	}
	\label{tab:range_sensitivity}
\end{table}

\begin{table}[tp]
	\centering
	\caption{The prediction errors of ChronoSteer across various large language models. The historical windows vary across \{2, 3, 4, 7\} days, while the prediction window is fixed at 1 day. A lower value denotes superior accuracy.}
	\resizebox{0.95\linewidth}{!}
	{
		\begin{tabular}{@{}cc|cc|cc|cc|cc|cc@{}}
	\toprule
	\multicolumn{2}{c|}{\multirow{2}{*}{Method}}                  & \multicolumn{2}{c|}{Unimodal} & \multicolumn{2}{c|}{Qwen2.5-32B} & \multicolumn{2}{c|}{QwQ-32B} & \multicolumn{2}{c|}{Qwen2.5-32B-VL} & \multicolumn{2}{c}{QVQ-72B}                                                                                                             \\ \cmidrule(l){3-12}
	\multicolumn{2}{c|}{}                                         & \multicolumn{2}{c|}{MSE}      & \multicolumn{2}{c|}{MSE}         & \multicolumn{2}{c|}{MSE}     & \multicolumn{2}{c|}{MSE}            & MSE                         & MAE                                                                                                       \\ \midrule
	\multicolumn{1}{c|}{\multirow{4}{*}{\rotatebox{90}{Energy}}}  & 2D                            & 0.765$\pm$0.000                  & 0.277$\pm$0.000              & 0.604$\pm$0.009                     & 0.267$\pm$0.007             & 0.573$\pm$0.012 & 0.265$\pm$0.007 & 0.584$\pm$0.015 & 0.266$\pm$0.006 & 0.723$\pm$0.016 & 0.272$\pm$0.005 \\
	\multicolumn{1}{c|}{}                                         & 3D                            & 0.763$\pm$0.000                  & 0.268$\pm$0.000              & 0.606$\pm$0.011                     & 0.267$\pm$0.005             & 0.591$\pm$0.008 & 0.267$\pm$0.004 & 0.558$\pm$0.011 & 0.260$\pm$0.005 & 0.716$\pm$0.013 & 0.268$\pm$0.004 \\
	\multicolumn{1}{c|}{}                                         & 4D                            & 0.697$\pm$0.000                  & 0.259$\pm$0.000              & 0.609$\pm$0.017                     & 0.258$\pm$0.009             & 0.613$\pm$0.011 & 0.253$\pm$0.005 & 0.545$\pm$0.015 & 0.248$\pm$0.003 & 0.666$\pm$0.011 & 0.258$\pm$0.004 \\
	\multicolumn{1}{c|}{}                                         & 7D                            & 0.573$\pm$0.000                  & 0.246$\pm$0.000              & 0.546$\pm$0.011                     & 0.244$\pm$0.004             & 0.524$\pm$0.006 & 0.243$\pm$0.005 & 0.533$\pm$0.016 & 0.244$\pm$0.004 & 0.562$\pm$0.007 & 0.243$\pm$0.004 \\ \midrule
	\multicolumn{1}{c|}{\multirow{4}{*}{\rotatebox{90}{Traffic}}} & 2D                            & 0.669$\pm$0.000                  & 0.571$\pm$0.000              & 0.520$\pm$0.008                     & 0.508$\pm$0.013             & 0.512$\pm$0.005 & 0.508$\pm$0.013 & 0.523$\pm$0.011 & 0.513$\pm$0.014 & 0.558$\pm$0.009 & 0.526$\pm$0.011 \\
	\multicolumn{1}{c|}{}                                         & 3D                            & 0.608$\pm$0.000                  & 0.551$\pm$0.000              & 0.439$\pm$0.010                     & 0.453$\pm$0.008             & 0.401$\pm$0.007 & 0.448$\pm$0.006 & 0.387$\pm$0.007 & 0.436$\pm$0.008 & 0.415$\pm$0.003 & 0.448$\pm$0.008 \\
	\multicolumn{1}{c|}{}                                         & 4D                            & 0.559$\pm$0.000                  & 0.522$\pm$0.000              & 0.374$\pm$0.010                     & 0.435$\pm$0.009             & 0.367$\pm$0.006 & 0.430$\pm$0.007 & 0.363$\pm$0.007 & 0.425$\pm$0.009 & 0.389$\pm$0.007 & 0.438$\pm$0.008 \\
	\multicolumn{1}{c|}{}                                         & 7D                            & 0.344$\pm$0.000                  & 0.390$\pm$0.000              & 0.287$\pm$0.006                     & 0.342$\pm$0.010             & 0.242$\pm$0.004 & 0.332$\pm$0.006 & 0.327$\pm$0.008 & 0.380$\pm$0.006 & 0.333$\pm$0.009 & 0.387$\pm$0.007 \\ \midrule
	\multicolumn{1}{c|}{\multirow{4}{*}{\rotatebox{90}{Weather}}} & 2D                            & 2.340$\pm$0.000                  & 1.126$\pm$0.000              & 1.848$\pm$0.058                     & 1.053$\pm$0.026             & 1.532$\pm$0.029 & 0.910$\pm$0.019 & 1.948$\pm$0.026 & 1.080$\pm$0.035 & 1.820$\pm$0.032 & 1.077$\pm$0.025 \\
	\multicolumn{1}{c|}{}                                         & 3D                            & 1.511$\pm$0.000                  & 0.967$\pm$0.000              & 1.159$\pm$0.026                     & 0.817$\pm$0.016             & 0.898$\pm$0.015 & 0.726$\pm$0.014 & 1.082$\pm$0.011 & 0.820$\pm$0.025 & 1.167$\pm$0.019 & 0.812$\pm$0.017 \\
	\multicolumn{1}{c|}{}                                         & 4D                            & 1.358$\pm$0.000                  & 0.909$\pm$0.000              & 1.049$\pm$0.012                     & 0.810$\pm$0.015             & 0.800$\pm$0.014 & 0.673$\pm$0.008 & 0.921$\pm$0.026 & 0.741$\pm$0.015 & 0.976$\pm$0.018 & 0.753$\pm$0.007 \\
	\multicolumn{1}{c|}{}                                         & 7D                            & 1.116$\pm$0.000                  & 0.822$\pm$0.000              & 1.004$\pm$0.019                     & 0.779$\pm$0.011             & 0.754$\pm$0.021 & 0.646$\pm$0.012 & 0.864$\pm$0.015 & 0.716$\pm$0.020 & 0.930$\pm$0.017 & 0.736$\pm$0.014 \\ \bottomrule
\end{tabular}
	}
	\label{tab11}
\end{table}

\begin{table}[tp]
	\centering
	\caption{The prediction errors of ChronoSteer across various text embedding models. The historical windows vary across \{2, 3, 4, 7\} days, while the prediction window is fixed at 1 day. A lower value denotes superior accuracy.}
	\resizebox{\linewidth}{!}
	{
		\begin{tabular}{@{}cc|cc|cc|cc|cc|cc@{}}
	\toprule
	\multicolumn{2}{c|}{\multirow{3}{*}{Method}}                  & \multicolumn{2}{c|}{\multirow{2}{*}{Unimodal}} & \multicolumn{2}{c|}{all-MiniLM-L6-v2} & \multicolumn{2}{c|}{voyage-3-lite} & \multicolumn{2}{c|}{bge-m3}     & \multicolumn{2}{c}{NV-Embed-v2}                                                                                                             \\
	\multicolumn{2}{c|}{}                                         & \multicolumn{2}{c|}{}                          & \multicolumn{2}{c|}{(Dim 384)}        & \multicolumn{2}{c|}{(Dim 512)}     & \multicolumn{2}{c|}{(Dim 1024)} & \multicolumn{2}{c}{(Dim 4096)}                                                                                                              \\ \cmidrule(l){3-12}
	\multicolumn{2}{c|}{}                                         & MSE                                            & MAE                                   & MSE                                & MAE                             & MSE                             & MAE             & MSE             & MAE             & MSE             & MAE                               \\ \midrule
	\multicolumn{1}{c|}{\multirow{4}{*}{\rotatebox{90}{Energy}}}  & 2D                                             & 0.765$\pm$0.000                       & 0.277$\pm$0.000                    & 0.747$\pm$0.014                 & 0.284$\pm$0.005                 & 0.685$\pm$0.008 & 0.295$\pm$0.004 & 0.573$\pm$0.012 & 0.265$\pm$0.007 & 0.601$\pm$0.011 & 0.288$\pm$0.003 \\
	\multicolumn{1}{c|}{}                                         & 3D                                             & 0.763$\pm$0.000                       & 0.268$\pm$0.000                    & 0.648$\pm$0.012                 & 0.273$\pm$0.008                 & 0.689$\pm$0.014 & 0.298$\pm$0.004 & 0.591$\pm$0.008 & 0.267$\pm$0.004 & 0.606$\pm$0.014 & 0.285$\pm$0.006 \\
	\multicolumn{1}{c|}{}                                         & 4D                                             & 0.697$\pm$0.000                       & 0.259$\pm$0.000                    & 0.624$\pm$0.009                 & 0.261$\pm$0.005                 & 0.571$\pm$0.014 & 0.266$\pm$0.005 & 0.613$\pm$0.011 & 0.253$\pm$0.005 & 0.582$\pm$0.012 & 0.270$\pm$0.005 \\
	\multicolumn{1}{c|}{}                                         & 7D                                             & 0.573$\pm$0.000                       & 0.246$\pm$0.000                    & 0.524$\pm$0.009                 & 0.246$\pm$0.003                 & 0.559$\pm$0.008 & 0.260$\pm$0.004 & 0.524$\pm$0.006 & 0.243$\pm$0.005 & 0.536$\pm$0.009 & 0.251$\pm$0.005 \\ \midrule
	\multicolumn{1}{c|}{\multirow{4}{*}{\rotatebox{90}{Traffic}}} & 2D                                             & 0.669$\pm$0.000                       & 0.571$\pm$0.000                    & 0.769$\pm$0.009                 & 0.631$\pm$0.011                 & 0.823$\pm$0.010 & 0.664$\pm$0.013 & 0.512$\pm$0.005 & 0.508$\pm$0.013 & 0.445$\pm$0.008 & 0.496$\pm$0.007 \\
	\multicolumn{1}{c|}{}                                         & 3D                                             & 0.608$\pm$0.000                       & 0.551$\pm$0.000                    & 0.586$\pm$0.008                 & 0.547$\pm$0.012                 & 0.451$\pm$0.007 & 0.478$\pm$0.007 & 0.401$\pm$0.007 & 0.448$\pm$0.006 & 0.438$\pm$0.010 & 0.470$\pm$0.007 \\
	\multicolumn{1}{c|}{}                                         & 4D                                             & 0.559$\pm$0.000                       & 0.522$\pm$0.000                    & 0.514$\pm$0.011                 & 0.522$\pm$0.010                 & 0.515$\pm$0.008 & 0.531$\pm$0.007 & 0.367$\pm$0.006 & 0.430$\pm$0.007 & 0.390$\pm$0.006 & 0.447$\pm$0.006 \\
	\multicolumn{1}{c|}{}                                         & 7D                                             & 0.344$\pm$0.000                       & 0.390$\pm$0.000                    & 0.301$\pm$0.004                 & 0.388$\pm$0.007                 & 0.290$\pm$0.005 & 0.364$\pm$0.007 & 0.242$\pm$0.004 & 0.332$\pm$0.006 & 0.255$\pm$0.004 & 0.347$\pm$0.007 \\ \midrule
	\multicolumn{1}{c|}{\multirow{4}{*}{\rotatebox{90}{Weather}}} & 2D                                             & 2.340$\pm$0.000                       & 1.126$\pm$0.000                    & 2.272$\pm$0.038                 & 1.140$\pm$0.021                 & 1.496$\pm$0.023 & 0.924$\pm$0.010 & 1.532$\pm$0.029 & 0.910$\pm$0.019 & 1.477$\pm$0.015 & 0.907$\pm$0.017 \\
	\multicolumn{1}{c|}{}                                         & 3D                                             & 1.511$\pm$0.000                       & 0.967$\pm$0.000                    & 1.434$\pm$0.020                 & 0.891$\pm$0.019                 & 1.134$\pm$0.025 & 0.824$\pm$0.017 & 0.898$\pm$0.015 & 0.726$\pm$0.014 & 1.065$\pm$0.024 & 0.775$\pm$0.016 \\
	\multicolumn{1}{c|}{}                                         & 4D                                             & 1.358$\pm$0.000                       & 0.909$\pm$0.000                    & 1.315$\pm$0.034                 & 0.860$\pm$0.015                 & 0.960$\pm$0.019 & 0.743$\pm$0.014 & 0.800$\pm$0.014 & 0.673$\pm$0.008 & 0.830$\pm$0.011 & 0.681$\pm$0.015 \\
	\multicolumn{1}{c|}{}                                         & 7D                                             & 1.116$\pm$0.000                       & 0.822$\pm$0.000                    & 1.026$\pm$0.026                 & 0.759$\pm$0.014                 & 0.874$\pm$0.011 & 0.716$\pm$0.016 & 0.754$\pm$0.021 & 0.646$\pm$0.012 & 0.849$\pm$0.011 & 0.695$\pm$0.013 \\ \bottomrule
\end{tabular}
	}
	\label{tab12}
\end{table}

\subsection{Ablation Study on Embedding Model}
\label{subappendix:Ablation_Study_on_Embedding_Model}

In the primary experiment, the text embedding model BGE-M3 is employed to transform textual revision instructions into high-dimension embedding vectors. To examine the impact of different text embedding models on system prediction accuracy, BGE-M3 is replaced with several widely used alternatives. Specifically, four models are selected: all-MiniLM-L6-v2, Voyage-3-lite, BGE-M3, and NV-Embed-v2. Based on the MTEB Leaderboard, these models show a progressive increase in representational capability.

As shown in Table \ref{tab12}, prediction accuracy improves incrementally with enhanced embedding capability. However, despite NV-Embed-v2 having nearly ten times the parameters of BGE-M3, its performance does not significantly surpass that of BGE-M3. This may be due to the high dimension of NV-Embed-v2, which tends to semantic dispersion. Given its larger embedding size, the current alignment dataset may be insufficient to effectively train an alignment module of this scale, limiting the potential of the model.

\subsection{Ablation Study on Time series Foundation Model}
\label{subappendix:Ablation_Study_on_Time_series_Foundation_Model}

\begin{table}[t]
	\centering
	\caption{The prediction errors of ChronoSteer across various time series foundation models from the same families. The historical windows vary across \{2, 3, 4, 7\} days, while prediction window is fixed at 1. A lower value denotes superior accuracy.}
	\resizebox{0.9\linewidth}{!}
	{
		\begin{tabular}{@{}ccc|cc|cc|cc|cc@{}}
	\toprule
	\multicolumn{3}{c|}{\multirow{2}{*}{Method}}                  & \multicolumn{2}{c|}{chronos-bolt-tiny}   & \multicolumn{2}{c|}{chronos-bolt-mini} & \multicolumn{2}{c|}{chronos-bolt-small} & \multicolumn{2}{c}{chronos-bolt-base}                                                                                                                   \\ \cmidrule(l){4-11}
	\multicolumn{3}{c|}{}                                         & Unimodal                                 & Multimodal                             & Unimodal                                & Multimodal                            & Unimodal        & Multimodal        & Unimodal        & Multimodal                                              \\ \midrule
	\multicolumn{1}{c|}{\multirow{8}{*}{\rotatebox{90}{Energy}}}  & \multicolumn{1}{c|}{\multirow{2}{*}{2D}} & MSE                                    & 0.653$\pm$0.000                         & {0.629$\pm$0.012}                     & 0.659$\pm$0.000 & {0.635$\pm$0.007} & 0.803$\pm$0.000 & {0.718$\pm$0.012} & 0.765$\pm$0.000 & {0.573$\pm$0.012} \\
	\multicolumn{1}{c|}{}                                         & \multicolumn{1}{c|}{}                    & MAE                                    & 0.265$\pm$0.000                         & {0.259$\pm$0.004}                     & 0.259$\pm$0.000 & {0.250$\pm$0.004} & 0.274$\pm$0.000 & {0.272$\pm$0.003} & 0.277$\pm$0.000 & {0.265$\pm$0.007} \\ \cmidrule(l){2-11}
	\multicolumn{1}{c|}{}                                         & \multicolumn{1}{c|}{\multirow{2}{*}{3D}} & MSE                                    & 0.665$\pm$0.000                         & {0.608$\pm$0.006}                     & 0.686$\pm$0.000 & {0.600$\pm$0.010} & 0.775$\pm$0.000 & {0.689$\pm$0.009} & 0.763$\pm$0.000 & {0.591$\pm$0.008} \\
	\multicolumn{1}{c|}{}                                         & \multicolumn{1}{c|}{}                    & MAE                                    & 0.262$\pm$0.000                         & {0.256$\pm$0.004}                     & 0.260$\pm$0.000 & {0.255$\pm$0.004} & 0.262$\pm$0.000 & {0.259$\pm$0.005} & 0.268$\pm$0.000 & {0.267$\pm$0.004} \\ \cmidrule(l){2-11}
	\multicolumn{1}{c|}{}                                         & \multicolumn{1}{c|}{\multirow{2}{*}{4D}} & MSE                                    & 0.611$\pm$0.000                         & {0.567$\pm$0.009}                     & 0.670$\pm$0.000 & {0.561$\pm$0.007} & 0.802$\pm$0.000 & {0.645$\pm$0.011} & 0.697$\pm$0.000 & {0.613$\pm$0.011} \\
	\multicolumn{1}{c|}{}                                         & \multicolumn{1}{c|}{}                    & MAE                                    & 0.256$\pm$0.000                         & {0.251$\pm$0.005}                     & 0.256$\pm$0.000 & {0.251$\pm$0.005} & 0.261$\pm$0.000 & {0.259$\pm$0.003} & 0.259$\pm$0.000 & {0.253$\pm$0.005} \\ \cmidrule(l){2-11}
	\multicolumn{1}{c|}{}                                         & \multicolumn{1}{c|}{\multirow{2}{*}{7D}} & MSE                                    & 0.578$\pm$0.000                         & {0.526$\pm$0.009}                     & 0.645$\pm$0.000 & {0.516$\pm$0.009} & 0.768$\pm$0.000 & {0.657$\pm$0.010} & 0.573$\pm$0.000 & {0.524$\pm$0.006} \\
	\multicolumn{1}{c|}{}                                         & \multicolumn{1}{c|}{}                    & MAE                                    & 0.250$\pm$0.000                         & {0.245$\pm$0.005}                     & 0.255$\pm$0.000 & {0.244$\pm$0.006} & 0.255$\pm$0.000 & {0.252$\pm$0.004} & 0.246$\pm$0.000 & {0.243$\pm$0.005} \\ \midrule
	\multicolumn{1}{c|}{\multirow{8}{*}{\rotatebox{90}{Traffic}}} & \multicolumn{1}{c|}{\multirow{2}{*}{2D}} & MSE                                    & 0.632$\pm$0.000                         & {0.466$\pm$0.009}                     & 0.662$\pm$0.000 & {0.555$\pm$0.007} & 0.685$\pm$0.000 & {0.574$\pm$0.010} & 0.669$\pm$0.000 & {0.512$\pm$0.005} \\
	\multicolumn{1}{c|}{}                                         & \multicolumn{1}{c|}{}                    & MAE                                    & 0.559$\pm$0.000                         & {0.484$\pm$0.005}                     & 0.577$\pm$0.000 & {0.567$\pm$0.011} & 0.584$\pm$0.000 & {0.520$\pm$0.009} & 0.571$\pm$0.000 & {0.508$\pm$0.013} \\ \cmidrule(l){2-11}
	\multicolumn{1}{c|}{}                                         & \multicolumn{1}{c|}{\multirow{2}{*}{3D}} & MSE                                    & 0.620$\pm$0.000                         & {0.467$\pm$0.005}                     & 0.605$\pm$0.000 & {0.381$\pm$0.006} & 0.613$\pm$0.000 & {0.411$\pm$0.010} & 0.608$\pm$0.000 & {0.401$\pm$0.007} \\
	\multicolumn{1}{c|}{}                                         & \multicolumn{1}{c|}{}                    & MAE                                    & 0.552$\pm$0.000                         & {0.481$\pm$0.005}                     & 0.554$\pm$0.000 & {0.447$\pm$0.007} & 0.547$\pm$0.000 & {0.476$\pm$0.013} & 0.551$\pm$0.000 & {0.448$\pm$0.006} \\ \cmidrule(l){2-11}
	\multicolumn{1}{c|}{}                                         & \multicolumn{1}{c|}{\multirow{2}{*}{4D}} & MSE                                    & 0.596$\pm$0.000                         & {0.391$\pm$0.007}                     & 0.564$\pm$0.000 & {0.382$\pm$0.009} & 0.593$\pm$0.000 & {0.423$\pm$0.008} & 0.559$\pm$0.000 & {0.367$\pm$0.006} \\
	\multicolumn{1}{c|}{}                                         & \multicolumn{1}{c|}{}                    & MAE                                    & 0.538$\pm$0.000                         & {0.430$\pm$0.006}                     & 0.526$\pm$0.000 & {0.461$\pm$0.007} & 0.526$\pm$0.000 & {0.492$\pm$0.010} & 0.522$\pm$0.000 & {0.430$\pm$0.007} \\ \cmidrule(l){2-11}
	\multicolumn{1}{c|}{}                                         & \multicolumn{1}{c|}{\multirow{2}{*}{7D}} & MSE                                    & 0.351$\pm$0.000                         & {0.299$\pm$0.005}                     & 0.353$\pm$0.000 & {0.247$\pm$0.003} & 0.351$\pm$0.000 & {0.256$\pm$0.003} & 0.344$\pm$0.000 & {0.242$\pm$0.004} \\
	\multicolumn{1}{c|}{}                                         & \multicolumn{1}{c|}{}                    & MAE                                    & 0.403$\pm$0.000                         & {0.380$\pm$0.006}                     & 0.399$\pm$0.000 & {0.344$\pm$0.004} & 0.397$\pm$0.000 & {0.360$\pm$0.008} & 0.390$\pm$0.000 & {0.332$\pm$0.006} \\ \midrule
	\multicolumn{1}{c|}{\multirow{8}{*}{\rotatebox{90}{Weather}}} & \multicolumn{1}{c|}{\multirow{2}{*}{2D}} & MSE                                    & 2.027$\pm$0.000                         & {1.616$\pm$0.022}                     & 2.246$\pm$0.000 & {1.630$\pm$0.029} & 2.137$\pm$0.000 & {1.585$\pm$0.036} & 2.340$\pm$0.000 & {1.532$\pm$0.029} \\
	\multicolumn{1}{c|}{}                                         & \multicolumn{1}{c|}{}                    & MAE                                    & 1.081$\pm$0.000                         & {0.964$\pm$0.013}                     & 1.141$\pm$0.000 & {0.943$\pm$0.014} & 1.116$\pm$0.000 & {0.940$\pm$0.020} & 1.126$\pm$0.000 & {0.910$\pm$0.019} \\ \cmidrule(l){2-11}
	\multicolumn{1}{c|}{}                                         & \multicolumn{1}{c|}{\multirow{2}{*}{3D}} & MSE                                    & 1.599$\pm$0.000                         & {1.043$\pm$0.018}                     & 1.747$\pm$0.000 & {1.033$\pm$0.013} & 1.616$\pm$0.000 & {1.024$\pm$0.016} & 1.511$\pm$0.000 & {0.898$\pm$0.015} \\
	\multicolumn{1}{c|}{}                                         & \multicolumn{1}{c|}{}                    & MAE                                    & 0.994$\pm$0.000                         & {0.787$\pm$0.013}                     & 1.024$\pm$0.000 & {0.772$\pm$0.016} & 0.992$\pm$0.000 & {0.762$\pm$0.010} & 0.967$\pm$0.000 & {0.726$\pm$0.014} \\ \cmidrule(l){2-11}
	\multicolumn{1}{c|}{}                                         & \multicolumn{1}{c|}{\multirow{2}{*}{4D}} & MSE                                    & 1.381$\pm$0.000                         & {0.913$\pm$0.014}                     & 1.409$\pm$0.000 & {0.869$\pm$0.015} & 1.461$\pm$0.000 & {0.880$\pm$0.017} & 1.358$\pm$0.000 & {0.800$\pm$0.014} \\
	\multicolumn{1}{c|}{}                                         & \multicolumn{1}{c|}{}                    & MAE                                    & 0.920$\pm$0.000                         & {0.732$\pm$0.013}                     & 0.934$\pm$0.000 & {0.715$\pm$0.014} & 0.942$\pm$0.000 & {0.704$\pm$0.011} & 0.909$\pm$0.000 & {0.673$\pm$0.008} \\ \cmidrule(l){2-11}
	\multicolumn{1}{c|}{}                                         & \multicolumn{1}{c|}{\multirow{2}{*}{7D}} & MSE                                    & 1.142$\pm$0.000                         & {0.795$\pm$0.015}                     & 1.124$\pm$0.000 & {0.802$\pm$0.015} & 1.169$\pm$0.000 & {0.756$\pm$0.017} & 1.116$\pm$0.000 & {0.754$\pm$0.021} \\
	\multicolumn{1}{c|}{}                                         & \multicolumn{1}{c|}{}                    & MAE                                    & 0.821$\pm$0.000                         & {0.676$\pm$0.011}                     & 0.826$\pm$0.000 & {0.675$\pm$0.012} & 0.838$\pm$0.000 & {0.650$\pm$0.006} & 0.822$\pm$0.000 & {0.646$\pm$0.012} \\ \bottomrule
\end{tabular}
	}
	\label{tab13}
\end{table}

\begin{table}[t]
	\centering
	\caption{The prediction errors of ChronoSteer across various time series foundation models from different families. The historical windows vary across \{2, 3, 4, 7\} days, while prediction window is fixed at 1. A lower value denotes superior accuracy.}
	\resizebox{0.9\linewidth}{!}
	{
		\begin{tabular}{@{}ccc|cc|cc|cc|cc@{}}
	\toprule
	\multicolumn{3}{c|}{\multirow{2}{*}{Method}}                  & \multicolumn{2}{c|}{timesfm-1.0-200m}    & \multicolumn{2}{c|}{timesfm-2.0-500m} & \multicolumn{2}{c|}{moirai-1.0-large} & \multicolumn{2}{c}{moirai-1.1-large}                                                                                                                   \\ \cmidrule(l){4-11}
	\multicolumn{3}{c|}{}                                         & Unimodal                                 & Multimodal                            & Unimodal                              & Multimodal                           & Unimodal        & Multimodal        & Unimodal        & Multimodal                                              \\ \midrule
	\multicolumn{1}{c|}{\multirow{8}{*}{\rotatebox{90}{Energy}}}  & \multicolumn{1}{c|}{\multirow{2}{*}{2D}} & MSE                                   & 0.804$\pm$0.000                       & {0.585$\pm$0.010}                    & 0.728$\pm$0.000 & {0.574$\pm$0.011} & 0.838$\pm$0.017 & {0.587$\pm$0.011} & 0.754$\pm$0.011 & {0.616$\pm$0.008} \\
	\multicolumn{1}{c|}{}                                         & \multicolumn{1}{c|}{}                    & MAE                                   & 0.302$\pm$0.000                       & {0.283$\pm$0.005}                    & 0.288$\pm$0.000 & {0.253$\pm$0.006} & 0.313$\pm$0.005 & {0.296$\pm$0.005} & 0.294$\pm$0.003 & {0.274$\pm$0.002} \\ \cmidrule(l){2-11}
	\multicolumn{1}{c|}{}                                         & \multicolumn{1}{c|}{\multirow{2}{*}{3D}} & MSE                                   & 0.713$\pm$0.000                       & {0.606$\pm$0.011}                    & 0.695$\pm$0.000 & {0.576$\pm$0.013} & 0.797$\pm$0.013 & {0.607$\pm$0.010} & 0.716$\pm$0.012 & {0.586$\pm$0.007} \\
	\multicolumn{1}{c|}{}                                         & \multicolumn{1}{c|}{}                    & MAE                                   & 0.361$\pm$0.000                       & {0.291$\pm$0.004}                    & 0.320$\pm$0.000 & {0.274$\pm$0.004} & 0.289$\pm$0.005 & {0.245$\pm$0.002} & 0.281$\pm$0.005 & {0.265$\pm$0.003} \\ \cmidrule(l){2-11}
	\multicolumn{1}{c|}{}                                         & \multicolumn{1}{c|}{\multirow{2}{*}{4D}} & MSE                                   & 0.671$\pm$0.000                       & {0.601$\pm$0.009}                    & 0.688$\pm$0.000 & {0.627$\pm$0.009} & 0.776$\pm$0.012 & {0.693$\pm$0.009} & 0.727$\pm$0.014 & {0.645$\pm$0.008} \\
	\multicolumn{1}{c|}{}                                         & \multicolumn{1}{c|}{}                    & MAE                                   & 0.280$\pm$0.000                       & {0.261$\pm$0.004}                    & 0.279$\pm$0.000 & {0.247$\pm$0.006} & 0.276$\pm$0.003 & {0.256$\pm$0.002} & 0.284$\pm$0.004 & {0.271$\pm$0.002} \\ \cmidrule(l){2-11}
	\multicolumn{1}{c|}{}                                         & \multicolumn{1}{c|}{\multirow{2}{*}{7D}} & MSE                                   & 0.629$\pm$0.000                       & {0.528$\pm$0.009}                    & 0.581$\pm$0.000 & {0.515$\pm$0.008} & 0.621$\pm$0.009 & {0.575$\pm$0.008} & 0.620$\pm$0.010 & {0.544$\pm$0.011} \\
	\multicolumn{1}{c|}{}                                         & \multicolumn{1}{c|}{}                    & MAE                                   & 0.319$\pm$0.000                       & {0.256$\pm$0.005}                    & 0.288$\pm$0.000 & {0.257$\pm$0.004} & 0.254$\pm$0.004 & {0.230$\pm$0.004} & 0.259$\pm$0.003 & {0.243$\pm$0.003} \\ \midrule
	\multicolumn{1}{c|}{\multirow{8}{*}{\rotatebox{90}{Traffic}}} & \multicolumn{1}{c|}{\multirow{2}{*}{2D}} & MSE                                   & 0.588$\pm$0.000                       & {0.527$\pm$0.006}                    & 0.681$\pm$0.000 & {0.491$\pm$0.009} & 0.928$\pm$0.012 & {0.581$\pm$0.015} & 0.853$\pm$0.013 & {0.538$\pm$0.016} \\
	\multicolumn{1}{c|}{}                                         & \multicolumn{1}{c|}{}                    & MAE                                   & 0.511$\pm$0.000                       & {0.483$\pm$0.007}                    & 0.568$\pm$0.000 & {0.525$\pm$0.007} & 0.713$\pm$0.013 & {0.567$\pm$0.015} & 0.655$\pm$0.008 & {0.529$\pm$0.008} \\ \cmidrule(l){2-11}
	\multicolumn{1}{c|}{}                                         & \multicolumn{1}{c|}{\multirow{2}{*}{3D}} & MSE                                   & 0.524$\pm$0.000                       & {0.423$\pm$0.007}                    & 0.544$\pm$0.000 & {0.406$\pm$0.007} & 0.645$\pm$0.012 & {0.442$\pm$0.008} & 0.655$\pm$0.007 & {0.421$\pm$0.006} \\
	\multicolumn{1}{c|}{}                                         & \multicolumn{1}{c|}{}                    & MAE                                   & 0.477$\pm$0.000                       & {0.434$\pm$0.007}                    & 0.516$\pm$0.000 & {0.428$\pm$0.009} & 0.572$\pm$0.010 & {0.479$\pm$0.011} & 0.569$\pm$0.010 & {0.488$\pm$0.006} \\ \cmidrule(l){2-11}
	\multicolumn{1}{c|}{}                                         & \multicolumn{1}{c|}{\multirow{2}{*}{4D}} & MSE                                   & 0.486$\pm$0.000                       & {0.336$\pm$0.009}                    & 0.526$\pm$0.000 & {0.365$\pm$0.008} & 0.607$\pm$0.008 & {0.421$\pm$0.006} & 0.522$\pm$0.007 & {0.431$\pm$0.007} \\
	\multicolumn{1}{c|}{}                                         & \multicolumn{1}{c|}{}                    & MAE                                   & 0.465$\pm$0.000                       & {0.436$\pm$0.008}                    & 0.492$\pm$0.000 & {0.428$\pm$0.008} & 0.551$\pm$0.008 & {0.446$\pm$0.007} & 0.482$\pm$0.007 & {0.447$\pm$0.005} \\ \cmidrule(l){2-11}
	\multicolumn{1}{c|}{}                                         & \multicolumn{1}{c|}{\multirow{2}{*}{7D}} & MSE                                   & 0.330$\pm$0.000                       & {0.257$\pm$0.003}                    & 0.294$\pm$0.000 & {0.228$\pm$0.005} & 0.487$\pm$0.008 & {0.324$\pm$0.005} & 0.421$\pm$0.005 & {0.258$\pm$0.004} \\
	\multicolumn{1}{c|}{}                                         & \multicolumn{1}{c|}{}                    & MAE                                   & 0.387$\pm$0.000                       & {0.340$\pm$0.006}                    & 0.361$\pm$0.000 & {0.324$\pm$0.006} & 0.472$\pm$0.007 & {0.373$\pm$0.007} & 0.459$\pm$0.007 & {0.338$\pm$0.006} \\ \midrule
	\multicolumn{1}{c|}{\multirow{8}{*}{\rotatebox{90}{Weather}}} & \multicolumn{1}{c|}{\multirow{2}{*}{2D}} & MSE                                   & 1.627$\pm$0.000                       & {1.541$\pm$0.023}                    & 1.565$\pm$0.000 & {1.319$\pm$0.020} & 2.103$\pm$0.031 & {1.593$\pm$0.028} & 2.536$\pm$0.051 & {1.559$\pm$0.052} \\
	\multicolumn{1}{c|}{}                                         & \multicolumn{1}{c|}{}                    & MAE                                   & 0.956$\pm$0.000                       & {0.903$\pm$0.012}                    & 0.973$\pm$0.000 & {0.907$\pm$0.014} & 1.129$\pm$0.022 & {0.998$\pm$0.015} & 1.196$\pm$0.015 & {0.923$\pm$0.016} \\ \cmidrule(l){2-11}
	\multicolumn{1}{c|}{}                                         & \multicolumn{1}{c|}{\multirow{2}{*}{3D}} & MSE                                   & 1.291$\pm$0.000                       & {0.896$\pm$0.019}                    & 1.307$\pm$0.000 & {0.910$\pm$0.021} & 1.794$\pm$0.020 & {1.002$\pm$0.035} & 1.996$\pm$0.027 & {1.289$\pm$0.027} \\
	\multicolumn{1}{c|}{}                                         & \multicolumn{1}{c|}{}                    & MAE                                   & 0.878$\pm$0.000                       & {0.744$\pm$0.010}                    & 0.886$\pm$0.000 & {0.734$\pm$0.017} & 1.042$\pm$0.020 & {0.743$\pm$0.012} & 1.081$\pm$0.016 & {0.751$\pm$0.016} \\ \cmidrule(l){2-11}
	\multicolumn{1}{c|}{}                                         & \multicolumn{1}{c|}{\multirow{2}{*}{4D}} & MSE                                   & 1.259$\pm$0.000                       & {0.834$\pm$0.019}                    & 1.157$\pm$0.000 & {0.811$\pm$0.016} & 1.463$\pm$0.024 & {0.859$\pm$0.023} & 1.774$\pm$0.010 & {0.989$\pm$0.023} \\
	\multicolumn{1}{c|}{}                                         & \multicolumn{1}{c|}{}                    & MAE                                   & 0.871$\pm$0.000                       & {0.687$\pm$0.017}                    & 0.836$\pm$0.000 & {0.693$\pm$0.012} & 0.929$\pm$0.023 & {0.712$\pm$0.013} & 1.009$\pm$0.013 & {0.646$\pm$0.011} \\ \cmidrule(l){2-11}
	\multicolumn{1}{c|}{}                                         & \multicolumn{1}{c|}{\multirow{2}{*}{7D}} & MSE                                   & 1.198$\pm$0.000                       & {0.789$\pm$0.017}                    & 1.065$\pm$0.000 & {0.743$\pm$0.015} & 1.309$\pm$0.024 & {0.804$\pm$0.017} & 1.416$\pm$0.018 & {0.796$\pm$0.018} \\
	\multicolumn{1}{c|}{}                                         & \multicolumn{1}{c|}{}                    & MAE                                   & 0.838$\pm$0.000                       & {0.671$\pm$0.017}                    & 0.804$\pm$0.000 & {0.658$\pm$0.013} & 0.877$\pm$0.015 & {0.685$\pm$0.012} & 0.897$\pm$0.020 & {0.684$\pm$0.013} \\ \bottomrule
\end{tabular}
	}
	\label{tab14}
\end{table}

In the primary experiment, the unimodal TSFM Chronos-Bolt-base serves as the backbone for ChronoSteer. To explore how varying scales of unimodal TSFM affect prediction accuracy, Chronos-Bolt-base is replaced with several other models differing in parameter size. Specifically, four models are tested: Chronos-Bolt-tiny, Chronos-Bolt-mini, Chronos-Bolt-small, and Chronos-Bolt-base, with parameter counts increasing progressively.

The experiment results in Table \ref{tab13} show that prediction accuracy does not clearly improve with an increase in the parameter of backbone, for either unimodal or multimodal tasks. However, our method demonstrates strong adaptability across backbones with different scales, consistently boosting multimodal prediction accuracy relative to unimodal results in all cases. Specifically, applying the proposed method improves prediction accuracy for Chronos-Bolt-tiny, Chronos-Bolt-mini, Chronos-Bolt-small, and Chronos-Bolt-base by 19.8\%, 23.1\%, 22.4\%, and 25.8\%, respectively.

Furthermore, to validate the generalization of ChronoSteer across different architectures, we evaluate its performance on various families of unimodal TSFMs. The results are summarized in Table \ref{tab14}. Consistent with the findings on model scales, our method yields uniform improvements, boosting the performance of timesfm-1.0-200m, timesfm-2.0-500m, moirai-1.0-large, and moirai-1.1-large by 18.2\%, 19.1\%, 26.6\%, and 28.9\%, respectively.

\section{Detailed Prediction Showcase}
\label{appendix:Detailed_Prediction_Showcase}

\subsection{Prediction Showcase on Code Book}
\label{subappendix:Prediction_Showcase_on_Code_Book}

\begin{figure}[t]
	\centering
	\resizebox{0.94\linewidth}{!}
	{
		\includegraphics{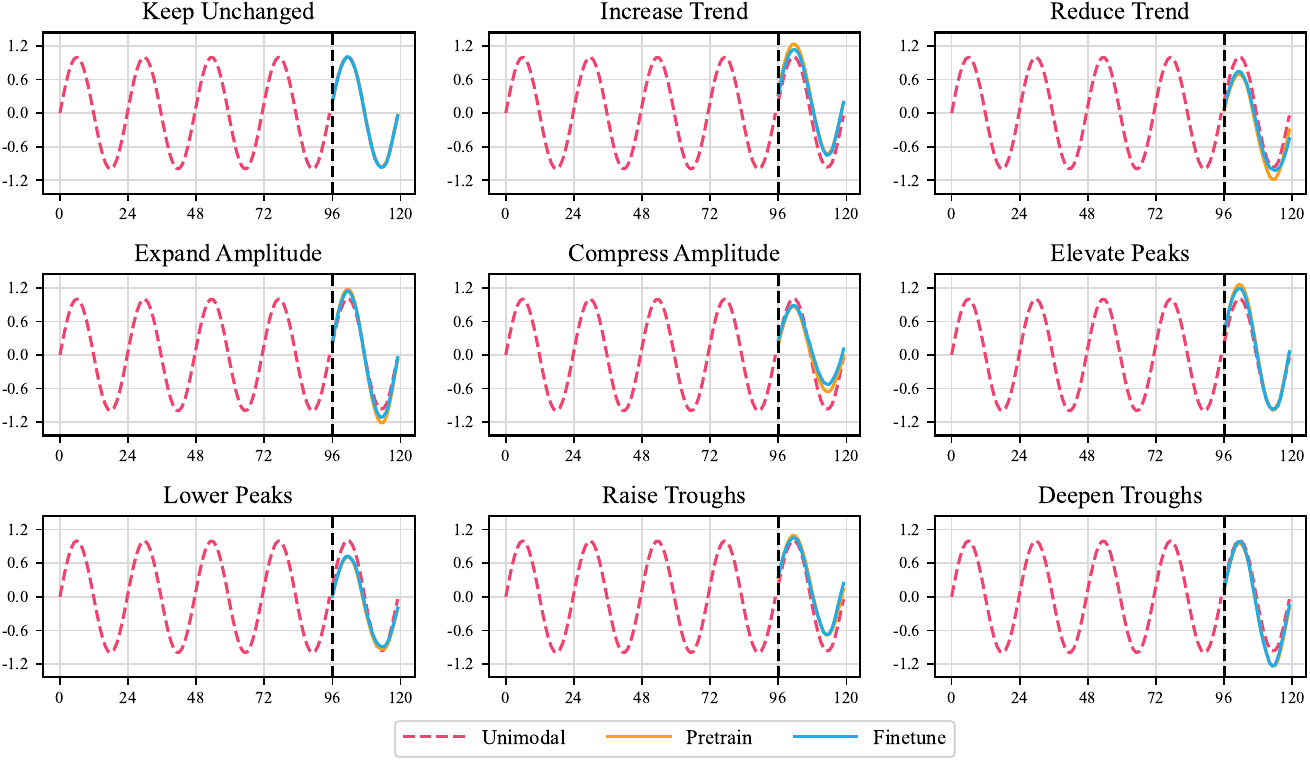}
	}
	\caption{The illustration of prediction showcases for ChronoSteer on the anchor revision instructions. The historical and prediction windows are fixed at 96 and 24 time steps, respectively.}
	\label{fig4}
\end{figure}

Figure \ref{fig4} illustrates the adjustment effects of nine anchor revision instructions. As shown, pre-training objectives rely on future series synthesized via functions, so adjustments by ChronoSteer after pre-training generally align with these transformation functions. However, after being fine-tuned on observed future series with pseudo-labeling techniques, ChronoSteer effectively recovers the fine-grained magnitude information lost during discretization and mitigates the expressive limitations of the synthetic transformation functions. It balances historical data and revision instructions better, yielding more accurate and realistic predictions. This improves its practical value in real-world scenarios.

\subsection{Prediction Showcase on Energy}
\label{subappendix:Prediction_Showcase_on_Energy}

\begin{figure}[t]
	\centering
	\resizebox{0.94\linewidth}{!}
	{
		\includegraphics{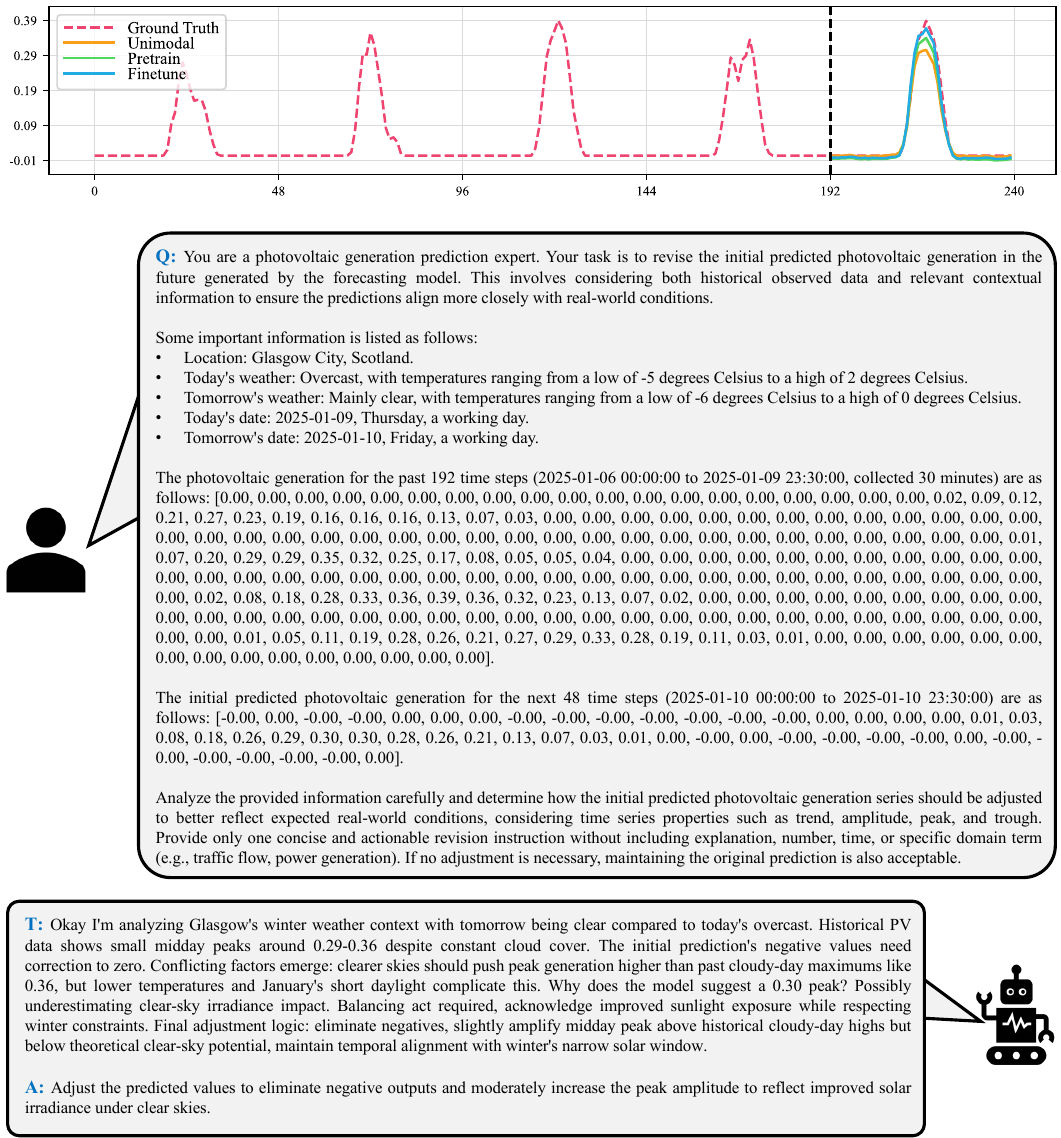}
	}
	\caption{The illustration of prediction showcases for ChronoSteer on the Energy sub-dataset, including complete context prompt, thought content, and revision instructions. The historical and prediction windows are fixed at 4 days and 1 day.}
	\label{fig5}
\end{figure}

Figure \ref{fig5} presents the comprehensive prediction visualization for the Energy dataset, comparing a 4-day historical series with a 1-day predicted series. The predictions include results from the unimodal TSFM, ChronoSteer after pre-training, and ChronoSteer after fine-tuning. Additionally, a detailed description of the prompt, reasoning process, and revision instructions for the Energy dataset is provided to support result replication. The foggy weather causes a gradual decline in peaks within the historical series. Without textual context, a unimodal TSFM continues this trend, forecasting even lower peaks. The weather forecast indicates \textit{"transitioning to clear skies tomorrow"}, so the LLM provides a revision instruction to raise the peaks and steers ChronoSteer to adjust its prediction. After fine-tuning, the adjustment magnitude more closely matches the observed future series.

\subsection{Prediction Showcase on Traffic}
\label{subappendix:Prediction_Showcase_on_Traffic}

\begin{figure}[t]
	\centering
	\resizebox{0.94\linewidth}{!}
	{
		\includegraphics{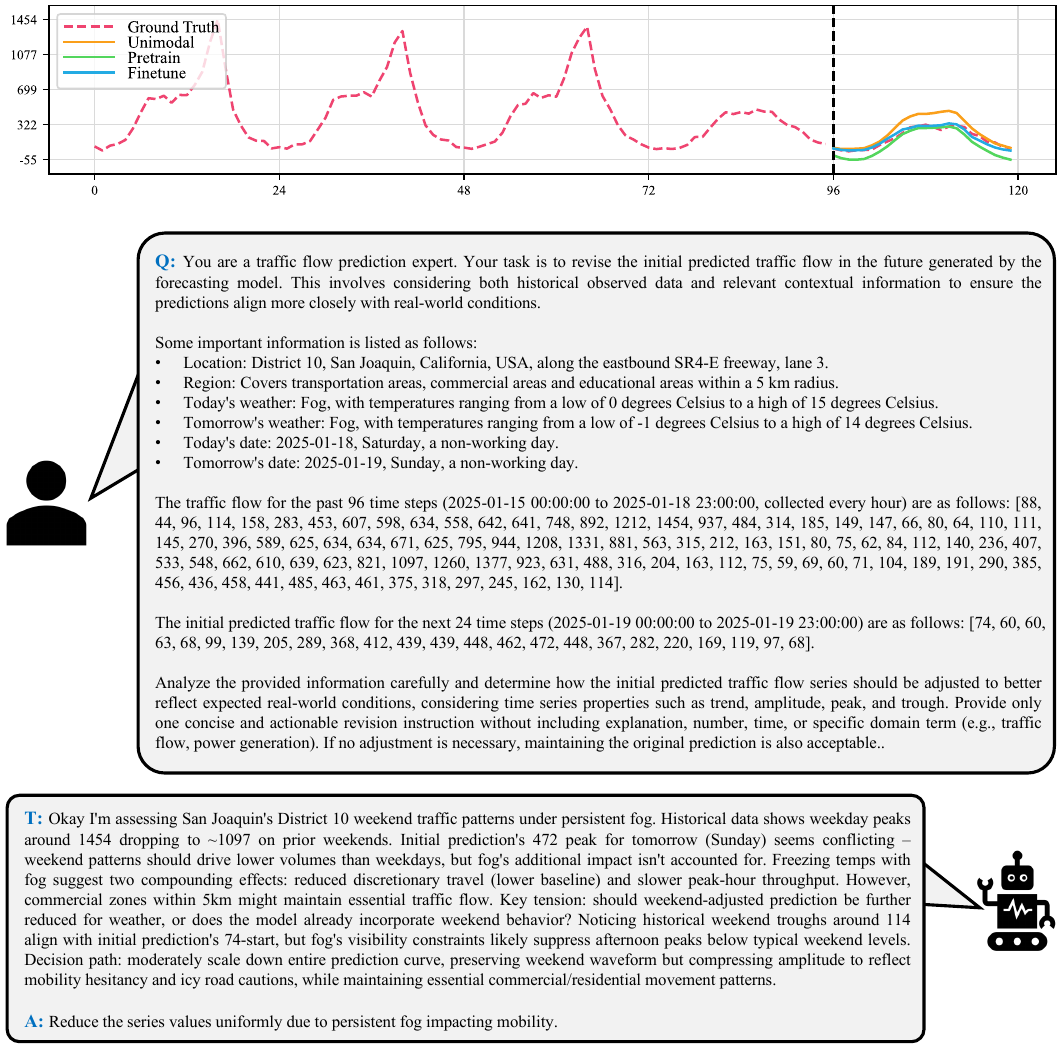}
	}
	\caption{The illustration of prediction showcases for ChronoSteer on the Traffic sub-dataset, including complete context prompt, thought content, and revision instructions. The historical and prediction windows are fixed at 4 days and 1 day.}
	\label{fig6}
\end{figure}

Figure \ref{fig6} presents the comprehensive prediction visualization for the Traffic dataset, comparing a 4-day historical series with a 1-day predicted series. The predictions include results from the unimodal TSFM, ChronoSteer after pre-training, and ChronoSteer after fine-tuning. Additionally, a detailed description of the prompt, reasoning process, and revision instructions for the Traffic dataset is provided to support result replication. The persistent heavy fog causes a further reduction in traffic flow peaks. Without textual information, a unimodal TSFM predicts a typical non-working day pattern similar to the previous day. The LLM provides the revision instruction \textit{"reduce series uniformly"}, and ChronoSteer adjusts its prediction accordingly. However, this instruction is imprecise, as it should target only peaks rather than the entire series. After pre-training, ChronoSteer strictly follows the instruction, resulting in unrealistic troughs. In contrast, the fine-tuned version integrates the instruction with historical series and partly mitigates this issue. This behavior improves the practical value of ChronoSteer in real-world scenarios.

\subsection{Prediction Showcase on Weather}
\label{subappendix:Prediction_Showcase_on_Weather}

\begin{figure}[t]
	\centering
	\resizebox{0.94\linewidth}{!}
	{
		\includegraphics{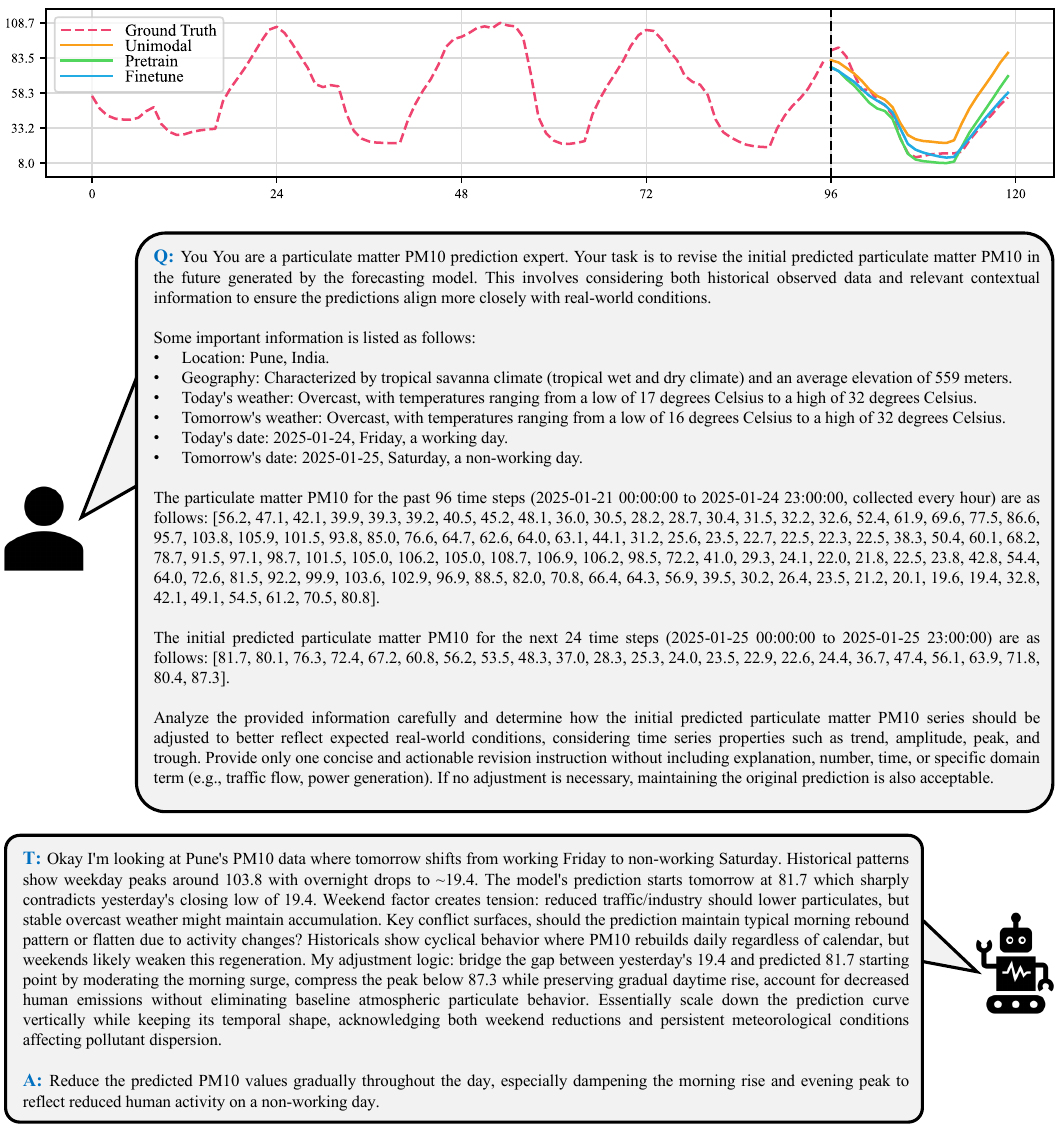}
	}
	\caption{The illustration of prediction showcases for ChronoSteer on the Weather sub-dataset, including complete context prompt, thought content, and revision instructions. The historical and prediction windows are fixed at 4 days and 1 day.}
	\label{fig7}
\end{figure}

Figure \ref{fig7} presents the comprehensive prediction visualization for the Weather dataset, comparing a 4-day historical series with a 1-day predicted series. The predictions include results from the unimodal TSFM, ChronoSteer after pre-training, and ChronoSteer after fine-tuning. Additionally, a detailed description of the prompt, reasoning process, and revision instructions for the Weather dataset is provided to support result replication. The transition from weekdays to non-working days reduces air pollution due to decreased human activity. Without textual information, a unimodal TSFM struggles to capture this pattern, leading to overestimated predictions. By integrating revision instructions, ChronoSteer lowers the prediction and improves accuracy. Compared to the pre-trained model, the fine-tuned ChronoSteer predicts troughs and peaks more accurately.

\end{document}